\documentclass[11pt]{report}
\usepackage{adjustbox,lipsum}
\usepackage{graphicx}
\usepackage{fullpage}
\usepackage{setspace}

\usepackage{amsthm}
\usepackage{subcaption}
\usepackage{latexsym}
\usepackage{helvet}
\usepackage{courier}
\usepackage{amsmath}
\usepackage{amssymb}
\usepackage{multirow}
\usepackage{mdwlist}
\usepackage{setspace}
\usepackage{tabularx}
\usepackage[noend]{algorithmic}
\usepackage{algorithm}
\usepackage{multirow}
\usepackage{bm}
\usepackage{longtable}

\usepackage{soul}
\usepackage{graphicx}
\usepackage{graphics}
\usepackage{latexsym}
\usepackage{helvet}
\usepackage{courier}
\usepackage{amsmath}
\usepackage{amssymb}
\usepackage{url}
\usepackage[noend]{algorithmic}
\usepackage{algorithm}
\usepackage{multirow}
\usepackage{mathtools}
\usepackage{enumitem}
\usepackage{color}
\usepackage{microtype}
\usepackage{tabularx}

\usepackage{amsthm}

\newcommand{\MYTITLE}{{Optimal Decision-Making in Mixed-Agent Partially Observable Stochastic Environments via Reinforcement Learning
}}

\newtheorem{thm}{Theorem}

\newtheorem{prop}{Proposition}
\newtheorem{defn}{Definition}

\newcommand{\mcesp}{\textsf{MCES-P}}
\newcommand{\mcesip}{\textsf{MCES-IP}}
\newcommand{\mcesmp}{\textsf{MCES-MP}}
\newcommand{\mcesppac}{\textsf{MCESP+PAC}}
\newcommand{\mcesmppac}{\textsf{MCESMP+PAC}}
\newcommand{\mcesippac}{\textsf{MCESIP+PAC}}
\newcommand\numberthis{\addtocounter{equation}{1}\tag{\theequation}}

\usepackage{blindtext}

\hypersetup{pdfinfo={
Title={\MYTITLE},
Author={Roi Ceren},
Subject={Dissertation},
Keywords={human modeling,
precision agriculture,
reinforcement learning,
model-free learning,
probably approximately correct learning}
}}

\makeatletter
\def\@part[#1]#2{%
    \ifnum \c@secnumdepth >-2\relax
      \refstepcounter{part}%
      \addcontentsline{toc}{part}{\partname~\thepart\hspace{1em}#1}%
    \else
      \addcontentsline{toc}{part}{#1}%
    \fi
    \markboth{}{}%
    {\centering
     \interlinepenalty \@M
     \normalfont
     \ifnum \c@secnumdepth >-2\relax
       \huge\bfseries \partname\nobreakspace\thepart
       \par
       \vskip 20\p@
     \fi
     \Huge \bfseries #2\par}%
    \@endpart}
\makeatother

\begin{document}

\newpage
\pagestyle{empty}
\vspace*{18pt}
\begin{center}
\textsc{\MYTITLE}\\[18pt]
by\\[18pt]
\textsc{Roi Ceren}\\[12pt]
(Under the Direction of Shannon Quinn)\\[12pt]
\textsc{Abstract}
\end{center}

Optimal decision making with limited or no information in stochastic environments where multiple agents interact is a challenging topic in the realm of artificial intelligence. Reinforcement learning (RL) is a popular approach for arriving at optimal strategies by predicating stimuli, such as the reward for following a strategy, on experience. RL is heavily explored in the single-agent context, but is a nascent concept in multiagent problems. To this end, I propose several principled model-free and partially model-based reinforcement learning approaches for several multiagent settings. In the realm of normative reinforcement learning, I introduce scalable extensions to Monte Carlo exploring starts for partially observable Markov Decision Processes (POMDP), dubbed \mcesp{}, where I expand the theory and algorithm to the multiagent setting. 
I first examine \mcesp{} with probably approximately correct (PAC) bounds in the context of multiagent setting, showing \mcesppac{} holds in the presence of other agents. I then propose a more sample-efficient methodology for antagonistic settings, \mcesippac{}. For cooperative settings, I extend \mcesp{} to the Multiagent POMDP, dubbed \mcesmppac{}.
I then explore the use of reinforcement learning as a methodology in searching for optima in realistic and latent model environments. First, I explore a parameterized Q-learning approach in modeling humans learning to reason in an uncertain, multiagent environment. Next, I propose an implementation of \mcesp{}, along with image segmentation, to create an adaptive team-based reinforcement learning technique to positively identify the presence of phenotypically-expressed water and pathogen stress in crop fields.

\singlespacing

\begin{list}{\sc Index words:\hfill}{\labelwidth 1.2in\leftmargin 1.4in\labelsep 0.2in}
\item 
\begin{flushleft}\singlespacing
human modeling,
precision agriculture,
reinforcement learning,
model-free learning,
probably approximately correct learning
\end{flushleft}
\end{list}

\newpage
\thispagestyle{empty}

\vspace*{18pt}
\begin{center}
\textsc{\MYTITLE}\\[18pt]
by\\[18pt]
\textsc{Roi Ceren}\\[12pt]
B.S., The University of Georgia, Athens, GA, 2010\\
\vfill
A Dissertation Submitted to the Graduate Faculty \\
of The University of Georgia in Partial Fulfillment \\
of the \\
Requirements for the Degree \\[10pt]
\textsc{Doctor of Philosophy}\\[36pt]
\textsc{Athens, Georgia}\\[18pt]
2018\\[18pt]
\end{center}

\newpage
\thispagestyle{empty}
\vspace*{5.5in}
\begin{center}
\copyright ~2018 \\
Roi Ceren\\
All Rights Reserved
\end{center}

\newpage
\thispagestyle{empty}
\vspace*{18pt}
\begin{center}
\textsc{\MYTITLE}\\[18pt]
by\\[18pt]
\textsc{Roi Ceren}
\end{center}
\vfill
\begin{flushleft}\singlespacing
\hskip 200pt {Approved:}\\
\vskip 12pt
\hspace*{200pt}\makebox[100pt][l]{Major Professor:}Shannon Quinn\\
\vskip 12pt
\hspace*{200pt}\makebox[100pt][l]{Committee:       }Khaled Rasheed\\
\hspace*{200pt}\makebox[100pt][l]{~                }Glen Rains\\
\vfill
Electronic Version Approved:\\[12pt]
Suzanne Barbour\\
Dean of the Graduate School\\
The University of Georgia\\
December 2018
\end{flushleft}

\newpage
\pagestyle{plain}
\pagenumbering{roman}
\setcounter{page}{4}
\vspace*{3.5in}
\begin{center}
{\large Dedicated to the people who drive me to pursue my dreams: \\my wife Alex, my son Peter, my mother Esther, and my alter egos Nathan and Robert.}
\end{center}

\title{\bf \MYTITLE}
\author{Roi Ceren}

\newpage

\pagestyle{plain}

\chapter*{Acknowledgments}
\addcontentsline{toc}{chapter}{Acknowledgments}

For my academic pursuits, I must primarily acknowledge the patience, kindness, and selflessness of my advisor, Shannon Quinn. Among all other faculty, no one has shown greater flexibility and eagerness to help me succeed when I struggled the most to find my place. Peerless in his acumen for the mix between thought leadership and programmatic skill, Shannon will continue to be a colleague and inspiration as I pursue a blend of research with delivery in my scholastic and industrial career. I thank my former advisor, Prashant Doshi, for 6 years of guidance.

Next, I thank my employer, SalesLoft, specifically my direct manager and CTO, Scott Mitchell, for being flexible, a sounding board, and, importantly, invested in my future. I feel remarkably blessed to perform as the current data science lead. To my many brilliant peers, including Kyle Bock, Sean Ogawa, Drew Pfundstein, Mike Sandt, Tyler Howard, and Sneha Subramanian, I couldn't succeed at work or in academics without you.

Most importantly, I thank my family, both blood and otherwise. I thank my neighbors, Robert and Beth Wilson, who I purchased a house to be near, and who became the launchpad for my career and renewed performance in academics. I thank all the Rays, Nathan and Robin, Nick and Christine, Jerry and Kathoise, who made me part of their family. I thank my mother and siblings, Esther, Merav, and Omri, for driving me my whole life to be the best I can be. Above all others, I thank my wife, Alexandria, whose contributions would produce a document exceeding the size of this dissertation. I thank you for your patience, your love, your support, and the son you are having with me. You give my life purpose.

\chapter*{Publications}
\addcontentsline{toc}{chapter}{Publications}
\begin{enumerate}

\item Roi Ceren, Shannon Quinn, Glen Rains. "Towards a Decentralized, Autonomous Multiagent Framework for Mitigating Crop Loss". \underline{In Preparation}

\item Roi Ceren, Scott Mitchell. "On Deriving Optimal Cadences via Engagement Score Maximization". \underline{In Preparation}, http://www2.salesloft.com/derived-cadences (2018)

\item Roi Ceren, Prashant Doshi, Keyang He. "Reinforcement Learning for Heterogeneous Teams with PALO Bounds", \underline{In Preparation}, arXiv preprint arXiv:1805.09267 (2018).

\item Roi Ceren, Prashant Doshi, and Bikramjit Banerjee. "Reinforcement Learning in Partially Observable Multiagent Settings: Monte Carlo Exploring Starts with PAC Bounds" To appear in \textit{Proceedings of the International Conference on Autonomous Agents and Multi-Agent Systems (AAMAS)}, 2016, Singapore, Singapore.

\item Adam Goodie, Matthew Meisel, Roi Ceren, Daniel Hall, and Prashant Doshi. "Evaluating and Improving Probability Assessment in an Ambiguous, Sequential Environment" In \textit{Current Psychology}, pp. 1-11, 2015.

\item Shu Zhang, Roi Ceren, and Khaled Rasheed. "EVOLMUSIC - A Preference Learning Accompanist" In \textit{Proceedings of the International Conference on Genetic and Evolutionary Methods (GEM)}, 2014.

\item Roi Ceren, Prashant Doshi, Matthew Meisel, Adam Goodie, and Daniel Hall. "On Modeling Human Learning in Sequential Games with Delayed Reinforcements" In \textit{IEEE International Conference on Systems, Man, and Cybernetics (IEEE SMC)}, pp. 3108-3113, 2013, Manchester, UK.

\end{enumerate}

\setcounter{tocdepth}{1}
\tableofcontents
\cleardoublepage
\addcontentsline{toc}{chapter}{\listfigurename}
\listoffigures  
\cleardoublepage
\addcontentsline{toc}{chapter}{\listtablename}
\listoftables 



\newpage
\pagenumbering{arabic}  

\chapter{Introduction}
\label{chap:intro}

In the vast majority of real-world scenarios, decision makers are faced with performing under limited or no information, where stimuli upon interacting with the environment is the only actionable metric to drive action. Reinforcement learning (RL) is a powerful tool for reasoning in the context of machine learning, derived from countless decades of research on human behavior~\cite{sutton}. By predicating stimuli solely on observations of an environment with features that are largely or entirely latent, decision makers can learn to behave optimally with regards to the elicited information~\cite{rl}.

For this reason, I investigate advancements in contemporary reinforcement learning research along two axis: theoretical contributions to advancing the search for optima under extremely limited information and, additionally, the performance of contextualized reinforcement learning in modeling mixed-agent environments.

\section{Advancing Theory in Multiagent Reinforcement Learning}

Making the best decision in a realistic, stochastic environment is a complex and computationally expensive task, even when the mechanics of that environment are known. The presence of other agents, also interacting and affecting the state of the environment, further complicates the process of selecting the most valuable behavior to adopt. In real world situations, the knowledge of how an environment works, even when acting alone, is highly unlikely to be available. In these situations, the agent can search for optimal behavior only through interaction, learning the value of executing strategies through experience.

The algorithms proposed in Chapters~\ref{chap:rlmas} and~\ref{chap:instantiate} are designed to accomplish the highly complex task of searching for optimal strategies in environments burdened by partial observability and several agents acting simultaneously. In contemporary works, even the task of optimal planning in the face of such stochasticity is a difficult task. I take this setting, and its associated complexity, to an even higher level of difficulty, where the details of stochasticity are not known to subject agents. With the application of reinforcement learning and principled sampling techniques, I propose several methods for arriving at optimal strategies under a variety of perspectives in multiagent systems (MAS), inspired by Monte Carlo Exploring Starts for POMDPs (\mcesp{})~\cite{perkins}, a method for model-free learning and hill-climbing to local optima for partially observable Markov Decision Processes (POMDPs).

\begin{enumerate}
\item \mcesp{} with PAC bounds \mcesppac{}: I establish statistical bounds via probably approximately correct learning (PAC learning) for the existing \mcesp{} algorithm in the \textit{multiagent} setting, where the subject agent behaves as if the opponent does not exist, referred to as the single-agent perspective. Previously, guarantees for \mcesppac{} existed only for the single agent context.
\item MCES for POMDPs, \mcesmp{} and \mcesmppac{}: I introduce \mcesp{} to the instant communication team setting, multiagent POMDPs (MPOMDPs), where a centralized controller explores joint policies for two or more agents, also providing guarantees on the optimality of the converged policy.
\item MCES for Interactive POMDPs, \mcesip{} and \mcesippac{}: In the most complex of settings, I further expand Monte Carlo Exploring Starts to include a partially model-based approach in which the subject agent learns and reasons about the behavior of an opponent in an attempt to speed up convergence to local optima. As with previous methods, I establish PAC guarantees of optimality. 
\end{enumerate}

These three extensions fill a significant and necessary gap in the nascent field of learning optimal strategies in multiagent systems under uncertainty. \mcesmppac{} generalizes the POMDP solution methodology of \mcesppac{} to learning in a truly multiagent domain in a model-free fashion, marking a significant departure from previous literature. While \mcesppac{} can be straightforwardly extended to the multiagent context, the explicit modeling in \mcesippac{} shows significant gains in scalability in sample complexity, as well.

\section{Applying Reinforcement Learning as a Model of Human Reasoning}

Canonical reinforcement learning is versatile and powerful unmodified, representing an excellent method reflecting perfectly rational behavior. In strategic domains, however, humans express \textit{descriptive} reasoning, often deviating from rational behavior due to cognitive biases. Chapter~\ref{chap:psych} explores a process model adapted from reinforcement learning which captures the effect of these biases. Inspired by behavioral game theory, this descriptive reinforcement learning algorithm fits a model using principled computational psychology parameters over data collected from humans playing a strategic, sequential game. Several biases are represented in this computational model.

\begin{enumerate}
\item \textit{Forgetfulness}: The phenomenon whereby humans subnormatively deteriorate previous knowledge. Normative Q-learning contains the learning parameter \textit{$\alpha$}, which is already able to capture this effect.
\item \textit{Spillover}: Adapted from \textit{eligibility traces}, the descriptive RL algorithm allows for the propagation of stimuli received in a physical state to nearby states, replicating the erroneous behavior of humans misattributing an experience to nearby locations.
\item \textit{Subproportional Weighting}: Established in the field of prospect theory, humans tend to categorically under- and over-weight probabilities in domains involving chance. In our experiments, humans are asked to make assessments of success.
\end{enumerate}

I show that this descriptive reinforcement learning model reflects the behavior of these subjects well when faced with uncertainty in an uncertain, multiagent environment. By fitting the parameters with a portion of data collected from experiments, where human participants are tasked with assessing the probability of success in a strategic game, I compare a variety of RL approaches in predicting these judgments, and illustrate and discuss the performance of the best of these.

\section{Applying Team Reinforcement Learning to Precision Agriculture}

The low-cost availability of imaging technology has given rise to the rapidly developing field of precision agriculture, often marked by the use of multispectral image collection via autonomous uninhabited aerial vehicles (AUAVs)~\cite{precag}. As an example, recent efforts combining AUAVs, normalized difference vegetation index (NDVI) imaging, and environmental barometric and water potential sensors have been used to create efficient autonomous systems for targeted crop field watering~\cite{farmbeats}. Additionally, many targeted image processing systems have been developed for the purpose of specific disease identification based on phenotypic expression, such as lesions, browning, and tumors~\cite{plant_detection}.

While precision watering techniques have dramatically improved yields for large-scale farms, the advent of autonomous intervention for disease propagation is nascent~\cite{plant_detection}. While some generalized models exist to detect these stresses, they have not been introduced to the distributed autonomous systems as in precision watering. To that end, I adopt the problem of identifying and predicting the onset of stresses (pest and pathogen) in crop fields via environmental sensor and image data, taken at various resolutions throughout a growing season. I factor the distribution of functional capabilities of our physical system into four distinct layers, comprised of satellites (layer 3), AUAVs (L2), autonomous uninhabited ground vehicles (AUGVS, L1), and static ground-level sensors (L0). Generally, the output of each layer (excluding L0) is used to inform the decision making of the layer below it by raising a \textit{call-to-action}, wherein the layer believes a stress is occurring based on phenotypic expression that differs in a geographical location.

\begin{figure} [htb]
\label{fig:intro_layers}
\centerline{\includegraphics[width=8cm]{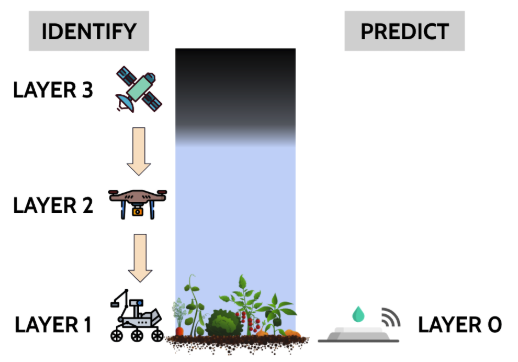}}
\caption{Layer composition of physical sensor modalities in the crop field problem domain. Images and Layer 0 data are used in each layer to inform subsequent layers of possible stress, where AUGVs are constrained by AUAVs, and AUAVs are constrained by satellites.}
\end{figure}

A common challenge of real-world problem domains, particularly in the agricultural domain, is the constraint on sample availability for machine learning. Since we are attempting to uncover the true state of stress in a crop field without prior knowledge, we propose a \textit{model-free} exploration of policies \textit{a la} Perkins' Monte Carlo Exploring Starts (MCES) for Partially Observable Markov Decision Processes (POMDPs), labeled \mcesp{}~\cite{perkins}. \mcesp{} iterates over memory-less policies that directly map actions to observations, instead of beliefs of the state~\cite{wiering}. As a first best effort towards our goal, we assemble our sensor modalities into a heterogeneous team, utilize an image processing technique to extract potentially stressed sectors, and learn policies that map these observations of phenotypic deviations to calls for intervention.

\section{Dissertation Organization}

The dissertation is organized as follows:

  \begin{center}
    \begin{longtable}{r|p{5in}}
  \hline
       Chapter~\ref{chap:intro} & Introduction to the model-free reinforcement learning domains that are tackled in this work.\\
       Chapter~\ref{chap:background} & Review of the background material utilized in this work, including relevant Markov models, approximation techniques, and state of the art in reinforcement learning techniques for multiagent systems and precision agriculture.\\
       Chapter~\ref{chap:relatedwork} & Review of related work in the fields of reinforcement learning, multiagent Markov models, and precision agriculture, in imaging and reasoning capacities.  Comparisons are made to the contributions of this dissertation.\\
       Chapter~\ref{chap:rlmas} & Methods for performing reasoning in multiagent systems via model-free reinforcement learning. I apply contemporary literature as a baseline, and expand theory with features capturing multiagent behavior.\\
       Chapter~\ref{chap:instantiate} & Expanded theory on Chap.~\ref{chap:rlmas} algorithms, ensuring statistical guarantees on optimality are met.\\
       Chapter~\ref{chap:psych} & Models for descriptively modifying reinforcement learning to capture latent features of human decision making for enhanced predictive accuracy.\\
       Chapter~\ref{chap:ag} & Aggregate reinforcement learning model for teams of agents, promoting capabilities to detect and predict the presence of a stress through image processing.\\
       Chapter~\ref{chap:appendix} & Appendix containing longform proofs of lemmas and propositions throughout the work.\\
    \hline
  \caption*{}
    \end{longtable} 
  \end{center}


\newpage
\chapter{Background}
\label{chap:background}

In this chapter, I introduce the relevant topics that define the scope of the learning algorithms proposed in the succeeding chapters. I first cover the frameworks, in order of generality, used to model agents interacting in an environment. I then introduce reinforcement learning as a departure point for agents learning optimal strategies in these environments. Next, I introduce a method for statistically guaranteeing the optimality of learned strategies. The following section covers a specific method leveraging reinforcement learning for POMDPs. I then cover necessary background for descriptive reinforcement learning as exhibited by humans in strategic environments. Lastly, I cover the state of the art in precision agriculture via imaging techniques.

\section{Models of Decision Making}
\label{sec:models}

The task of making decisions in a sequential, stochastic environment can be modeled with the Markov decision process (MDP) framework~\cite{mdp}, which describes the features of the environment, including the physical states of the environment, the possible actions available to the decision maker (hereafter called an \textit{agent}), the probabilities of transitioning between states given an action, and the rewards of states and, potentially, performing actions in them. Solutions to the MDP map the physical states of the environment to an action, and optimal search techniques over these solutions attempt to converge to one that maximizes the expected reward of following it.

In many cases, the physical state may be hidden from the subject agent, instead being revealed by a noisy observation, an extension called the partially observable MDP (POMDP)~\cite{pomdp}. The POMDP contains the same features as the MDP, but further includes observations the agent may receive and the probabilities of receiving some observation from a state. When interacting with the environment, the agent does not know which state it is in, but instead receives one of these observations. Solutions may map observations to actions or, given the observation function, may map states to actions and require inferring about the possible state given received observations. This inference is formalized in the belief update.

These frameworks, which are limited to agents interacting with environments alone, are extensively well-explored. The last decade has witnessed an explosion in focus on providing such formalizations to multiagent systems (MAS), which include environments of varying observability where multiple agents interact. The cooperative multiagent MDP (COM-MDP)~\cite{commdp}, in which agents are able to instantly communicate their actions to one another prior to execution, is an example of the simplest of such settings, but already illustrates an exponentially larger state and action space. The multiagent POMDP (MPOMDP)~\cite{mpomdp} is like the COM-MDP except the state is hidden. MPOMDPs are a special case of the more difficult decentralized POMDP (DEC-POMDP)~\cite{decpomdp}. Here, communication is not assumed, but is still within the confines of team settings, in that a joint team reward is given to all agents. The interactive POMDP (I-POMDP)~\cite{doshijair05} is one of the most computationally difficult of the multiagent generalizations of POMDPs. In addition to the exponentially larger state and action space, collaboration is not incentivized (and is, in fact, often suboptimal) and opponent behavior is difficult to predict (and must be inferred about). I-POMDP agents must reason not only about the state from observations, but also the motivations and actions of the opponent, anticipating their moves, potentially considering that the opponent may, in fact, be modeling the subject.

Formally, the POMDP is defined as a tuple $\langle \mathcal{S}, \mathcal{A}, \Omega, T, O, \mathcal{R} \rangle$, where $\mathcal{S}$ is the set of physical states, $\mathcal{A}$ defines the set of actions the subject agent may take, $\Omega$ is the set of observations the subject agent receives from physical states, $T: \mathcal{S}\times\mathcal{A}\rightarrow \mathcal{S}$ defines the transition probabilities of arriving in state $s'$ from state $s$ when taking action $a$. $\Omega$ defines the set of observations an agent may receive, with $O: \mathcal{S}\times\mathcal{A}\rightarrow\Omega$ defining the probabilities of receiving one from a state taking an action. The multiagent configurations covered in this dissertation includes the set of agents in the frame, $\mathcal{I}$. For the MPOMDP, $\mathcal{S}$, $\mathcal{A}$, and $\Omega$ are the joint of the states, actions, and observations over all agents $i\in\mathcal{I}$. The I-POMDP also includes $m_j\in M_j\triangleq \langle h_j, \pi_j \rangle$, which is the set of models, including a history of observation-action pairs and a policy for the opponent.

Even with knowledge of the mechanics of the environment, such as the transition and observation functions, finding optimal solutions to the variety of highly complex multiagent generalizations of POMDPs is very computationally expensive. However, the burden is significantly worsened without explicit knowledge of these functions. In my research, I propose frameworks for learning the optimal policy even in such settings (referred to as \textit{model-free} learning), a task that is often significantly more difficult than simply planning in stochastic environments.

\section{Reinforcement Learning}
\label{sec:rl}

In many real world scenarios, knowledge of the model of the environment in the MDP, such as the state transition or observation functions, are not known. In this case, the agent can rely instead on experience through its interactions, learning which behavior is most valuable through trial-and-error. While several supervised learning techniques exist, the assumption of an expert advising or correcting the subject agent heavily limits the scope of applicable problems. In the realm of unsupervised learning, no approach is more fundamental or widespread in machine learning literature as reinforcement learning~\cite{kaelbling}.

Simply, the behavior of a reinforcement learning agent can be characterized as a balance between \textit{exploring} the space of possible strategies (usually referred to as a \textit{policy}) and \textit{exploiting} the action-value knowledge acquired from interaction, selecting actions that maximize its expected reward~\cite{rl}. Consequently, the process of reinforcement learning is largely an \textit{online} one, where experience is acquired by an agent executing actions and recording their results.

Temporal difference learning ($TD(\lambda)$) calculates the value of being in a particular state of the MDP when executing a given policy (a configuration referred to as \textit{on-policy})~\cite{tdlambda}. This simple technique represents one of the earliest and most important reinforcement learning implementations applicable to MDPs. $TD(\lambda)$ is formalized in the following equation.

\begin{align}
\label{eq:td}
V(s;\alpha,\lambda)=V(s)+\alpha(r(s)+\gamma\cdot V(s')-V(s))e(s;\lambda)
\end{align}

Here, $\alpha$ is a learning rate (which can be fixed or depreciated over time to prioritize exploitation as more data is acquired), $r(s)$ is the immediate reward and $s'$ the resultant state when taking the action prescribed by the policy at state $s$, and $\gamma$ is a depreciation factor for future rewards. $e(s;\lambda)$ is defined as the \textit{eligibility trace}, representing a short-term memory of the interaction with state $s$, allowing the quick propagation of reward from future steps to eligible states~\cite{eligibilitytraces}. The eligibility trace is dependent on the state $s$ and is multiplied by $\lambda$, often depreciated (i.e. $0<\lambda<1$).

Eligibility traces are shown to significantly speed up learning for frequently visited states~\cite{etracespeedup}. When performing the value update after experiencing the transition of state $s$ to its next state $s'$, the state receives an augmented reward relative to its eligibility, defined as

\begin{align}
\label{eq:et}
	e(s;\lambda) = \left\{
	\begin{array}{lr}
	\gamma\lambda e_{t-1}(\hat{s}) & \text{if } s\neq\hat{s}\\
	\gamma\lambda e_{t-1}(\hat{s})+1, & \text{if } s=\hat{s}
	\end{array}
	\right.
\end{align} 

Watkins' Q-learning~\cite{watkins} is a well-known off-policy TD-learning technique, in which the accumulated value is additionally predicated on the execution of an action. Q-learning is considered \textit{off-policy} as it allows an agent to adopt any $a$, not necessarily the one prescribed by a particular strategy, when calculating value. Equation~\ref{eq:qlearn} describes the Q-learning update process.

\begin{align}
\label{eq:qlearn}
Q(s,a;\alpha)\gets Q(s,a)+\alpha(r(s)+\gamma \max_{a'}Q(s',a')-Q(s,a))
\end{align}

Q-learning differs from traditional TD-learning in two significant ways. First, the use of state-action pairs allows the flexibility to explore strategies not prescribed by the policy. Second, note that the Q-value for the next state includes a $\max$ operator for selecting the action, which serves as an estimation of the best achievable future reward, not an estimate of the policy value. In this sense, Q-learning learns action values relative to the greedy policy and can learn as the policy changes.

SARSA~\cite{sarsa} is an on-policy RL technique, as in canonical TD-learning, but also predicates values on the state-action tuple, like Q-learning. Consequently, SARSA learns the action-values relative specifically to the policy being executed. SARSA updates similarly to Q-learning, but does not select a greedy next action, instead just selecting the action the policy prescribes at state $s$.

\begin{align}
\label{eq:sarsa}
Q(s,a;\alpha)\gets Q(s,a)+\alpha(r(s)+\gamma Q(s',a')-Q(s,a))
\end{align}

Both Q-learning and SARSA can utilize eligibility traces by a simple application of $e(s;\lambda)$ as in Eqs.~\ref{eq:td} and~\ref{eq:et}. Note that the inclusion of the action in eligibility traces is unnecessary, but possible.

\section{PALO: PAC Learning for Sequential Strategies}
\label{sec:pac}

Probably approximately correct (PAC) learning is a human-inspired technique for probabilistically bounding the errors that may arise in the acquisition of information~\cite{pac}. These bounds are generated \textit{a la} Hoeffding's inequality, a method to estimate sample average deviation, $\bar{x}$, from an unknown population mean, $\mu$. This requires user-specified parameters bounding the probabilistic error of the sample average, $\delta\in(0,1)$, and the numerical distance of the sample average to the mean, $\epsilon>0$.

\begin{align}
Pr(|\bar{x}-\mu|>\epsilon)\leq 2\cdot exp\{-2k\left(\frac{\epsilon}{\Lambda}\right)^2\}
\end{align}

Here, $\Lambda$ is defined as the minimum and maximum value possible when sampling $\mu$ and $k$ is the sample count of $\bar{x}$. The significant trait of Hoeffding's inequality is the ability to bound the error without explicit knowledge of the population mean. By bounding the expression with $\delta$, $k$ can be resolved to the lower bound for the required sample satisfying the requirements of $\epsilon$ and $\delta$.

Probably Approximately Locally Optimal (PALO), is a reinforcement learning technique that leverages PAC bounds to hill-climb through a series of \textit{performance elements} (problem solvers), $\Theta$~\cite{PALO}. In each stage of PALO, a set of neighboring performance elements (defined as an element that differs only in a single decision from the original element) are generated and sampled according to the associated PAC bounds. The goal of PALO is to transform an initial performance element until it cannot find a transformation that increases the expected value. To do this, PALO samples each performance element $k$ times, derived from the inequality as follows.

\begin{align}
\label{eq:palok}
k=\left\lceil 2\left(\frac{\Lambda[\Theta_m]}{\epsilon}\right)^2\ln\frac{2|\mathcal{T}[\Theta_m]|}{\delta_m} \right\rceil
\end{align}

where $m$ is the count of transformations that have been performed, $\Lambda[\Theta_m]$ is defined as the maximum difference between $\Theta_m$ and any of its neighbors, defined later in this section, $\mathcal{T}$ is the neighborhood of the element, and $\delta_m$ is the current $\delta$, calculated as 

\begin{align}
\delta_m=\frac{6\delta}{\pi^2 m^2}
\end{align}

PALO updates the sampled average value difference of each neighbor element $\Theta'$ against the current selected element with the equation

\begin{align}
\Delta(\Theta_m,\Theta',i)\gets\Delta(\Theta_m,\Theta',i-1)+[c(\Theta',q_i)-c(\Theta_m,q_i)]
\end{align}

where $c[\Theta,q]$ is the empirical value of executing performance element $\Theta$ for query $q$. PALO climbs to the neighbor if the difference satisfies the approximate value bound $\epsilon$.

\begin{align}
\Delta(\Theta_m,\Theta',i)/i > \epsilon(m,i,k)
\end{align}

This expression allows PALO to terminate before the sample count $i$ reaches the sample bound $k$. While there is more variance in the empirically derived expected value, $\epsilon$ is larger for smaller $i$, shrinking to $\epsilon$ as $i\rightarrow k$.

\begin{align}
	\epsilon(m,i,k) = \left\{
	\begin{array}{lr}
	\Lambda(\Theta_m,\Theta')\sqrt{\frac{1}{2i}{\ln}\frac{2(k-1)|\mathcal{T}(\Theta_m)|}{\delta_m}} & \text{if } i < k\\
	\frac{\epsilon}{2} & \text{if } i=k \\
	+\infty & \text{otherwise}
	\end{array}
	\right.
\end{align}

$\Lambda(\Theta_m,\Theta')$ is the range between the maximum and minimum differences between an element and its neighbor. The maximum distance is defined as $\Lambda[\Theta_m]$, used in Eq.~\ref{eq:palok}. PALO terminates when the current element dominates all its neighbors by $1-\epsilon(m,i,k)$. Utilizing these derived bounds, PALO provides PAC statistical guarantees formalized in the following theorem, adapted from Thm.~1 of \cite{PALO}.

\begin{thm}
\label{thm:palo}
PALO incrementally produces a series of performance elements $\Theta_1, \Theta_2, ...\Theta_m$ such that, for every element in the series $\Theta_{j-1}, \Theta_j$, and with probability $1-\delta$:
\begin{enumerate}
\item $\Theta_j$ dominates $\Theta_{j-1}$ in expected value, and
\item $\Theta_m$ is $\epsilon$-locally optimal; that is, no performance element $\Theta'\in\mathcal{T}(\Theta_m)$ dominates $\Theta-m$ and $\Theta_m$ is within $\epsilon$ of a local optima
\end{enumerate}
\end{thm}

\section{Monte Carlo Exploring Starts for POMDPs}
\label{sec:mces}

Perkins' Monte Carlo Exploring Starts for POMDPs (\mcesp{}) implements model-free reinforcement learning via Q-learning as a mechanism for solving MDPs online~\cite{perkins}. \mcesp{} differs from Q-learning for MDPs in that, instead of evaluating the action-values empirically sampled in states or for observation histories~\cite{thrun}, the algorithm evaluates the policy itself. Leveraging a variation of Q-learning called \textit{exploring starts}~\cite{rl}, \mcesp{} evaluates the action-values of a randomly selected observation history in the policy. To arrive at an optimal solution to the POMDP, the algorithm locally explores the neighborhood of a policy at each stage. If the empirically calculated expected value of a neighbor dominates the original policy, \mcesp{} transforms the policy to the neighbor. If no better neighbor can be located for the given policy, \mcesp{} terminates. Algorithm~\ref{alg:MCESP} formalizes the \mcesp{} process as proposed in~\cite{perkins}.

\begin{algorithm}[!ht]
  \small
  \caption{\mcesp{}}
  \label{alg:MCESP}
  \begin{algorithmic}[1]
    \REQUIRE Q-value table initialized; initial policy $\pi$ that is greedy w.r.t. Q-values; learning rate schedule $\alpha$; horizon $T$; sample count bound $k$
    \STATE $c_{o,a}\gets 0$ for all $o$ and $a$
    \STATE $m\gets 0$
    \REPEAT
    \STATE Choose some $o$ and $a\in A(o)$
    \STATE Generate a trajectory, $\tau$, according to $\pi\gets (o,a)$.
    \STATE $Q_{o,a}\gets(1-\alpha(m,c_{o,a}))Q_{o,a}+\alpha(m,c_{o,a})R_{post-o}(\tau)$
    \STATE $c_{o,a}\gets c_{o,a}+1$
    \IF {$\max_{a'} Q_{o,a'}-\epsilon(m,c_{o,a'},c_{o,\pi(o)},k)>Q_{o,\pi(o)}$}
    \STATE $\pi(o)\gets a'\in \arg\max_{a'}Q_{o,a'}-\epsilon(m,c_{o,a'},c_{o,\pi(o)},k)$
    \STATE $m\gets m+1$
    \STATE $c_{o'',a''}\gets 0$ for all $o''$ and $a''$
    \ENDIF
    \UNTIL{termination}
  \end{algorithmic}
\end{algorithm}

In the simplest approach MCESP-SAA (or MCESP using Sample Average Approximation), $\alpha(n,i)=\frac{1}{i+1}$, $\epsilon(n,i,j)=+\infty$ if $i=j<k$, and $k$ is arbitrarily selected. This method iteratively examines every observation and action in a round robin fashion. The expressions in line 5 are defined as the following.

\begin{defn}
\label{def:tau}
Let the trajectory $\tau$ be a tuple representing a horizon $T$ sample of the environment, containing the histories of observations received, actions taken, and rewards obtained when executing a policy, $\tau=\langle a^0,r^0,o^1,a^1,r^1,...o^{T-1},a^{T-1},r^{T-1}\rangle$.
\end{defn}

\begin{defn}
\label{def:transform}
Let $\pi \gets (o, a)$ denote a policy $\pi$ in which the action in the policy on receiving the observation $o$ is replaced with $a$. As such, 
$\pi \gets (o, a)$ is a policy in the neighborhood of $\pi$.
\end{defn}

Note that the Q-values are updated with the value present in $R_{post-o}(\tau)$, which is the rewards obtained in the trajectory $\tau$ after observing the observation $o$. Similarly, the function $R_{pre-o}(\tau)$ represents the reward in $\tau$ that precedes $o$. The value of the policy $\pi$ can therefore be decomposed as the following expression.

\begin{align*}
V^{\pi} &= E^{\pi}\{R(\pi)\} \\
&= E^{\pi}\{R_{pre-o}(\tau)\}+E^{\pi}\{R_{post-o}(\tau)\}
\end{align*}

Recall that the goal of reinforcement learning is to arrive at an optima by searching for solutions that maximize expected value. Though \mcesp{} only utilizes the rewards following the observation, since the policies are only transformed on actions at the selected observation (that is, $Q^{\pi}_{o,a}=E^{\pi\gets(o,a)}\{R_{post-o}(\tau)\}$), this property is still maintained.

\begin{align}
\label{eq:expectedreward}
\begin{split}
&V^{\pi}+\epsilon\geq V^{\pi\gets(o,a)} \\
\Longleftrightarrow & E^{\pi}\{R_{pre-o}(\tau)\}+E^{\pi}\{R_{post-o}(\tau)\} + \epsilon \geq E^{\pi\gets(o,a)}\{R_{pre-o}(\tau)\}+E^{\pi\gets(o,a)}\{R_{post-o}(\tau)\} \\
\Longleftrightarrow &E^{\pi}\{R_{post-o}(\tau)\} + \epsilon \geq E^{\pi\gets(o,a)}\{R_{post-o}(\tau)\} \Longleftrightarrow Q^{\pi}_{o,\pi(o)}+\epsilon\geq Q^{\pi}_{o,a}
\end{split}
\end{align}

\mcesp{}'s success rate at finding optima is highly dependent on the accuracy of the empirically derived expected action-values. With low $k$, the sample average has high variance, and \mcesp{} may therefore erroneously transform or terminate. If each observation is sampled with high $k$, \mcesp{} will transform to policies with strictly monotonically increasing value and terminate at a local optima.

\mcesp{} with PAC guarantees (hereafter \mcesppac{}) introduces PAC bounds to provide guarantees for this property, reproducing the parameters in PALO using $\alpha(m,i)=\frac{1}{i+1}$, $\Lambda(\pi)$ as the range of rewards generated by $R_{post-o}$ for any $o$ when executing $\pi$, $\delta_m=\frac{6\delta}{m^2\pi^2}$, and the following sample count and comparison definitions.

\begin{align}
\label{eq:mcesppac}
k_m=\left\lceil 2\left(\frac{\Lambda(\pi)}{\epsilon}\right)^2\ln\frac{2N}{\delta_m} \right\rceil
\end{align}

where $N=|\mathcal{A}_i|\cdot\frac{|\Omega_i|^T-1}{|\Omega_i|-1}-1$ is the number of neighboring transformed policies. $\Lambda$ is defined following the comparison bound definition.

\begin{align}
\label{eq:mcespeps}
	\epsilon(m,i,j,k) = \left\{
	\begin{array}{lr}
	\Lambda(\pi,\pi')\sqrt{\frac{1}{2i}{\ln}\frac{2(k_m-1)N}{\delta_m}} & \text{if } i=j < k\\
	\frac{\epsilon}{2} & \text{if } i=j=k_m \\
	+\infty & \text{otherwise}
	\end{array}
	\right.
\end{align}

Recall that $\Lambda$ in PALO bounds the maximum distance between the minimum and maximum samples taken from a policy and its neighbor. The \mcesp{} version, replacing the sample average with Q-values, is defined below.

\begin{align}
\Lambda(\pi_i,\pi'_i) & \triangleq \max_{\tau} ~(Q_{\pi_i} - Q_{\pi'_i}) - \min_{\tau} ~(Q_{\pi_i} - Q_{\pi'_i})\nonumber\\
&\leq \sum \limits_{t =0}^{T-1} \left \{(R_{i,max} - R_{i,min}) - (R_{i,min} - R_{i,max})\right \}\nonumber\\
&= \sum \limits_{t=0}^{T-1} 2(R_{i,max}- R_{i,min})
= 2T(R_{i,max}- R_{i,min})
\label{eqn:Lambda_mcesp}
\end{align}

$\Lambda(\pi)$ is merely defined as $\max_{\pi'}\Lambda(\pi,\pi')$ where $\pi'\in\mathcal{N}(\pi)$, or a neighbor of $\pi$. Reproducing PALO's guarantees, \mcesppac{} therefore has the following theoretical property.

\begin{thm}
\label{thm:mcesppac}
\mcesppac{} incrementally produces a series of policies $\pi_1,\pi_2,...\pi_m$ and with probability $1-\delta$, $\pi_j$ dominates $\pi_{j-1}$ for all $0<j\leq m$ in expected value, and $\pi_m$ is $\epsilon$-locally optimal
\end{thm}

\section{Imaging in Precision Agriculture}

The availability of high-dimensional multispectral image data of crop fields in the last few years~\cite{landsat} has dramatically increased the development of computational systems designed to analyze and interpret crop image data. In parallel, NDVI was established as a powerful metric for image data, as it computes visual attributes of crop fields while eliminating non-vegetative properties~\cite{ndvi}.

Using temporally-evolving image processing via NDVI imaging in crops is a nascent and quickly growing field~\cite{landsat_methods}. Contemporary work largely focuses on retrospective curve-fitting, as in time series analysis on Advanced Very-High-Resolution Radiometer (AVHRR)~\cite{avhrr} and Moderate Resolution Imaging Spectroradiometer (MODIS)\cite{modis} data, with several focusing on the root-mean-squared deviation (RMSD) metric over pairwise  pixel differences as an image comparison methodology~\cite{landsat}.

As I explore online environments, and therefore must make immediate estimates of possibly early or ongoing crop stress, retrospective models do not satisfy my needs. Therefore, I instead focus on adapting these methods to online settings, exploring methodologies using pairwise image differences.

\newpage
\chapter{Related Work}
\label{chap:relatedwork}

\section{Learning in Multiagent Systems}

Several methods tackle the task of optimal planning in partially observable environments both from the single and multiagent perspectives. For the former, POMDPs are concerned with arriving at optimal policies predicated on observations of hidden states given explicit knowledge of environment mechanics~\cite{aibook}. For the latter, frameworks include the MPOMDP~\cite{mpomdp} and decentralized POMDP (Dec-POMDP)~\cite{decpomdp} both of which adopt the joint planning perspective in cooperative settings, and the interactive POMDP (I-POMDP)~\cite{doshijair05} that adopts a subjective perspective to cooperative and non-cooperative settings.

A strong majority of the focus on MPOMDPs is concerned with optimal planning assuming knowledge of environment mechanics. However, a recent effort is concerned with the task of learning the underlying mechanics prior to planning as a form of model-based RL. Bayes-Adaptive POMDPs (BA-POMDPs)~\cite{bapomdp} generates approximate observation and transition functions via online sampling, which is extended to the MPOMDP setting in \cite{bampomdp}. Purely model-free learning in the cooperative multiagent setting is, to the best of our knowledge, currently unexplored.

Monte Carlo Q-Alternating~\cite{mcqalt} is a quasi model-based RL approach that uses online Q-learning in a cooperative multiagent environment where the policy of the other agent is held fixed. While no model is known a priori, Monte Carlo Q-Alternating contains an intermediate step of estimating model parameters and leveraging them for planning. After converging, the first agent fixes its policy and the other agent learns. \mcesmp{} similarly requires shared knowledge and communication, but allows simultaneous learning by all agents while maintaining convergence requirements and is purely model-free. Additionally, the instantiation \mcesmppac{} provides statistical guarantees of local optimality.

Factored-value partially observable Monte Carlo planning~\cite{amato} provides a  scalable approach to solving MPOMDPs by factoring the value function to exploit the structure of multiagent systems but requires either explicit knowledge of the environment models or a reasonably close transition model learned using the BA-MPOMDP application of factored-value Monte Carlo planning. $\epsilon$-optimality is established for the former case, but not the latter.

A  Bayes-Adaptive  I-POMDP   (BA-IPOMDP)~\cite{baipomdp}  maintains  a
vector  of latent  models  of environment  mechanics  and updates  its
belief over these models online.  In contrast to learning policies for
all agents, \mcesip{} focuses on  learning the policy of an individual
self-interested agent that shares  its environment with cooperative or
noncooperative  agents. It  does not  require explicit  models of  the
environment, which are potentially infinite in the case of BA-IPOMDPs.
Effectively, BA-IPOMDP casts model-free learning as planning over an infinite space. Along similar lines, Hoang and Low~\cite{Hoang13:General}  show how a flat Dirichlet Multinomial distribution may be utilized  to represent the posterior in interactive Bayes-optimal RL by an agent interacting with other self-interested agents. Differing from our context, the state is assumed to be perfectly observable.

\section{Models of Human Learning}

 Reinforcement learning has received much attention as a computational
 technique for modeling human play in strategic games. Several approaches
 to applying default reinforcement learning have been explored in order
 to explain phenomena in human decision making, such as eligibility
 traces to generalize reward stimuli in economic games~\cite{tdlambda}.
 Additionally, the state space has been generalized in grid-world problems
 to simulate value association between like states~\cite{leffler}.
 Several extensions to default reinforcement learning
 have been used to explain phenomena in human decision
 making, such as neural spikes~\cite{tdlambda} and principled grid-
 world generalizations~\cite{leffler}.

 Behavioral game theoretic extensions have been applied to single-shot
 and repeated games~\cite{roth95,roth98}. This differs  from  our context  of
 repeated   sequential  games   with   a  dynamic   state.   Erev   and
 Roth~\cite{roth95}  applied  reinforcement  learning to  public  good,
 market and ultimatum games. In this application, cognitive biases such
 as  the  spill  over  of  attraction to  neighboring  strategies  were
 modeled.   In follow  up  work, Erev  and Roth~\cite{roth98}  expanded
 their  analysis  by   demonstrating  their  descriptive  reinforcement
 learning  model's  predictive  capabilities  against a  large  set  of
 available  games.    As  in  their  previous  work,   the  games  were
 single-shot and simple in design.
 
\section{Image Processing and Automated Systems for Precision Agriculture}

Machine learning in agricultural domains falls under the body of work characterized by the category of precision agriculture, tackled by a variety of fields, including agriculture, agronomics, computer science, robotics, engineering, and physics. In particular, the relevant subtopics I explore include disease detection, nutrient deficiency, and insufficient water potential. This data provides a basis for precision agro-management, such as through spot spraying, targeted water irrigation and nitrogen application.

The most recent advance in precision agriculture is the FarmBeats initiative driven by Microsoft AI~\cite{farmbeats}, in which a variety of network-accessible sensor modalities, including soil water potential sensors and AUAVs, are arranged to provide automated and targeted water intervention. This methodology is powerful for tackling stresses due to underwatering, but is incapable of detecting the presence of pathogens, pests and nutrient deficiency, which express themselves phenotypically. 

Concerning the goal of disease identification and intervention, the wide array of contemporary efforts leverage phenotypic expressions of stress largely via thermal detection~\cite{khanal} and are often specific to the expression from a specific disease~\cite{plant_detection}. What remains is a generalized model that encompasses the variety of stresses in a model-free way. That is, instead of seeking a particular expression, learn the correlation between erroneous growth patterns, leaf and fruit necrosis or chlorosis, leaf spots, leaf striations and wilting (as caused by stress) and the available image and environmental data. 

\newpage
\chapter{Normative Reinforcement Learning in Multiagent Settings}
\label{chap:rlmas}

In this chapter, I propose several templates for extending Perkins' \mcesp{} in multiagent settings. I begin with a brief redefinition of \mcesp{} to accommodate the solution space (hereafter referred to as policy space), which maps observation sequences, including private observations of opponent behavior, to actions, as opposed to single observations as in its original formulation. I introduce two novel settings: \mcesmp{}, which reflects the team MPOMDP setting, and \mcesip{}, solving problems in the I-POMDP setting. 

In these sequential multiagent settings, the Q-value are updated with $R_{post-\vec{o}}(\tau)$, or the reward following the observation \textit{sequence} $\vec{o}$ in trajectory $\tau$. As $V^{\pi}=E^{\pi}\{R_{pre-\vec{o}}(\tau)\}+E^{\pi}\{R_{post-\vec{o}}(\tau)\}$ in the sequential setting, the comparison in Eq.~\ref{eq:expectedreward} holds. I redefine trajectory $\tau$ for the multiagent setting and transformation function $\pi\gets(\vec{o}_i,a)$ similarly to Defs.~\ref{def:tau} and~\ref{def:transform}.

It is important to note that the maximum observation sequence length of these policies is bound by horizon $T$. That is, the policies are designed only for problems that last up to $T$ rounds.

\begin{defn}
\label{def:tau2}
Let $\tau$ $=$ $(a^0, r^0, o^1, a^1, r^1,$ $\ldots,o^T, a^T, r^T)$ denote the trajectory of all agents, where $o^t$ denotes the joint observations $o=\langle \bar{o}_1,\bar{o}_2,...\bar{o}_Z \rangle$ at time step $t$, $a^t$ be the profile of their actions, and $r$ be the reward. Trajectory $\tau$ is composed of the individual agent trajectories $\tau_i$ $=$ $(a_i^0, r_i^0, \bar{o}_i^1, a_i^1, r_i^1, \ldots, \bar{o}_i^T, a_i^T, r_i^T)$, $i = 1, \ldots, Z$ where $\bar{o}_i^t, a_i^t, r_i^t$ are agent $i$'s observation, action and reward in joints $o^t, a^t, r^t$, respectively. $\bar{o}$ may be a single private observation, a single public observation (received by all agents), or a combination of the two, depending on the context.
\end{defn}

Consequently, $R_{post-\vec{o}_i}(\tau)$ is the cumulative reward following $\vec{o}_i$ in the $\tau_i$ component of $\tau$. 

\begin{defn}
\label{def:transform2}
Let $\pi_i \gets (\vec{o}_i, a)$ denote agent $i$'s policy $\pi_i$ in which the action in the policy on receiving the observation history $\vec{o}_i$ is replaced with $a$. As such, 
$\pi_i \gets (\vec{o}, a)$ is a policy in the neighborhood of $\pi_i$.
\end{defn}

This chapter defines the templates of the three multiagent extensions of \mcesp{}. In Ch.~\ref{chap:instantiate}, I cover their instantiations, including a policy search space optimization technique and the theoretical contributions of PAC, as well as experimental results.

\section{MCES-P: The Single Agent Perspective}
\label{sec:mcesp}

Algorithm~\ref{alg:MCESP2} redefines \mcesp{} in the sequential, multiagent domain. In both \mcesp{} and \mcesip{}, the observations received at each time step of the sequence is a tuple of both a public observation of the physical state of the environment and private signal of the opponents action, $\bar{o}_i=\{o,o_i\}$. The subject agent receives only their individual portion of the trajectory $\tau_i\in\tau$. Additionally, for both \mcesp{} and \mcesip{}, I assume the opponent enacts a deterministic policy or a mixed policy stochastically selected from two or more deterministic policies.

\begin{algorithm}[!h]
  \caption{\mcesp{} in the multiagent setting}
  \label{alg:MCESP2}
  \begin{algorithmic}[1]
    \REQUIRE Q-value table initialized; initial policy $\pi_i$ that is greedy w.r.t. Q-values; learning rate schedule $\alpha$; horizon $T$; sample count bound $k$
    \STATE $c_{\vec{o}_i, a_i} \gets 0$ for all $\vec{o}_i$ and $a_i$ 
    \STATE $m \gets 0$
    \REPEAT
    \STATE Pick some observation history $\vec{o}_i$ and action $a_i$
    \STATE Modify $\pi_i$ to $\pi_i \gets (\vec{o}_i, a_i)$
    \STATE Generate trajectory $\tau$ of length $T$ according to $\pi_i \gets (\vec{o}_i, a_i)$ ~~\small (this involves simulating the implicit policies of other agents as well)
    \STATE $Q_{\pi_i \gets \vec{o}_i, a_i} \gets (1-\alpha(m, c_{\vec{o}_i, a_i}))~Q_{\pi_i \gets \vec{o}_i, a_i} +\alpha(m, c_{\vec{o}_i,a_i})~ R_{post-\vec{o}_i}(\tau)$
    \STATE $c_{\vec{o}_i, a_i} \gets c_{\vec{o}_i, a_i} + 1$
    \IF {$\max_{{a}_i'} Q_{\pi_i \gets \vec{o}_i, a'_i} - Q_{\pi_i} > \epsilon(m, c_{\vec{o}_i,a_i}, c_{\vec{o}_i,\pi_i(\vec{o}_i)})$}
    \STATE $\pi_i(\vec{o}_i) \gets a'_i$ where $a'_i$ $\in$ arg $\max Q_{\pi_i \gets \vec{o}_i, a'_i}$ 
    \STATE $m \gets m + 1$
    \FORALL {$\vec{o}_i,  a_i $}
    \STATE $c_{\vec{o}_i, a} \gets 0$
    \ENDFOR
    \ENDIF
    \UNTIL{termination}
  \end{algorithmic}
\label{alg:mcesp}
\end{algorithm}

The algorithm begins with an initialized Q-table and count vectors over each observation sequence, which contains a sequence of public and private observations, and actions (lines 1-2). \mcesp{} then randomly selects an observation sequence and samples the environment using the initial policy and the action-transformed policy on that observation sequence (lines 4-6). If, after $k$ samples, one of these neighbors dominates the current policy, \mcesp{} transforms to the new policy, updates the transformation count, and begins sampling again on a new sequence (lines 9-13). If no neighbor dominates this policy, \mcesp{} terminates.

\mcesp{} in the multiagent setting therefore proceeds identically to the canonical case in Perkins' work, except where the policies are mappings from observation sequences to actions, and the sequence contains tuples of public and private observations. The significant changes to \mcesp{} appear in the theoretical contributions of introducing PAC, presented in Sec.~\ref{sec:guarantees}. Additionally, the changes to \mcesp{} are not the primary contributions of my work, and serve only as a baseline comparison to the template \mcesip{}.

\section{MCES-MP: The Team Setting}
\label{sec:mcesmp}

As noted in Sec.~\ref{sec:rl}, very few methodologies exist for performing RL in multiagent settings. As MPOMDPs are a generalization of POMDPs, where observations and actions are replaced by their joint, extending \mcesp{} is a very attractive proposal. In the MPOMDP, each agent receives a noisy private observation of the environment, $\bar{o}=o_i$, without observations of the opponent action. These observations are communicated instantaneously to a centralized controller that proposes policy transformations to each agent, which are executed in unison. The controller then evaluates the empirical Q-value for each agent and decides whether the transformations should be selected. Algorithm~\ref{alg:MCESMP} defines the \mcesmp{} process.

\begin{algorithm}
  \caption{\mcesmp{}}
  \label{alg:MCESMP}
  \begin{algorithmic}[1]
    \REQUIRE Profile of agent policies $\{\pi_i\}_{i=1}^Z$, that are greedy w.r.t. initial Q-values; learning rate schedule $\alpha$; error $\epsilon$; probability $\delta$; horizon $T$; sample count bound $k$
    \vspace{0.1in}
    
    \FORALL {$i\in\mathcal{I}$}
    \FORALL {$\vec{o}_i,a_i$}
    \STATE $c^i_{\vec{o}_i,a_i} \gets 0$
    \ENDFOR
    \ENDFOR
    \STATE $m \gets 0$
    \REPEAT
    \STATE Pick some joint observation history $\vec{o}$ (=$\{\vec{o}_i\}_{i=1}^Z$), and joint action $\vec{a}$ (=$\{a_i\}_{i=1}^Z$)
    \FORALL {$i\in\mathcal{I}$}
    \STATE Construct neighboring policy $\pi_i \gets (\vec{o}_i, a_i)$
    \ENDFOR
    \STATE Generate trajectory $\tau$ (=$\{\tau_i\}_{i=1}^Z$) of length $T$ with each agent using its transformed policy $\pi_i \gets (\vec{o}_i, a_i)$
    \FORALL {$i\in\mathcal{I}$}
    \STATE $Q^i_{\vec{o}_i, a_i} \gets (1-\alpha(m, c^i_{\vec{o}_i, a_i}, c^i_{\vec{o}_i, \pi_i(\vec{o}_i)})) \cdot Q^i_{\vec{o}_i, a_i} +\alpha(m, c^i_{\vec{o}_i, a_i}, c^i_{\vec{o}_i, \pi_i(\vec{o}_i)}) \cdot R_{post-\vec{o}_i}(\tau)$
    \STATE $c^i_{\vec{o}_i, a_i} \gets c^i_{\vec{o}_i, a_i} + 1$
    \ENDFOR
    \IF {$\bigwedge\limits_{i \in \mathcal{I}} \left (\max\limits_{a_i' \in \mathcal{A}_i}   Q^i_{\vec{o}_i, a_i'}-\epsilon(n, c^i_{\vec{o}_i,a_i}, c^i_{\vec{o}_i, \pi_i(\vec{o}_i)}) > Q^i_{\vec{o}_i, \pi_i(\vec{o})}\right )$}
    \STATE $m \gets m + 1$
    \FORALL {$i\in\mathcal{I}$}
    \STATE $\pi_i(\vec{o}_i) \gets a'_i$
    \FORALL {$\vec{o}_i,a_i$}
    \STATE $c^i_{\vec{o}_i,a_i} \gets 0$
    \ENDFOR
    \ENDFOR
    \ENDIF
    \UNTIL{termination}
  \end{algorithmic}
\end{algorithm}

\mcesmp{} proceeds similarly to \mcesp{} in the multiagent setting. \mcesmp{} begins with a joint set of individual policies for all agents $i\in\mathcal{I}$, an initial Q-table for each agent, and a count vector over the number of joint trajectories the agent has received when considering an individual observation sequence and action (lines 1-3). \mcesmp{} then selects a random joint observation sequence and joint action for the subject agents (line 6), proposes transformations for each which are sampled simultaneously (lines 7-8), and evaluates whether the new policies are better for all agents (lines 9-12). If, after $k$ samples, \textit{every} agent's transformed policy dominates their initial policy, \mcesmp{} transforms (line 13). If there is no joint policy in which all agents benefit (even if some agents may benefit from it), \mcesmp{} terminates.

\mcesmp{} differs in a few notable ways from \mcesp{}. First, solutions to the MPOMDP are over joint policies and, in the canonical case, therefore joint observations mapped to joint actions. However, in our formulation, each agent receives an individual observation of the environment, thus policies in \mcesmp{} are single observations to actions without private observations of opponent actions. Second, the Q-values are predicated on individual rewards, which are factored for each agent, and therefore each agent must maintain their own Q-table. Third, the comparison determining whether a transformation is selected must be \textit{mutually beneficial}, and is thus the conjunction of each transformation satisfying the domination criteria.

\section{MCES-IP: Modeling Non-cooperative Opponents}
\label{sec:mcesip}

Interactive POMDPs (I-POMDPs)~\cite{doshijair05} define the setting in which a subject agent and one or more opponents interact simultaneously in a sequential, partially observable environment. The physical state in the I-POMDP is augmented to the interaction state, which includes not only the location of the subject agent but also the belief over the opponents location and model. The model of the opponent can be simply a deterministic mapping of observations to actions (called a \textit{subintentional model}), or a model nearly as complex as the subject agent itself, which relies on beliefs over the opponents' models of the subject themselves.

Monte Carlo Exploring Starts for I-POMDPs (\mcesip{})~\cite{mcesip} searches over the same policy space as \mcesp{}, with both public observations of the environment and private signals of opponent actions, but additional predicates empirically derived expected rewards on opponent action sequences. These sequences are derived from a calculated belief over a finite set of deterministic policies representing subintentional opponent models. First, I introduce the \mcesip{} template in Alg.~\ref{alg:mcesip} for one opponent and discuss the how beliefs are generated after.

\begin{algorithm}[!h]
  \caption{\mcesip{}}
  \begin{algorithmic}[1]
    \REQUIRE Q-value table initialized; initial policy, $\pi_i$, that is greedy w.r.t. Q-values; prior on set of models $M_j$; learning rate schedule $\alpha$; horizon $T$; sample count bound $k$
    \STATE $c_{\vec{o}_i, a_i}^{\vec{a}_j} \gets 0$
    \STATE $m \gets 0$
    \REPEAT
    \STATE Pick some observation history, $\vec{o}_i$, and $a_i$
    \STATE Modify $\pi_i$ to $\pi_i \gets (\vec{o}_i, a_i)$
    \STATE Generate trajectory $\tau$ of length $T$ according to $\pi_i \gets (\vec{o}_i, a_i)$
    \STATE Generate belief sequence $\vec{b}_i$ based on $\tau$
    \STATE Obtain most probable action sequence $\vec{a}_j$ from $\vec{b}_i$
     \STATE $Q_{\pi_i \gets \vec{o}_i, a_i}^{\vec{a}_j} \gets (1-\alpha(m, c_{\vec{o}_i, a_i}^{\vec{a}_j})) \cdot Q_{\pi_i \gets \vec{o}_i, a_i}^{\vec{a}_j} +\alpha(m, c_{\vec{o}_i,a_i}^{\vec{a}_j}) \cdot R_{post-\vec{o}_i}(\tau)$
    \STATE $c_{\vec{o_i}, a_i}^{\vec{a}_j} \gets c_{\vec{o}_i, a_i} ^{\vec{a}_j} + 1$
    \IF {$\max_{a_i'} Q_{\pi_i \gets \vec{o}_i, a_i'}^{\vec{a}_j} - \epsilon^{\vec{a}_j}(m, c_{\vec{o}_i,a_i}^{\vec{a}_j}, c_{\vec{o}_i,\pi_i(\vec{o}_i)}^{\vec{a}_j})>Q_{\pi_i}^{\vec{a}_j}$}
    \STATE $\pi_i(\vec{o}_i) \gets a_i'$ where $a_i' \in \text{arg} \max_{a_i'} Q_{\pi_i \gets \vec{o}_i, a_i'}^{\vec{a}_j}$
    \STATE $m \gets m + 1$
    \FORALL {$\vec{o}_i, a_i, \vec{a}_j$}
    \STATE $c_{\vec{o}_i, a_i}^{\vec{a}_j} \gets 0$
    \ENDFOR
    \ENDIF
    \UNTIL{termination}
  \end{algorithmic}
  \label{alg:mcesip}
\end{algorithm}

\mcesip{} begins with an initial policy as in \mcesp{}, but additionally a finite space of opponent models $M_j$, which contains the true opponent model, though which model is being used is unknown. Similarly, a random observation sequence and action are selected and a trajectory generated from the neighboring policy (lines 4-6). However, since the Q-function is predicated not only on the observation sequence and action but also the likely sequence of actions the opponent has taken, \mcesip{} must additionally reason about which models the opponent is using based on the private observations it receives.

An additional assumption of \mcesip{} is that the subject agent knows the probabilities of receiving a private observation when an opponent takes an action; that is, $Pr(o_i^{t+1}|a_i^t,a^t_j)=O_i(o_i^{t+1}|a_i^t,a_j^t)$ for all observations and actions are known. With this, a belief sequence $\vec{b}_i$ can be generated as follows. Let $M^t_j$ be the set of models of an opponent agent $j$ at time step $t$, where $m_j^t\in M_j^t$ contains the $t$-length action-observation history of $j$ and a policy $\pi_j$, thus $m_j^t\triangleq \langle h_j^t,\pi_j \rangle$. Agent $i$ can update its beliefs over the model space $M_j$ at each step of the trajectory with the following equation.

\begin{align}
\label{eqn:belief-update}
  &b_i'(m_j^{t+1}|a_i^t,o^{t+1},\omega_i^{t+1}, b_i) = \sum\limits_{m_j^t \in M_j^t} b_i(m_j^t)\sum\limits_{a_j^t\in A_j} Pr(a_j^t|m_j^t)\nonumber\\
  &\times O_i(\omega_i^{t+1}|a_i^t,a_j^t)~\delta_K(h_j^{t+1},\text{APPEND}(h_j^t,a_j^t,o^{t+1}))
\end{align}

The Kronecker delta function $\delta_K$ is 1 if the updated history of agent $j$ $h^{t+1}_j$ matches the previous history $h_j^t$ with the action $a_j^t$ and public observation $o^{t+1}$, generated by APPEND. Considering the space of possible opponent actions, the belief update computes the probability of receiving the private observation in $\tau$ at time step $t$ and propagates the probability to the previous belief of $m_j^t$. The sequence of beliefs $\vec{b}_i$ is then calculated from a prior uniform probability and updated according to Eq.~\ref{eqn:belief-update} for each time step. The action sequence in $\vec{a}_j$ is simply calculated by selecting the most probable model at each belief, $\arg\max_{m^t_j}b^t_i(m^t_j)$, and then selecting the most probable action based on the model's history and policy, $\arg\max_{a_j}Pr(a_j|m_j)$.

\mcesip{} uses the above to calculate the belief sequence (line 7) and, consequently, the most probable action sequence (line 8). Predicating the Q-value on the action sequence, observation sequence, and candidate action, \mcesip{} samples trajectories until $k$ samples of each observation sequence and action for \textit{any} action sequence is satisfied. If a better neighbor is found, the policy is transformed to the neighbor (line 11). Otherwise, the algorithm terminates.

\section{Concluding Remarks}

This chapter reintroduced \mcesp{}, a reinforcement learning approach for finding solutions to POMDPs. We extend \mcesp{} in the context of the multiagent setting. Whereas in canonical \mcesp{} the local neighborhood are policies mapping single observations to actions, in our extension \mcesp{} transforms over observation sequences. While the majority of novelty in the multiagent \mcesp{} extension arrives in its PAC extension, \mcesppac{}, the redefinition serves as an excellent departure point for the two novel applications, \mcesmp{} and \mcesip{}.

The field of reinforcement learning approaches to multiagent systems is still in the early stages. Since MPOMDPs can be seen as a very straightforward generalization of POMDPs to joint team settings, \mcesp{} is tantalizing as a inspiration. \mcesmp{} searches for solutions to an underlying MPOMDP, where a centralized controller explores a set of joint policies, whose expected rewards are generated empirically from two or more agents acting simultaneously and communicating their rewards instantaneously. By sampling joint policies and individually updating factored Q-functions, \mcesmp{} is able to explore strictly beneficial joint policies and hill climb to local optima.

I-POMDPs are significantly more complex than MPOMDPs, as agents don't communicate and, often, have conflicting goals, leading to antagonistic settings. \mcesip{} explores solutions to the I-POMDP setting, where a subject agent explicitly models an opponent given a private observation function providing the probabilities of receiving observations from opponent actions. \mcesip{} explores individual policies mapping public and private observations to actions, as in \mcesp{}, but also updates a belief over opponent models based on sampled trajectories. From this, a maximal likelihood sequence of opponent actions is derived and augment the expectation over action-observation values. Then, when each observation-action combination is sampled $k$ times for any of these action sequences, \mcesip{} either transforms to a dominant neighbor or terminates at the local optima.

\newpage
\chapter{Instantiating Multiagent MCES}
\label{chap:instantiate}

Two significant hurdles preclude the effectiveness of the algorithms presented in Ch.~\ref{chap:rlmas}: selecting the appropriate $k$ to guarantee accuracy of the sample averages and the extreme burden of sampling observation sequences that appear rarely. For the first hurdle, \mcesp{} provides an elegant principled method for providing these guarantees in its PAC extension, \mcesppac{}. In a similar approach, I prove that these theoretical bounds can be applied to the significantly more complex multiagent setting in Sec.~\ref{sec:guarantees}.

The second hurdle is due to the nature of Monte Carlo sampling, which is inherently a form of rejection sampling. For these algorithms to terminate, \textit{every} observation sequence and action must be explored up to $k$ times. However, many of these observation sequences occur very rarely. Imagine the widely-known POMDP problem domain, the Tiger problem. In this domain, the subject agent faces two doors, one with a tiger behind it and one with a pot of gold. If the agent opens the door with the tiger, they are eaten, incurring a reward of -100. However, if the agent opens the door with the gold, they gain a reward of 10. Instead of opening a door, the agent may also opt to listen to glean where the tiger is. With some noise (often a probability of 0.85), the agent hears a growl from the correct door. Therefore, it is highly unlikely that the agent would hear a growl from a different door every round. However, for \mcesp{} to terminate, it must receive $k$ samples of this trajectory!

An important observation about these sequences is that they, due to their rarity, also indicate a relatively lower impact on the expected reward of the policy. Simply, the extreme computation cost may not be worth the gain in reward from a more optimal action on that sequence. I introduce a principled method for removing these rare sequences from the policy search space in Sec.~\ref{sec:pruning}.

In Sec.~\ref{sec:mcesresults}, I introduce several multiagent problem domains of varying complexity to evaluate the effectiveness of \mcesmppac{} and \mcesippac{}, the latter of which is compared to the multiagent extension of \mcesppac{}.

\section{Statistically Guaranteeing Optimality}
\label{sec:guarantees}

In Sec.~\ref{sec:mcesp}, I introduced the PAC extension of \mcesp{}. In this algorithm, Perkins leverages probably approximately correct learning to bound the variance on the sample average collected online in an extension I refer to as \mcesppac{}. Recall that, given the user-defined parameters $\delta$ and $\epsilon$, \mcesppac{} guarantees that, with probability 1-$\delta$, each selected transformed policy is guaranteed to dominate the original policy and, when \mcesppac{} terminates, the final policy is $\epsilon$-locally optimal. However, these bounds are guaranteed for the single-agent POMDP context. To proceed to the multiagent context, I first redefine \mcesppac{}'s comparison bound and sample count bounds.

\subsection{\mcesppac{} for Multiagent Settings}

The main observation of the multiagent version of \mcesp{} is that the observation space is quite a bit larger. Where the original setting involved only receiving a single observation indicating the physical state of the environment, the subject agent additionally receives a private signal correlating the action the opponent has taken in the last round.

Formally, where $\bar{o}\in\Omega$ for the single agent context is a singleton of the public observation $\bar{o}=o$, the multiagent context adds the private observation such that $\bar{o}=\{o,o_i\}$. Recall the Tiger problem defined in Sec.~\ref{chap:instantiate}. The multiagent Tiger problem, where two or more agents simultaneously open doors, may include a private signal of the opponent actions, defined by $\Omega_i=\{Silence, Open Left, Open Right\}$, a noisy signal indicating the opponent listened, opened the left door, or opened the right door, respectively. For a horizon of $T=3$, a policy in the single agent version of the Tiger problem has a neighborhood $N=20$, whereas the multiagent version has a neighborhood of $N=129$!

While rather straightforward, this expansion of the neighborhood is sufficient for calculating the bounds for the more complex multiagent setting and is proved in Sec.~\ref{sec:mcesppacmas_proof}. For clarity and continuity, Eq.~\ref{eq:mcesppac} and~\ref{eq:mcespeps} are repeated below.

\begin{align*}
k_m=\left\lceil 2\left(\frac{\Lambda(\pi)}{\epsilon}\right)^2\ln\frac{2N}{\delta_m} \right\rceil
\end{align*}

\begin{align*}
	\epsilon(m,p,q,k_m) = \left\{
	\begin{array}{lr}
	\Lambda(\pi,\pi')\sqrt{\frac{1}{2p}{\ln}\frac{2(k_m-1)N}{\delta_m}} & \text{if } p=q < k_m\\
	\frac{\epsilon}{2} & \text{if } p=q=k_m \\
	+\infty & \text{otherwise}
	\end{array}
	\right.
\end{align*}

Including the expanded observation sequence, \mcesppac{} provides statistical guarantees of optimality to the transformed policies it generates as specified in Thm.~\ref{thm:mcesppacmas}.

\begin{thm}
\label{thm:mcesppacmas}
\mcesppac{} in the multiagent setting incrementally produces a series of policies $\pi_1,\pi_2,...\pi_m$ and with probability $1-\delta$, $\pi_j$ dominates $\pi_{j-1}$ for all $0<j\leq m$ in expected value, and $\pi_m$ is $\epsilon$-locally optimal
\end{thm}

\subsection{\mcesmppac{}}

Recall that MPOMDPs expand the frame for states $\mathcal{S}$, actions $\mathcal{A}$, and observations $\Omega$ to their joints over all agents. Each agent, however, operates in the truly single agent context, and communicate their observations and rewards to a centralized controller. This controller then iterates over joint policies $\vec{\pi}=\{\pi_1,\pi_2,...\pi_Z\}, Z=|\mathcal{I}|$ to hill-climb to an $\epsilon$-locally optimal joint policy.

The introduction of individual sample average approximations $Q_i$ significantly complicates the calculation of PAC bounds. Essentially, since the controller iterates over several policies, the errors that may occur when sampling and transforming due to these empirical estimations are multiplicatively greater to the order of agents in the environment. Consequently, the agent space $Z$ impacts the comparison bound and sample count bound in the following fashion.

\begin{align*}
	\epsilon(m,p,q,k_m) = \left\{
	\begin{array}{lr}
	\Lambda(\pi_i,\pi'_i)\sqrt{\frac{1}{\sqrt{2p}}ln\frac{\sqrt[2Z]{(4Z-2)(k_m-1)}N}{\sqrt[2Z]{\delta_m}}} & \text{if } p=q < k_m\\
	\frac{\epsilon}{2} & \text{if } p=q=k_m \\
	+\infty & \text{otherwise}
	\end{array}
	\right.
\end{align*}

where

\begin{align*}
k_m = \left\lceil 2\left(\frac{\Lambda(\pi_i)}{\epsilon}\right)^2 ln\left(\frac{\sqrt[2Z]{4Z-2}N}{\sqrt[2Z]{\delta_m}}\right)\right\rceil
\end{align*}

and $\delta_m$ is defined as in \mcesppac{}. Since the comparison method is a conjunct of a comparison between each individual agent's policy against a transformation, the neighborhood is bound by the largest local neighborhood for any agent, such that

\begin{align*}
N\leq |A_i|^Z\left(\frac{|\Omega_i|^Z-1}{|\Omega_i|-1}-1\right)
\end{align*}

\mcesmppac{} then terminates when, after $k_m$ samples, there is no joint observation sequence $\vec{o}$ and joint action $\vec{a}$ that proposes a better neighbor for every agent $i\in\mathcal{I}$, or, prior to $k_m$ samples

\begin{align*}
Q^i_{\vec{o}_i,a_i}< Q^i_{\vec{o}_i,\pi_i(\vec{o}_i)}+\epsilon-\epsilon(m,c^i_{\vec{o}_i,a_i},c^i_{\vec{o}_i,\pi_i(\vec{o}_i)},k_m)
\end{align*}

for all agents, every individual sequence $\vec{o}_i\in\vec{o}$, and individual action $a_i\in\vec{a}$. With these augmented bounds, \mcesppac{} is able to provide similar guarantees to \mcesppac{} as in Thm.~\ref{thm:mcesmppac}.

\begin{thm}
\label{thm:mcesmppac}
\mcesmppac{} incrementally produces a series of joint policies $\vec{\pi}_1,\vec{\pi}_2,...\vec{\pi}_m$ and with probability $1-\delta$, $\vec{\pi}_j$ dominates every agents policy in $\vec{\pi}_{j-1}$ for all $0<j\leq m$ in expected value, and $\vec{\pi}_m$ is $\epsilon$-locally optimal, where there exists no joint policy $\vec{\pi}\in neighbor(\vec{\pi}_m)$ where all agents' transformed policies are better than their counterpart in $\vec{\pi}_m$.
\end{thm}

This theorem is further proven in Sec.~\ref{sec:mcesmppac_proof}. It is in the context of two agents for the interest of brevity, but holds for any number of agents.

\subsection{\mcesippac{}}

While \mcesippac{} is set in the same context as \mcesppac{}, it becomes noticeably more computationally expensive by performing a belief update. However, maintaining an expectation over the actions an opponent has taken in a trajectory has a profound theoretical implication: the subject agent can refine their knowledge about the minimum and maximum rewards achievable by that observation sequence. To clarify this point, I first introduce the derived comparison and sample count bounds for \mcesippac{}. For a given error $\epsilon$ and probability $\delta$ let,
	\begin{align}
	\label{eq:mcesipeps}
	\epsilon^{\vec{a}_j}(m,p,q,k_m) = \left\{
	\begin{array}{lr}
	\Lambda^{\vec{a}_j}(\pi_i,\pi'_i)\sqrt{\frac{1}{2p}{\ln}\frac{2(k_m-1)N}{\delta_m}} & \text{if } p=q < k_m\\
	\frac{\epsilon}{2} & \text{if } p = q \geq k_m \\
	+\infty & \text{otherwise}
	\end{array}
	\right.
	\end{align}
where 
\begin{align*}
  k_m = \left\lceil 2\frac{(\Lambda^{\vec{a}_j}(\pi_i,\pi'_i))^2}{\epsilon^2} \ln\frac{2N}{\delta_m}\right\rceil
\end{align*}

Note that $\Lambda$ has additionally been predicated by the action sequence of opponent $j$, $\vec{a}_j$. Here,  $\Lambda^{\vec{a}_j}(\pi_i,\pi'_i)$ is  an upper  bound on  the range of  the difference in  action-values between two  policies given $j$'s   action  sequence is  $\vec{a}_j$.   Let   $R_{i,max}^{a_j}$  $=$
$\max_{s,a_i} R_i(s,a_i,a_j)$  and analogously  for $R_{i,min}^{a_j}$;
these specific values are assumed to be known. Following the form of Eq.~\ref{eqn:Lambda_mcesp}:
\begin{align}
\Lambda^{\vec{a}_j}(\pi_i,\pi'_i) & = \max_{\tau} ~\left(Q^{\vec{a}_j}_{\pi_i} - Q^{\vec{a}_j}_{\pi'_i}\right) - \min_{\tau} ~\left(Q^{\vec{a}_j}_{\pi_i} - Q^{\vec{a}_j}_{\pi'_i}\right)\nonumber\\
&\leq \sum \limits_{t \in T} \left(R_{i,max}^{a_j^t} - R_{i,min}^{a_j^t}\right) - \left(R_{i,min}^{a_j^t} - R_{i,max}^{a_j^t}\right)\nonumber\\
&= \sum \limits_{t \in T} 2\left(R_{i,max}^{a_j^t} - R_{i,min}^{a_j^t}\right)
\label{eqn:Lambda_mcesip}
\end{align}

This observation leads to the crucial observation contributing to the significance of the \mcesippac{} extension. In spite of the fact that performing the belief update incurs a significant computational cost, the following proposition holds.

\begin{prop}[Reduced sample complexity]
	For any predicted action sequence, $\vec{a}_j$, $$\Lambda^{\vec{a}_j}(\pi_i,\pi'_i) \leq \Lambda(\pi_i,\pi'_i)$$
\label{prop:sample_complexity}
\vspace{-0.5cm}
\end{prop} 

The proof for Prop.~\ref{prop:sample_complexity} appears in the Appendix in Sec.~\ref{sec:samplecomplexity_proof}. Simply, Prop.~\ref{prop:sample_complexity} states that the sample count bound $k_m$ for \mcesippac{} is bound by that of \mcesppac{} and, in fact, is often less depending on the structure of the reward function. Specifically, the greater impact that the opponent action has on the range of rewards available to the subject agent, the more significant the gap becomes. This property is rather significant, as $k_m$ grows quadratically with regards to $\Lambda$. However, as the Q-table grows to the order of the size of the opponent action space $|A_j|^T$, the sample bound must be reduced by $\frac{|\Omega|^T-1}{|\Omega|-1}$ to justify the inclusion of $\vec{a}_j$. In our experiments, this was certainly demonstrated.

If private signals provide perfect information about $j$'s actions, then 
\mcesippac{} terminates when,
$$ Q_{\pi_i \gets \vec{o}_i, a_i'}^{\vec{a}_j} < Q_{\pi_i}^{\vec{a}_j} + \epsilon - \epsilon^{\vec{a}_j}(m, c_{\vec{o}_i,a_i}^{\vec{a}_j}, c_{\vec{o}_i,\pi_i(\vec{o}_i)}^{\vec{a}_j})$$
for  all  $\vec{o}_i$,  $a_i'  \neq  \pi_i(\vec{o}_i)$,  and  \mcesippac{} has
encountered  at most $\frac{|\Omega|^T   -  1}{|\Omega|-1}$   many   distinct
$\vec{a}_j$  in the  trajectories for  each $\vec{o}_i$,  $a_i'$ pair.
Under perfect monitoring, for  the comparison
threshold,  sample  bound  and   probability  as  defined  above,  the
following theorem obtains for \mcesippac{}.

\begin{thm}
\label{thm:mcesippac}
\mcesippac{} incrementally produces a series of policies $\pi_1,\pi_2,...\pi_m$ and with probability $1-\delta$, $\pi_j$ dominates $\pi_{j-1}$ for all $0<j\leq m$ in expected value, and $\pi_m$ is $\epsilon$-locally optimal
\end{thm} 

The  proof of  this  theorem  proceeds analogously  to  the proof  for
Theorem~\ref{thm:mcesppacmas} where $\Lambda$ is replaced with $\Lambda^{\vec{a}_j}$.

I generalize the above results  on \mcesippac{} for the  case of
{\em  imperfect  monitoring}  --  when the  probability  of  error  in
estimating $\vec{a}_j$  is known, say  $\delta_e$.  This unique error may arise when the Q-value for $\vec{a}_j$, $Q^{\vec{a}_j}$, is placed in the wrong bin (see line 9 of Algorithm~\ref{alg:mcesip}), leading to non-i.i.d. samples for that Q-value.     Fortunately,    given    $\delta_e$,
Theorem~\ref{thm:mcesippac} can be generalized   to   the   case   of   independent   but
non-identically  distributed  samples using  a  more  general form  of
Hoeffding's  inequality.  Analogously  to Eq.~\ref{eqn:Lambda_mcesip},
let  $\bar{\Lambda}^{\vec{a}_j}$ be  an upper  bound on  the range  of
differences in action-values  for all $j$'s action  sequences that are
{\em different} from $\vec{a}_j$.  Then  for the case of $p=q<k_m$, I
redefine $\epsilon^{\vec{a}_j}(m,p,q,k_m)$ as,
\small
\begin{align*}
\epsilon^{\vec{a}_j}(m,p,q,k_m)=\sqrt{(1-\delta_e)(\Lambda^{\vec{a}_j})^2+\delta_e(\bar{\Lambda}^{\vec{a}_j})^2}\sqrt{\frac{1}{2p}\ln\frac{2(k_m-1)N}{\delta_m}}
\end{align*}
\normalsize
where $p=q<k_m$ and identically to Eq.~\ref{eq:mcesipeps} otherwise, and $k_m$ is redefined as
$$k_m=\left\lceil\frac{2((1-\delta_e)(\Lambda^{\vec{a}_j})^2+\delta_e(\bar{\Lambda}^{\vec{a}_j})^2)}{\epsilon^2}\ln\frac{2N}{\delta_m}\right\rceil$$

Algorithm~\ref{alg:mcesip}  requires  a  slight modification  for  this  case.   For
convenience,                                                       let
$\zeta^{\vec{a}_j}=\max_{a_i'} Q_{\pi_i       \gets       \vec{o}_i,
  a_i'}^{\vec{a}_j}              -             Q_{\pi_i}^{\vec{a}_j}$.
Then, line 11 of Algorithm 2 changes to the following:
$$(1-\delta_e)\zeta^{\vec{a}_j}+\delta_e
  \bar{\zeta}^{{\vec{a}_j}'}           >
  (1-\delta_e)\epsilon^{\vec{a}_j}+\delta_e
  \bar{\epsilon}^{{\vec{a}_j}'}$$

  The implicit assumption above  is that when $Q^{\vec{a}_j}$ receives
  a wrong sample meant for  bin ${\vec{a}_j'}$, the action sequence is
  equally likely  to be any ${\vec{a}_j'}  \neq \vec{a}_j$. Therefore,
  $\bar{\zeta}^{\vec{a}_j'}$ is  the mean of  $\zeta^{\vec{a}_j'}$ for
  all  $\vec{a}_j'  \neq  \vec{a}_j$  seen  so  far,  and  analogously
  $\bar{\epsilon}^{{\vec{a}_j}'}$  is  also  the  mean.   Notice  that
  insisting on  the test of line  11 for every $\vec{a}_j$  before the
  current policy  is changed, would be  a stronger form of  this test,
  and hence also  sufficient. Finally, 
  note that when $\delta_e=0$,  \mcesippac{} is recovered for perfect
  monitoring as a special case of this setting.

\section{Avoiding Rare Observation Sequences}
\label{sec:pruning}

Recall that the templates for \mcesppac{}, \mcesmppac{}, and \mcesippac{} all rely on exploring up to $k_m$ samples of every observation sequence-action pair up to a horizon of $T$. In these algorithms, a form of Monte Carlo method called \textit{rejection sampling} is used, whereby only the $\tau$ that contain the target observation sequence are used, and if a representative sample is not generated, it is discarded. Unfortunately, many observation sequences may very rarely occur, as noted in the introduction of this chapter, requiring significantly more trajectories be run than $k_m$. Regardless, for the MCES algorithms to terminate, $k_m$ samples of even these sequences are required. An obvious approach would be to remove these rare sequences from consideration entirely, a method already used in planning for multiagent systems~\cite{epruning,banerjeepruning}, though this may further prevent the algorithm from finding an optima. In this section, I introduce a simple method for pruning the policy search space and provide the statistical bounds of optimality for searching in this pruned space.

As noted, the act of removing rare sequences has the beneficial effect of avoiding expensive computation, but removes neighboring policies from consideration that may have higher expected reward, introducing \textit{regret}. Luckily, a less likely observation sequence is unlikely to add a significant portion of that reward. To bound the maximum regret introduced by pruning, I introduce a user-defined parameter $\phi$ that serves as an upper-bound on the proportion of maximum reward that may be limited as regret.

In the context of the MCES algorithms, avoiding $\vec{o}_i$ for transformation means forgoing the largest $R_{post-\vec{o}_i}$ rewards, upper bounded by $\max R_{post-\vec{o}_i}-\min R_{post-\vec{o}_i}$. Consequently, regret in the context of multiagent systems for an individual agent is bounded by

\begin{align}
\label{eq:regret}
  regret_{\vec{o}_i} & \leq Pr(\vec{o}_i;\vec{\pi})\left ( \max_{\tau} R_{post-\vec{o}_i}(\tau) - \min_{\tau} R_{post-\vec{o}_i}(\tau) \right ) \nonumber\\
                     & = Pr(\vec{o}_i;\vec{\pi})~(T - len(\vec{o}_i))~\left ( R_{i,max} - R_{i,min} \right )
\end{align}

where $\vec{\pi}=\{\pi_1,\pi_2,...\pi_Z\}$, the joint of all agent policies, and $Pr(\vec{o}_i;\vec{\pi})$ is the likelihood of agent $i$ observing sequence $\vec{o}_i$ when all agents execute the actions prescribed by their policies. I  normalize the regret to be a proportion of total reward as

\begin{align*}
\bar{regret}_{\vec{o}_i}=\frac{regret_{\vec{o}_i}}{T(R_{i,max}-R_{i,min})}
\end{align*}

Unfortunately, the policies of other agents in $\vec{\pi}$ are unknown for \mcesp{} and \mcesip{}. Even if they were, $Pr(\vec{o}_i;\vec{\pi})$ is unknown, as the algorithms are model-free, excluding \mcesmp{} from being able to calculate the regret, as well. However, I propose a methodology for computing an approximate calculation for $Pr(\vec{o}_i)$ from experience on-line.

Where $\mathcal{P}$ is the set of observation sequences pruned from the search space, and $\phi$ is the regret bound provided by the user, \mcesp{} can obtain $\mathcal{P}$ in the following way: Calculate the regret for each sequence $\vec{o}_i$ and sort in ascending order. Add each sequence to $\mathcal{P}$ until the next addition would exceed $\phi$, guaranteeing $\sum\limits_{\vec{o}_i\in\mathcal{P}}\bar{regret}_{\vec{o}_i}\leq\phi$. Obviously, increasing the regret bound allows for more sequences to be pruned. An alternative method for pruning is to merely select a random observation sequence and, if the proportion of regret it introduces when combined with the current set $\mathcal{P}$ doesn't exceed $\phi$, prune it.

\begin{algorithm}[!ht]
  \caption{Multiagent \mcesp{}-Prune}
  \label{alg:prune}
  \begin{algorithmic}[1]
    \REQUIRE Q-value table initialized; initial policy $\pi_i$ that is greedy w.r.t. Q-values; learning rate schedule $\alpha$; horizon $T$; sample count bound $k$; regret bound $\phi$
    \STATE $c_{\vec{o}_i, a_i} \gets 0$ for all $\vec{o}_i$ and $a_i$ 
    \STATE $m \gets 0$
    \STATE $\mathcal{P}\gets\emptyset$
    \STATE $c^{\phi}_{\vec{o}_i}\gets 0$
    \REPEAT
    \STATE Pick some observation history $\vec{o}_i$ and action $a_i$
    \IF {$\vec{o}_i\in\mathcal{P}$}
    \STATE Go to 6
    \ENDIF
    \STATE Modify $\pi_i$ to $\pi_i \gets (\vec{o}_i, a_i)$
    \STATE Generate trajectory $\tau$ of length $T$ according to $\pi_i \gets (\vec{o}_i, a_i)$ ~~\small (this involves simulating the implicit policies of other agents as well)
    \STATE $Q_{\pi_i \gets \vec{o}_i, a_i} \gets (1-\alpha(m, c_{\vec{o}_i, a_i}))~Q_{\pi_i \gets \vec{o}_i, a_i} +\alpha(m, c_{\vec{o}_i,a_i})~ R_{post-\vec{o}_i}(\tau)$
    \STATE $c_{\vec{o}_i, a_i} \gets c_{\vec{o}_i, a_i} + 1$
    \STATE $c^{\phi}_{\vec{o}_i}\gets c^{\phi}_{\vec{o}_i}+1$
    \IF {$\max_{{a}_i'} Q_{\pi_i \gets \vec{o}_i, a'_i} - Q_{\pi_i} > \epsilon(m, c_{\vec{o}_i,a_i}, c_{\vec{o}_i,\pi_i(\vec{o}_i)})$}
    \STATE $\pi_i(\vec{o}_i) \gets a'_i$ where $a'_i$ $\in$ arg $\max Q_{\pi_i \gets \vec{o}_i, a'_i}$ 
    \STATE $m \gets m + 1$
    \FORALL {$\vec{o}_i,  a_i $}
    \STATE $c_{\vec{o}_i, a} \gets 0$
    \ENDFOR
    \ENDIF
    \IF {$\sum\limits_{\vec{o}'_i\in\mathcal{P}\bigcup\vec{o}_i}\bar{regret}_{\vec{o}'_i}\leq\phi+\rho(\sum_{\vec{o}'_i}c^{\phi}_{\vec{o}'_i})$}
    \STATE $\mathcal{P}\gets\mathcal{P}\bigcup\vec{o}_i$
    \ENDIF
    \UNTIL{termination}
  \end{algorithmic}
\end{algorithm}

Algorithm~\ref{alg:prune} describes the random addition process for \mcesp{}, though it is trivially extended to \mcesmp{} and \mcesip{}, which share the same general design. Two important distinctions can be observed in the augmented algorithm. Pruning will skip any observation sequence that is contained in the pruning set (initially empty) on lines 7-8. Lines 19-20 add the last observed observation sequence $\vec{o}_i$ into the pruned set $\mathcal{P}$ if the cumulative normalized regret for all pruned sequences including $\vec{o}_i$ remains less than or equal to the bound $\phi+\rho(\sum_{\vec{o}'_i}c^{\phi}_{\vec{o}'_i})$, where

\begin{align*}
\rho(i) = \left\{
	\begin{array}{lr}
	0 & \text{if } i\geq k\\
	+\infty & \text{otherwise}
	\end{array}
	\right.
\end{align*}

$\mathcal{P}$ consequently remains empty unless a required amount of samples are collected to best reflect the true observation sequence probabilities. In the experiments, this sample bound is set to $\frac{k_m}{2}$.

\section{Experiments and Results}
\label{sec:mcesresults}

In this section, I demonstrate the effectiveness of the three multiagent RL techniques with PAC guarantees and search space pruning in solving several sequential, partially observable domains. First, I prove that \mcesippac{} shows remarkable improvement over the sample requirements in \mcesppac{} when solving three non-cooperative domains: the multiagent Tiger problem, a novel autonomous Unmanned Aerial Vehicle (AUAV) predator-prey domain, and a simplified but high-dimension version of the Money Laundering problem. Second, I show that \mcesmppac{} converges to good $\epsilon$-local joint policies for two team problems: the cooperative multiagent Tiger problem, with horizons $T=3$ and $T=4$, and a very large Firefighting problem, with both $3$ and $4$ agents interacting simultaneously.

\subsection{\mcesppac{} and \mcesippac{}}
\label{sec:pipresults}

\begin{figure*}[!ht]
\centering
	\begin{minipage}{1.3in}
	\centerline{\includegraphics[height=1.5in]{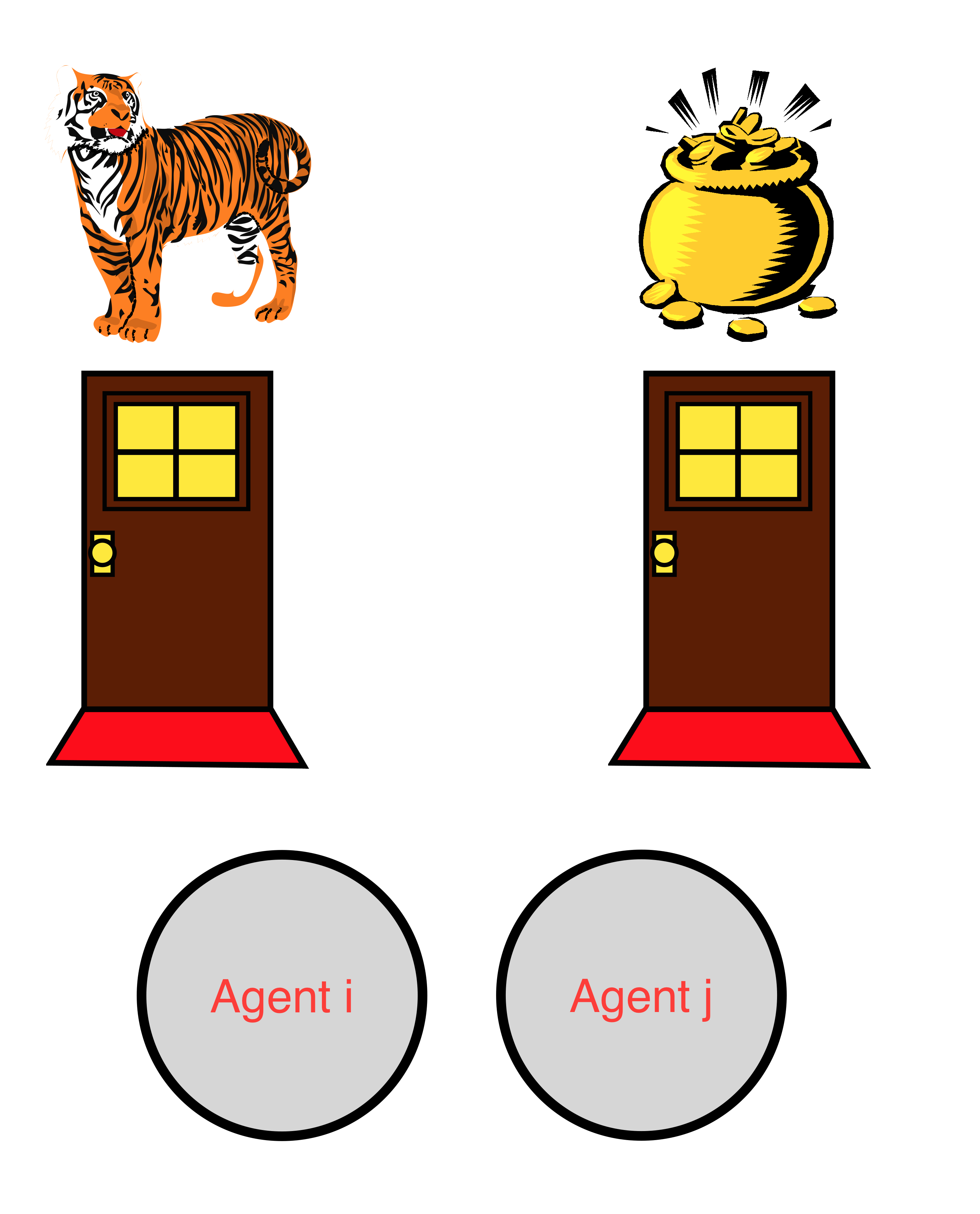}}
	\centerline{$(a)$ Multiagent tiger}
	\end{minipage}
	\hspace{-0.65cm}
	\begin{minipage}{2.7in}
	\centerline{\includegraphics[height=1.5in]{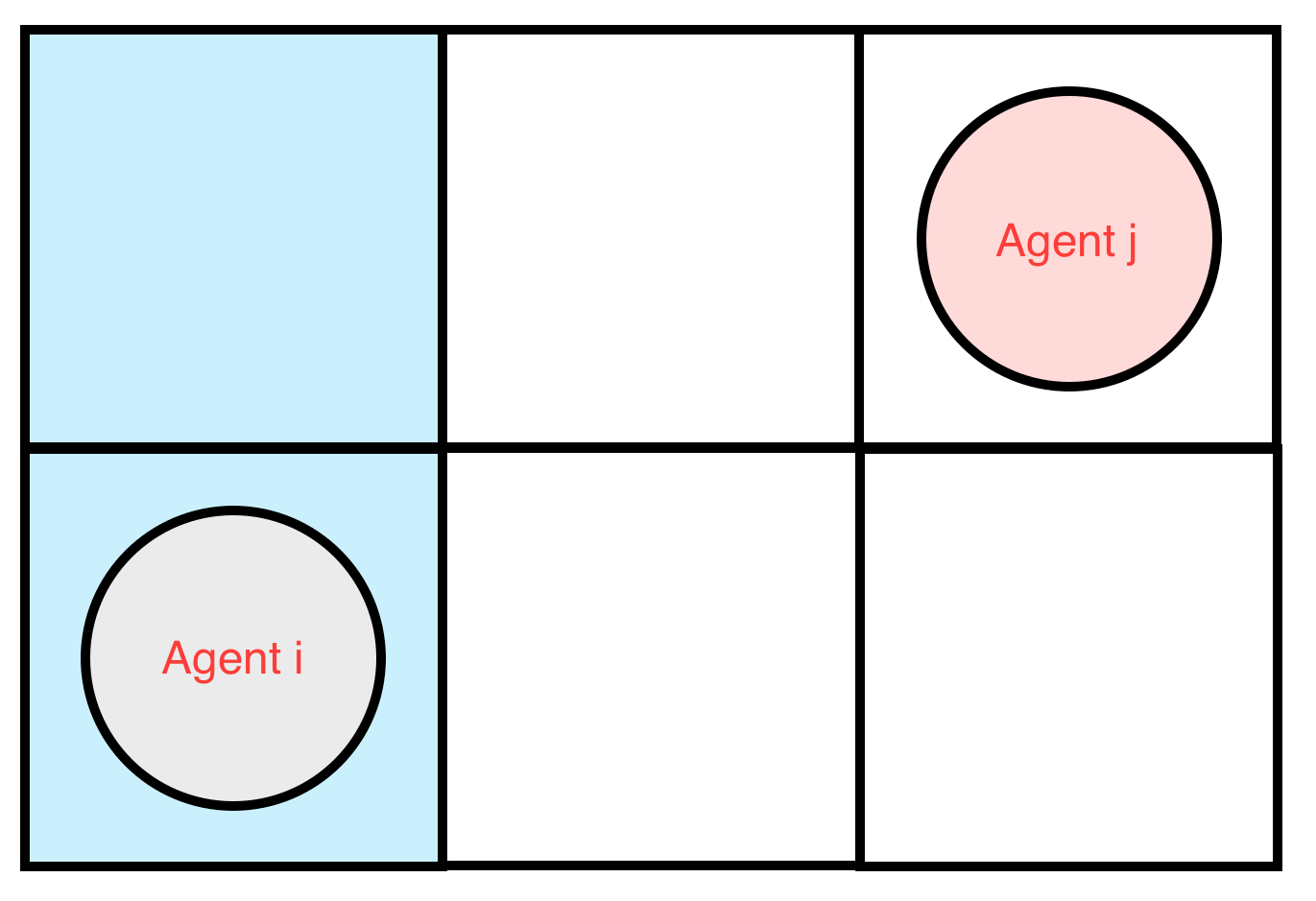}}
	\centerline{$(b)$ 3$\times$2 \textsf{AUAV}}
	\end{minipage}
	\hspace{-0.65cm}
	\begin{minipage}{2.7in}
	\centerline{\includegraphics[height=1.5in]{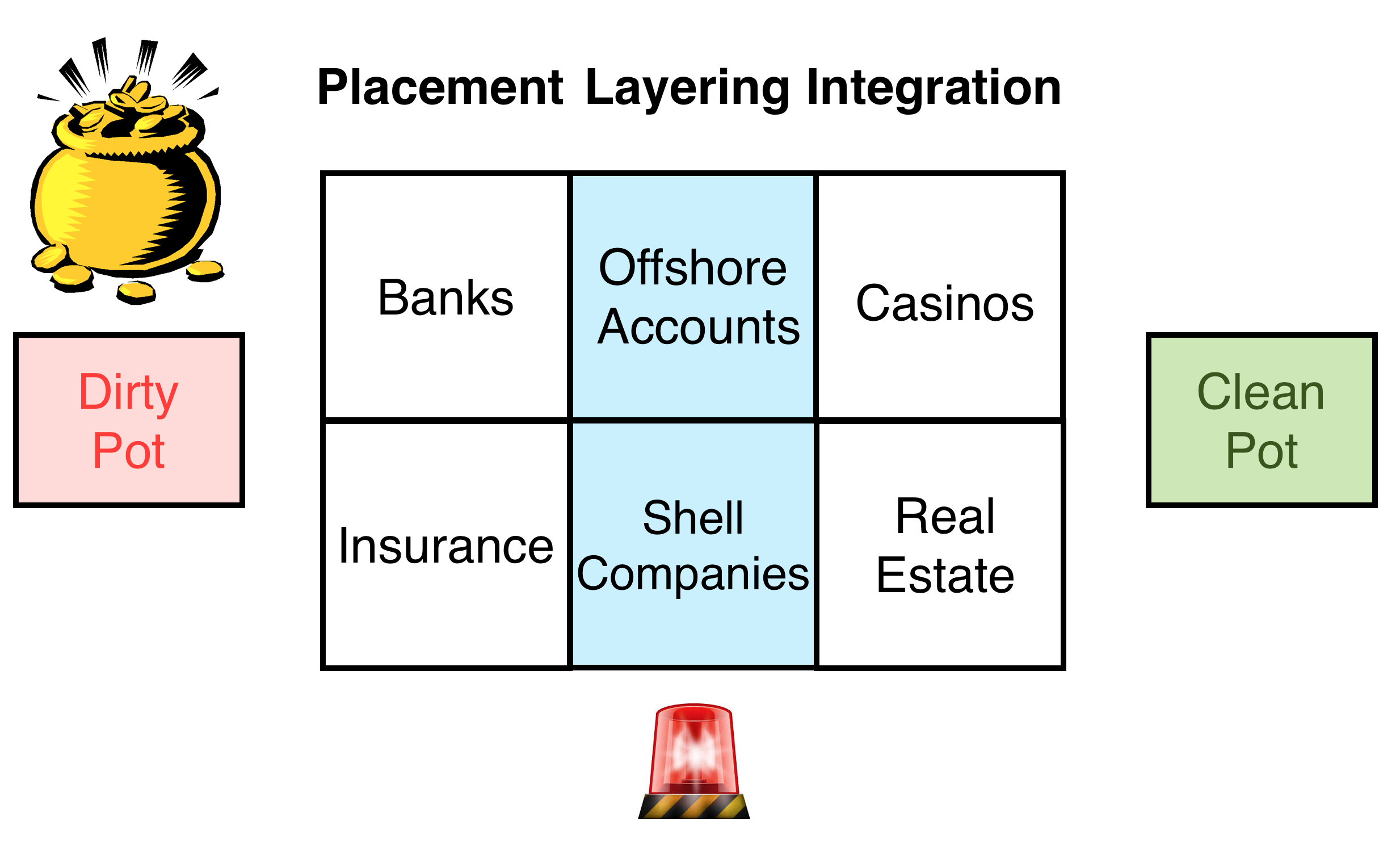}}
	\centerline{$(c)$ Money Laundering}
	\end{minipage}
	\caption{\small Problem domains for \mcesppac{} and \mcesippac{}}
	\label{fig:pipdomains}
\end{figure*}

Chapter~\ref{chap:instantiate} introduced the single Tiger problem to demonstrate the burden of rare observation sequences. The multiagent Tiger problem (Fig.~\ref{fig:pipdomains}.a) strongly parallels this domain. Two states define the space of physical configurations: the Tiger behind the left door or the right door. Each agent may take one of three actions: listen, open the left door, or open the right door. When listening, the subject agent receives a correct observation with probability 0.85 as to which door the tiger is behind, and hears the wrong door otherwise. Additionally, the subject agent receives a private signal of the action the opponent has taken with probability 0.6, or a random signal indicating the other actions uniformly otherwise. When a door is opened, the tiger randomly moves to another door. Opening the door with the gold grants $10$ points, while being eaten by the tiger costs $-100$ points. Listening results in a reward of $-1$. In addition, to predicate the reward on opponent actions, each agent loses half of what the opponent gains, leading to a maximum of $60$ points and a minimum of $-105$.

The 3$\times$2 \textsf{AUAV} problem, inspired by ~\cite{auavproblem}, is a predator-prey game in which the subject agent (predator) attempts to catch the opponent (prey) in a 3$\times$2 grid before they arrive at the goal sector, illustrated in Fig.~\ref{fig:pipdomains}.b. The subject agent (agent $i$) begins play in the bottom-left sector, while the opponent $j$ begins in the top-right sector. The subject may move up, left, or right, while the opponent can move down, left, or right. Each round, both agents receive a public observation indicating whether or not both agents are in the same row, the same column, the same sector, or none of these. The subject receives a private signal indicating which direction the prey is moving, useful when the public observation doesn't have information as to the relative location. The probabilities of correct observations are the same as in multiagent Tiger. If the subject catches the prey, they gain a reward of $100$, but receive $-100$ if the subject makes it to the left column, and the subject cannot catch the prey in these sectors.

The last domain, Money Laundering (ML) problem~\cite{mldomain} comprises a setting where a blue team (subject agent) seeks to confiscate illicit money that the  opponent red team is laundering. The red team can move money from the initial state to a series of placement states (banks and insurance), to layering states (offshore accounts and shell companies), to integration states (casinos and real estate), and to the safe clean pot. The blue team may place a sensor at each of these locations or confiscate the illicit funds. Each agent receives a noisy public observation indicating whether the money and sensor are in the same location, in the same laundering state, or if neither are the case. The blue team also receives a noisy observation of the last action of the red team. The blue team receives $10$ points for catching the opponent and $-100$ if the opponent makes it to the clean pot or if they attempt to confiscate money in a different sector than the illicit funds reside. Figure~\ref{fig:pipdomains}.c illustrates an example start state for ML, where the blue team begins with sensors on the layering states.

Table~\ref{tab:Domain} summarizes the domain statistics and parameter settings. For all domains and both methods, each agent has a $15\%$ chance of receiving a noisy public observation and $40\%$ chance of a noisy private observation. The opponent in each game follows a single policy (stationary environment) or fixed distribution over multiple policies (nonstationary environment).

\begin{table}[!h]
  \centering
  \small
    \begin{tabular}{|c|c|}
      \hline
      \textbf{Domain} & \textbf{Specifications} \\
      \hline
      \multirow{2}[2]{*}[-0.2em]{Multiagent Tiger} & $\epsilon=0.05$, $\delta=0.1$, $\phi=0.15$, $T=3$, \\&$|\Omega|=2$, $|A_i|=3$, $|A_j|=3$, $|\Pi_j|=14$ \\
      \hline
      \multirow{2}[2]{*}[-0.2em]{3$\times$2 {\textsf{AUAV}}} & $\epsilon=0.1$, $\delta=0.1$, $\phi=0.2$, $T=3$, \\ &$|\Omega|=4$, $|A_i|=3$, $|A_j|=3$, $|\Pi_j|=4$ \\
      \hline
      \multirow{2}[2]{*}[-0.2em]{{Money Laundering}} & $\epsilon=0.1$, $\delta=0.15$, $\phi=0.2$, $T=3$, \\ &$|\Omega|=4$, $|A_i|=4$, $|A_j|=5$, $|\Pi_j|=8$ \\
      \hline
    \end{tabular}%
    \caption{\small Parameter configurations for the three problem domains.}
 \label{tab:Domain}%
\end{table}

We simulate $i$'s policies with opponent $j$ following either a single policy or a mixture of two policies. These policies are picked from a predefined set $\Pi_j$. As per the policy space specified in Table~\ref{tab:Domain}, 105 games of the multiagent Tiger problem, 9  of the 3$\times$2 \textsf{AUAV} problem, and 13 of ML comprise the data set. 


\begin{figure*}[!t]
\centering
	\begin{minipage}{2.2in}
	\centerline{\includegraphics[width=2.2in]{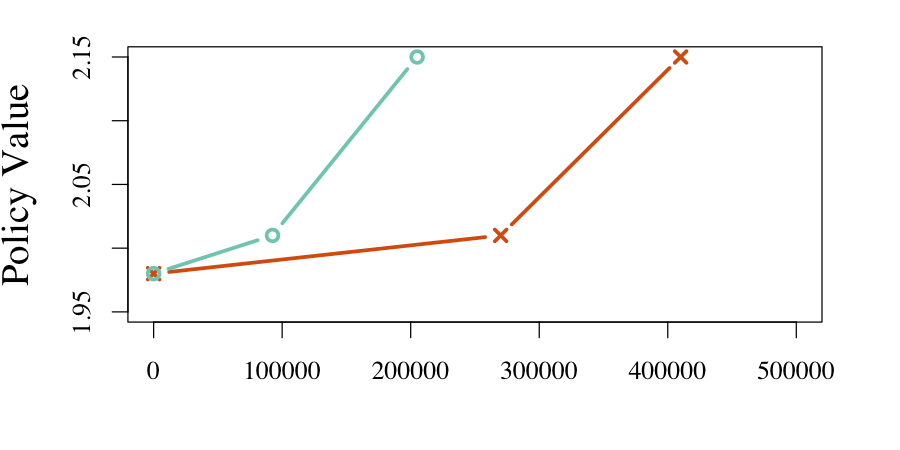}}
	\end{minipage}
	\hspace{-0.65cm}
	\begin{minipage}{2.2in}
	\centerline{\includegraphics[width=2.2in]{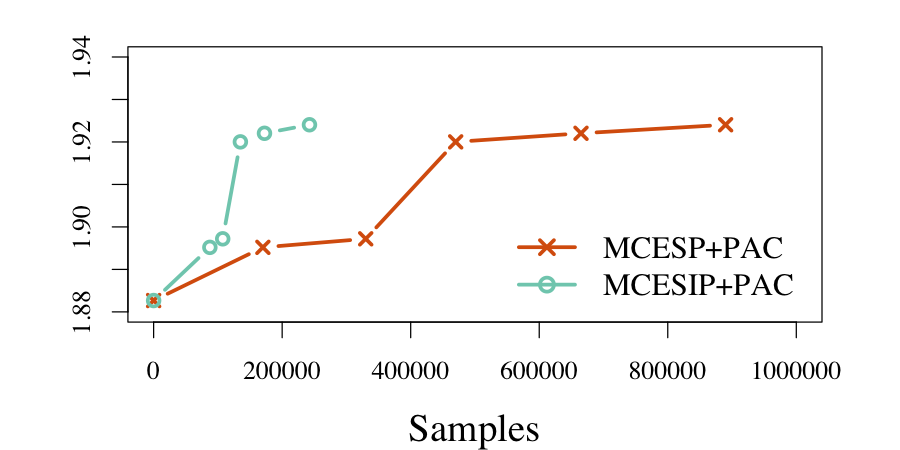}}
	\end{minipage}
	\hspace{-0.65cm}
	\begin{minipage}{2.2in}
	\centerline{\includegraphics[width=2.2in]{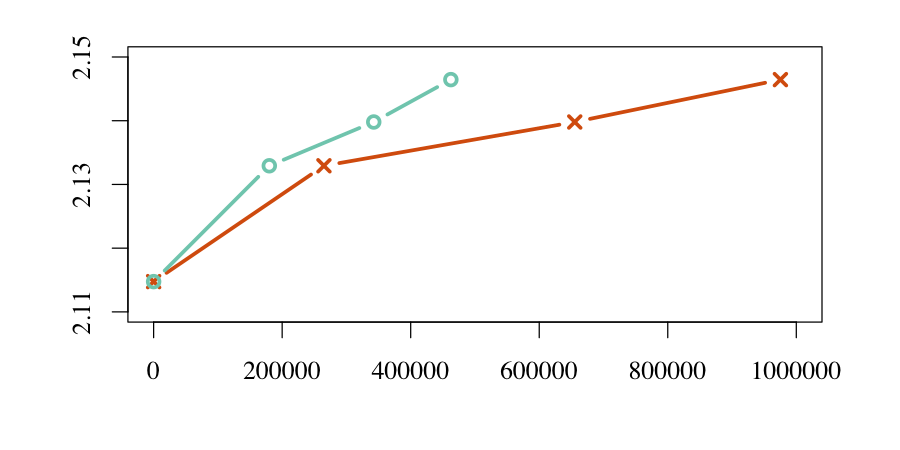}}
	\end{minipage}
	\caption{\small Example policy transformation paths with intermediate values for 3 different opponents in the multiagent Tiger problem. The right most transformation paths are for an opponent with mixed strategy.}
	\label{fig:transform}
\end{figure*}
First, I show that \mcesppac{} and \mcesippac{} demonstrates the PAC guarantee of monotonically increasing successive transformations. Figure~\ref{fig:transform} illustrates three example runs and the values for \mcesppac{} and \mcesippac{} given the same opponent. Different runs undergo varying number of transformations with some policies not transforming at all because they are $\epsilon$-locally optimal initially itself. As shown in Fig.~\ref{fig:transform}, each successive transformation results in a higher value and in no case is the final policy lower in value than the initial policy as should be expected. In every case, \mcesppac{} requires more samples to transform than \mcesippac{}.

\begin{table}[!ht]
	\centering
	\scalebox{0.76}{%
		\begin{tabular}{|c|c|c|c|c|c|c|}
			\hline 
			\textbf{Domain} & \textbf{Method} & \textbf{Policy} & \shortstack{\textbf{Mean \# of samples}\\\textbf{per transform}}& \shortstack{\textbf{Mean}\\\textbf{bound on $k_m$}}\\
			\hline \hline
			\multirow{4}{*}{Tiger} & \multirow{2}{*}{\textbf{\mcesppac{}}} & \textbf{Single} & 156,328 $\pm$ 21,012 & 265,948 $\pm$ 7,909 \\
			\cline{3-5}
			& & \textbf{Mixed} & 219,057 $\pm$ 9,521 & 263,565 $\pm$ 6,101\\
			\cline{2-5}
			& \multirow{2}{*}{\textbf{\mcesippac{}}} & \textbf{Single} & 72,740 $\pm$ 5,963 & 117,590 $\pm$ 3,309 \\
			\cline{3-5}
			& & \textbf{Mixed} & 117,504 $\pm$ 6,678 & 119,313 $\pm$ 1,089 \\
			\hline \hline
			\multirow{4}{*}{3$\times$2 AUAV} & \multirow{2}{*}{\textbf{\mcesppac{}}} & \textbf{Single} & 32,397 $\pm$ 2,816 & 91,443 $\pm$ 637 \\
			\cline{3-5}
			& & \textbf{Mixed} & 42,126 $\pm$ 1,689 & 96,328 $\pm$ 118\\
			\cline{2-5} 
			& \multirow{2}{*}{\textbf{\mcesippac{}}} & \textbf{Single} & 6,437 $\pm$ 68  & 20,397 $\pm$ 206  \\
			\cline{3-5}
			& & \textbf{Mixed} & 19,499 $\pm$ 1,304 & 22,763 $\pm$ 169 \\
			\hline \hline
			\multirow{4}{*}{ML} & \multirow{2}{*}{\textbf{\mcesppac{}}} & \textbf{Single} & 20,717 $\pm$ 2,418 & 34,726 $\pm$ 617 \\
			\cline{3-5}
			& & \textbf{Mixed} & 20,247 $\pm$ 4,974 & 35,612 $\pm$ 490\\
			\cline{2-5}
			& \multirow{2}{*}{\textbf{\mcesippac{}}} & \textbf{Single} & 1,947 $\pm$ 330 & 24,172 $\pm$ 448 \\
			\cline{3-5}
			& & \textbf{Mixed} & 3,174 $\pm$ 536 & 24,347 $\pm$ 482 \\
			\hline
		\end{tabular}
	}
	\caption{\small Mean effective sample size and theoretical bound across the stages for the three problem domains, stratified over method and whether the opponent follows a single policy or a mixed set of policies.}
	\label{tbl:avgs}
\end{table}
Second, Table~\ref{tbl:avgs} lists the mean of the theoretical sample bound $k_m$ across the different stages $m$ over all runs and the mean of the effective number of samples over all runs that were utilized by both methods. In validation of our theoretical result, \mcesippac{} requires remarkably fewer samples to transform per action sequence, around half in multiagent Tiger, about a quarter in 3$\times$2 \textsf{AUAV}, and nearly a tenth in Money Laundering. Additionally, the bound on $k_m$ is also significantly less compared to the bound for \mcesppac{}. Due to stochasticity in the simulations and finite sampling bounds, \mcesppac{} and \mcesippac{} may deviate in transformation paths. However, in over $80\%$ of the runs, both result in the same converged policy with \mcesippac{} converging on average under half the number of samples taken by \mcesppac{}.

\begin{table}[!ht]
	\centering
	\scalebox{0.79}{%
		\begin{tabular}{|c|c|c|c|c|c|}
			\hline
			\textbf{Pruning} & \textbf{Metric} & \textbf{Multiagent Tiger}& \textbf{3$\times$2 \textsf{AUAV}} & \textbf{ML} \\
			\hline \hline
			\multirow{2}{*}{\textbf{Without}} & \textbf{Neighborhood} & 128 & 470 & 636 \\
			\cline{2-5}
			& \textbf{Total $k_m$} & 15,893,387 & 3,704,396 & 18,911,460 \\
			\hline
			\multirow{2}{*}{\textbf{With}} & \textbf{Neighborhood} & 26 & 32 & 76 \\
			\cline{2-5}
			& \textbf{Total $k_m$} & 2,093,328 & 624,057 & 2,259,860 \\
			\hline
		\end{tabular}
	}
	\caption{\small Neighborhood size and total $k_m$ values for all the domains using their respective parameters in Table~\ref{tab:Domain}, with and without pruning for both \mcesppac{} and \mcesippac{}. Note that the total bound on samples reduces by almost an order of magnitude.}
	\label{tbl:prune}
\end{table}
Finally, observation sequence pruning plays a crucial role in improving the scalability of \mcesppac{} and \mcesippac{}, dramatically reducing the search space and run time while minimizing the impact on incurred regret. I list the mean of the required total $k_m$ values across all observation sequences for both problem domains in Table~\ref{tbl:prune}, both with and without pruning. 
As the policy search space for both methods is the same, the regret due to pruning the search space does not depend on the method used. Observation sequence pruning benefits both methods equally in reducing the policy search space as I demonstrate in Table~\ref{tbl:prune}.

Each neighborhood is calculated with a horizon of 3. For the multiagent Tiger problem, the size of the observation sequence space per round is 6 (2 public $\times$ 3 private observations) with 3 possible actions, resulting in a maximum neighborhood of 128. Given a regret bound of 0.15, 34 of 43 distinct observation sequences are eliminated on average, resulting in a neighborhood of 26. For 3$\times$2 \textsf{AUAV}, the observation space is 12 (4 public $\times$ 3 private observations) with 3 actions, resulting in a neighborhood of 470. On average, 146 of 157 observation sequences are pruned for a regret bound of 0.2, leaving 32 neighbors. ML's observation space is 9 (3 public $\times$ 3 private observations) with 7 actions, resulting in a neighborhood of 636. With a regret bound of 0.2, 80 of 91 sequences are pruned leaving 76 neighbors.

\subsection{\mcesmppac{}}

\begin{figure*}[!t]
\centering
	\begin{minipage}{2.1in}
	\centerline{\includegraphics[width=2.1in]{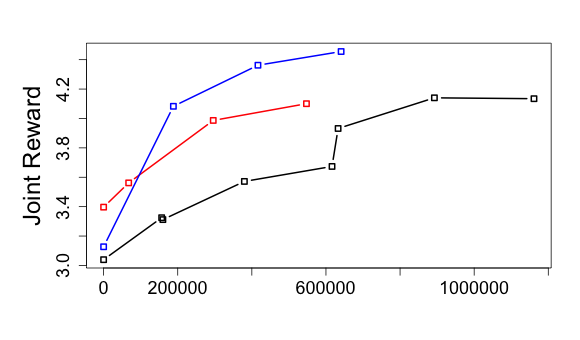}}
	\centerline{{$(a)$}}
	\end{minipage}
	\hspace{0cm}
	\begin{minipage}{2.1in}
	\centerline{\includegraphics[width=2.1in]{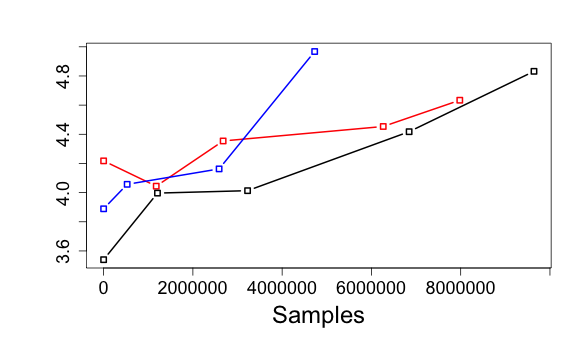}}
	\centerline{{$(b)$}}
	\end{minipage}
	\begin{minipage}{2.1in}
	\centerline{\includegraphics[width=2.1in]{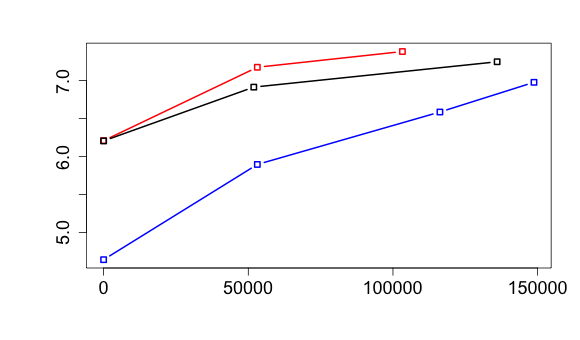}}
	\centerline{{$(c)$}}
	\end{minipage}
	\caption{\small Example policy transformation paths with true intermediate values for 3 different trials of the $(a)$ Multiagent Tiger $T=3$, $(b)$ Multiagent Tiger $T=4$, and $(c)$ Firefighting problems. Note that \mcesmppac{} transforms due to empirically sampled Q-values, which may be in error, illustrated in $(b)$ for the trial represented in red.}
	\label{fig:mptransform}
	\vspace{-0.05in}
\end{figure*}

I test \mcesmppac{}'s ability to hill-climb to $\epsilon$-local optima in two domains: the multiagent Tiger problem with horizon $T=3$ and $T=4$, as defined previously but with team-based rewards, and 3- and 4-agent versions of the 3-horizon Firefighting domain~\cite{firefighting}.

The team setting multiagent Tiger problem follows the same state, action, and observation configuration seen in Sec.~\ref{sec:pipresults}, but no private signals are received. Additionally, both agents receive the same reward. When both agents open the correct door, the team receives a reward of 20. When both open the wrong door, the reward is $-100$. Opening different doors results in a reward of $-100$, both listening results in $-2$, one agent listening while the other opens the correct door is $9$, and listening while the wrong door is opened is $-101$.

\begin{figure}[!h]
\centering
	\includegraphics[width=2.5in]{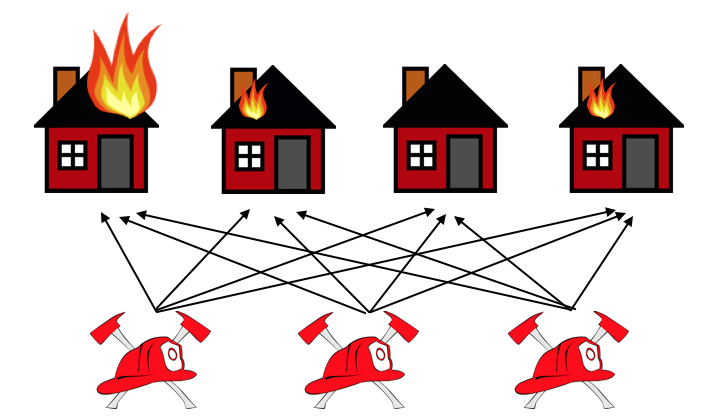}
	\caption{\small Illustration of an initial state of Firefighting problem with $Z=3$, $n_h=4$, and $n_f=3$. Each agent may move to one of the houses every round.}
	\label{fig:ff}
	\vspace{-0.05in}
\end{figure}

The Firefighting domain is a significantly larger domain where $Z$ agents are initially placed in front of $n_h$ houses, each burning with an intensity $0\leq f_h < n_f$. As an action, an agent can move to any one of the houses. An agent gets a noisy observation of a fire in their location with a $0.2$ probability if $f=0$, $0.5$ if $f=1$, and $0.8$ otherwise. When no agents are in a house, it may catch fire or increase in intensity with $0.8$ probability if a neighbor is on fire, or, if not, cannot catch fire but may increase in intensity with $0.4$ probability. One agent in a house will guaranteed lower its intensity by 1, unless a neighbor is on fire, resulting in a $0.6$ probability. Two or more agents in a house extinguishes its flames. Agents receive a negative reward relative to the order of fire intensities in each house, $r=\sum_{h\in n_h}f_h$. In our experiments, we explore the domain with $Z=3$ agents, $n_h=4$ houses, and $n_f=3$ fire intensity. To test scalability, we also text $Z=4$ agents, but with one less house, $n_h=3$. Figure~\ref{fig:ff} illustrates an example initial state of our domain for this configuration.

For each trial in both domains, we select a random pairing of private observation sequences to individual actions for each agent, representing an initial joint policy. \mcesmppac{} is tested 100 times for the multiagent Tiger problem with a horizon of 3 and the Firefighting problem. Additionally, we run 40 trials of the multiagent Tiger problem with an additional planning horizon and 60 trials of the Firefighting problem with an additional agent. Table~\ref{tab:mpDomain} lists the domain parameter specifications for these trials.

\begin{table}[!h]
	\small
  \vspace{-0.05em}
  \centering
	\scalebox{0.94}{%
    \begin{tabular}{| c | c |}
      \hline
      \textbf{Domain} & \textbf{Specifications} \\
      \hline
      \multirow{2}[2]{*}[-0.2em]{\shortstack{Multiagent Tiger\\T=3}} & $Z=2$, $|S|=2$, $\epsilon=0.05$, $\delta=0.1$, \\&$\phi=0.15$, $T=3$, $|\Omega_i|=2$, $|A_i|=3$ \\
      \hline
      \multirow{2}[2]{*}[-0.2em]{\shortstack{Multiagent Tiger\\T=4}} & $Z=2$, $|S|=2$, $\epsilon=0.1$, $\delta=0.15$, \\&$\phi=0.2$, $T=4$, $|\Omega_i|=2$, $|A_i|=3$ \\
      \hline
      \multirow{2}[2]{*}[-0.2em]{\shortstack{Firefighting\\Z=3}} & $Z=3$, $|S|=5,264$, $\epsilon=0.1$, $\delta=0.15$, \\ &$\phi=0.2$, $T=3$, $|\Omega_i|=2$, $|A_i|=3$ \\
      \hline
      \multirow{2}[2]{*}[-0.2em]{\shortstack{Firefighting\\Z=4}} & $Z=4$, $|S|=2,213$, $\epsilon=0.15$, $\delta=0.15$, \\ &$\phi=0.2$, $T=3$, $|\Omega_i|=2$, $|A_i|=3$ \\
      \hline
    \end{tabular}%
	}
    \caption{{\small Parameter configurations for the problem domains.}}
 \label{tab:mpDomain}%
	\vspace{-0.05in}
\end{table}

Table~\ref{tbl:mpavgs} lists the metrics collected for the 100 trials of the multiagent Tiger $T=3$ and Firefighting problems, and 40 trials of the multiagent Tiger $T=4$ problem. Figure~\ref{fig:mptransform} includes plots illustrating the true intermediate policy values for 3 distinct trials of multiagent Tiger for both $T=3$ and $T=4$, as well as the Firefighting problem $Z=3$ and $Z=4$. In every case except one transformation in $(b)$, \mcesmppac{} transformed to a joint policy whose true value dominated the origin joint policy.

\begin{table}[!t]
	\centering
	\scalebox{0.77}{%
		\begin{tabular}{|c|c|c|c|c|}
			\hline
			\multirow{3}{*}{\textbf{Metric}} & \multicolumn{4}{c|}{\textbf{Domain}} \\
			\cline{2-5}
			& \multicolumn{2}{c|}{\textbf{Multiagent Tiger}} & \multicolumn{2}{c|}{\textbf{Firefighting}} \\
			\cline{2-5}
			& $T=3$ & $T=4$ & $Z=3$ & $Z=4$ \\
			\hline
			\hline
			\shortstack{\textbf{Mean Initial}\\\textbf{Value}} & \shortstack{	1.36	 $\pm$	0.03	} & \shortstack{	1.81	 $\pm$	0.05	} & 2.28	 $\pm$ 	0.02 & 2.82	 $\pm$ 	0.35 \\
			\hline
			\shortstack{\textbf{Mean Converged}\\\textbf{Value}} & \shortstack{	1.93	 $\pm$	0.03	} & \shortstack{	2.51	 $\pm$0.04	} & 2.40	 $\pm$ 	0.02 & 3.01	 $\pm$ 	0.22 \\
			\hline
			\shortstack{\textbf{Mean \# of}\\\textbf{transformations}}& \shortstack{	2.74	 $\pm$ 0.16	} & \shortstack{	4.41	 $\pm$	0.29	} & 0.72	 $\pm$ 	0.07 & 0.51	 $\pm$ 	0.54 \\
			\hline
			\shortstack{\textbf{Mean \# of samples}\\\textbf{per transform}} & \shortstack{	166,973	 $\pm$	5,245	} & \shortstack{	69,411	 $\pm$	3,060	} & 44,739	 $\pm$ 	861 & 15,535	 $\pm$ 	2,450 \\
			\hline
			\shortstack{\textbf{Mean $k_m$}} & \shortstack{	202,911	 $\pm$	2,282	} & \shortstack{	95,194	 $\pm$	1,346	} & 45,297	 $\pm$ 	905 & 16,145	 $\pm$ 	284 \\
			\hline
		\end{tabular}
	}
	\caption{\small Average trial metrics for multiagent Tiger and the Firefighting problem}
	\label{tbl:mpavgs}
\end{table}

For the multiagent Tiger $T=3$ problem, about $2.7$ joint transformations were suggested per trial, resulting in a statistically significant increase in value over the initial random policy. Additionally, due to \mcesmppac{}'s guarantees for monotonically increasing intermediate values, nearly every transformation in all our trials resulted in an increased reward value for each agent. Multiagent Tiger $T=4$ demonstrates a similarly statistically significant gain in reward, though the gain is much more noticable. Additionally, $T=4$ transformed twice as often. However, the added planning horizon required over twice the computation time. Where the $T=3$ configuration took $2.28$ hours on average, $T=4$ required $5.71$ hours. In both planning horizons, \mcesmppac{} was able to converge early, with the empirical samples taken per convergence being remarkably lower (around $75\%$) compared to the maximum sample bound $k_m$, demonstrating the sample efficiency of the PAC extension.

In Firefighting $Z=3$, since agents moving to and from houses often both prevents fires from growing and, often, extinguishes them completely, many random initial policies are quite close to optima, reflected in the average transformation count of $0.72$. However, for highly suboptimal policies, such as all agents moving to the same house and remaining, the reward is increased by, on average, $1.52$ times the initial value. Even so, the average reward difference is statistically significant considering random start policies. Unfortunately, the empirical samples taken per transform was not significantly lower than the maximum requirement. This lack of significance extends to the converged values as well in the $Z=3$ domain, lending weight to the observation that the glut of agents makes it easy to extinguish fires. For $Z=4$, not only is the value not significant and the sample count not significantly lower than the max, but the amount of transformations was quite regularly 0 and usually between 0 and 1.

\begin{table}[!t]
	\centering
	\scalebox{0.75}{%
		\begin{tabular}{|c|c|c|c|c|c|c|}
			\hline
			\multirow{2}{*}{\textbf{Pruning}} & \multirow{2}{*}{\textbf{Metric}} & \multicolumn{2}{c|}{\textbf{Multiagent Tiger}} & \multicolumn{2}{c|}{\textbf{Firefighting}} \\
			\cline{3-6}
			& & $T=3$ & $T=4$ & $Z=3$ & $Z=4$\\
			\hline \hline
			\multirow{2}{*}{\textbf{Without}} & \textbf{Neighborhood} & 189 & 756 & 4,671  & 4,671\\
			\cline{2-6}
			& \textbf{Total $k_m$} & 40,865,013 & 70,994,448 & 253,358,616 & 87,975,281 \\
			\hline
			\multirow{2}{*}{\textbf{With}} & \textbf{Neighborhood} & 33 & 93 & 64  & 64 \\
			\cline{2-6}
			& \textbf{Total $k_m$} & 7,135,161 & 8,733,444 & 3,455,872 & 1,200,003 \\
			\hline
		\end{tabular}
	}
	\caption{\small Neighborhood size and total $k_m$ values for the domains using their respective parameters in Table~\ref{tab:Domain}, with and without pruning for both \mcesmp{}.}
	\label{tbl:mpprune}
	\vspace{-0.9em}
\end{table}

Policy search space pruning plays a strong role in promoting the scalability of \mcesmppac{} by limiting the neighborhood of local joint policies. This, in turn, limits the total maximum bound on $k_m$ required for transformation. Table~\ref{tbl:mpprune} illustrates the large effect of pruning the joint policy search space.

Both domains benefit from pruning the joint policy search space, resulting in a dramatically reduced bound on samples required. However, the benefit for Firefighting is immense. One key observation about this reduction is that Firefighting has an additional agent, which exponentially increases the joint policy space, but additionally has far less stochasticity in the observation function than multiagent Tiger. This, in turn, significantly increases the size of the individual pruned observation sequence sets and, as a result, reflects in the joint neighborhood size.

For the multiagent Tiger problem, each agent has an individual observation space of 2 per round and can select from 3 actions, resulting in a candidate space for 3 horizons of 21 policies and 189 joint policies. With $\phi=0.15$, on average 156 of the 189 joint policies are removed. For 4 horizons, the individual policy space is 45 with 756 joint policies. With $\phi=0.2$, on average, 93 joint policies remain after pruning. In the Firefighting problem $Z=3$, for each agent, with a per round observation space of 2 and with 4 actions possible and for 3 horizons, 28 policies and 4,672 joint policies comprise the neighborhood. With $\phi=0.2$, 64 policies remain. Note that, for $Z=4$, the neighborhood is the same size, as there is one more agent but one less house (and thus action). However, since $\epsilon$ is higher, it results in a lower bound on the total required samples.

\section{Concluding Remarks}

In this chapter, I presented instantiations for all three MCES algorithms, including a methodology guaranteeing the $\epsilon$-local optimality and monotonically increasing value in transformations as well as a scalability heuristic for pruning the policy search space. I then tested the instantiations against a variety of partially observable, sequential, multiagent settings.

\mcesippac{} dramatically reduces the sample complexity of \mcesppac{} with reductions on sample bounds and empirical sample counts ranging between $50\%$ to $75\%$ less than \mcesppac{}. In nearly every one of trials from each of the three domains, \mcesippac{} is able to achieve the same optima as \mcesppac{} despite the fraction of samples taken. By introducing sequence pruning, both instantiations benefit from requiring nearly $1/5$th the number of samples than the unpruned neighborhood. 

\mcesmp{} has the unenviable task of taking samples of extremely large joint state spaces, even for toy domains. With the inclusion of joint policy search space pruning, the \mcesmp{} template benefits from a significant reduction in run time and sample complexity at the expense of a bounded increase in regret and a relaxation of the pure model-free trait of the algorithm.

We tested \mcesmppac{} in two domains, the multiagent Tiger problem and a large 3-agent version of the Firefighting domain commonly used in the cooperative multiagent community. In each case, \mcesmppac{} resulted in a statistically significant increase in the converged reward over initial start policies, while also benefiting from requiring less samples than the maximum bound due to early convergence. Thus, \mcesmppac{} offers a strong first effort in tackling purely model-free learning in cooperative multiagent settings.

\newpage
\chapter{Descriptive Reinforcement Learning as a Model of Human Reasoning}
\label{chap:psych}

As demonstrated in the previous chapters, reinforcement learning serves as an elegant and simple tool for arriving at optimal decision making. RL as a concept finds its roots in human learning, dating back far into the branch of psychology established by behaviorists~\cite{behaviorism_watson,behaviorism,experiment_skinner}. Mathematically founded for normative decision making in temporal-difference learning~\cite{tdlambda}, computational psychologists have long noted that humans, when tasked with making decisions, differ predictably from optimal behavior.

In this chapter, I introduce a parameterized descriptive reinforcement learning algorithm to predict the behavior of humans operating in a strategic, sequential environment~\cite{ceren2013modeling}. Leveraging concepts derived from behavioral game theory, machine learning, and prospect theory, I augment TD-learning, Q-learning, and the on-policy SARSA algorithms with parameters meant to capture the effect of biases precluding optimal behavior. Then, with data derived from real-world trials on human subjects, I train and test the algorithm for accuracy.

The data used for tuning the descriptive model was derived from a series of experiments involving human subjects playing a  strategic, sequential  game.  
Performed via a series of studies held in conjunction with the psychology department at the University of Georgia, participants  in  this study  observed  an unmanned  aerial
vehicle (UAV) navigate using differing trajectories through sectors in
a 4x4  grid, while  the theater was  shared with another  hostile UAV.
Participants were asked to assess the likelihood of their observed UAV reaching
a goal sector without being spotted by the hostile UAV whose movement is
fixed  but not  revealed, at  a series  of decision  points.   In this
complex  context,  we observed  remarkable  learning  and  provide  an
aggregate behavioral model of the learning.

I first describe the experiment in detail in Sec.~\ref{sec:uavexp} in order to motivate the discussions in the rest of the chapter. In Sec.~\ref{sec:biases}, I cover the relevant cognitive biases derived from behavioral game theory and prospect theory. I follow with Sec.~\ref{sec:behavioralqlearning}, which augments three popular RL approaches with parameters representing the descriptive reasoning discussed previously. Section~\ref{sec:biasresults} evaluates the predictive capabilities of the model using data collected from the psychology experiments.

\section{UAV Game: Experiment Evaluating Human Judgment}
\label{sec:uavexp}

\begin{figure}[!ht]
\begin{center}
 	\begin{minipage}{2.8in}
		\centerline{
		\includegraphics[width=2in]{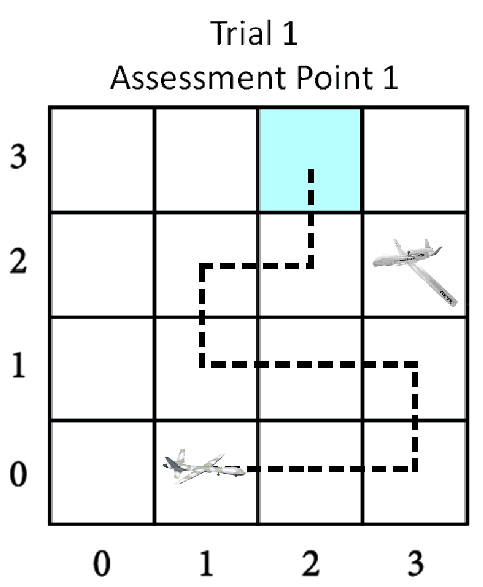}}
		{\centerline{{\small $(a)$}}}
	\end{minipage}
	\begin{minipage}{2.8in}
		 \centerline{\includegraphics[width=2in]{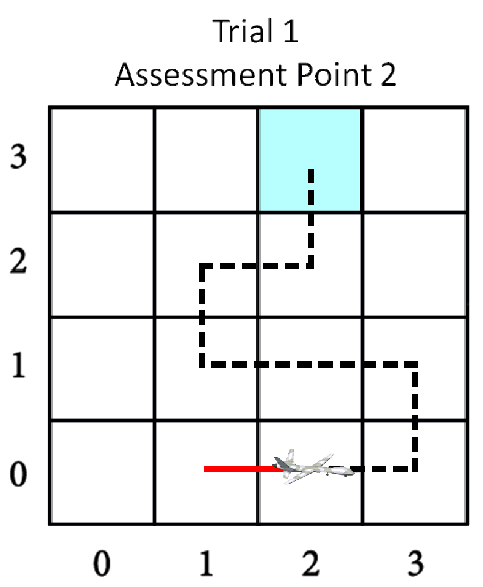}}
		{ \centerline{{\small $(b)$}}}		
	\end{minipage}
\end{center}
\caption{Two initial  decision points  of a trial  in the UAV  game as
  shown to the participants. $(a)$ For the first decision point at the
  starting location of the participant's UAV, the participant is shown
  the initial  location of  the other UAV  to confirm the  presence of
  another  UAV  in  the  theater.   $(b)$ Portion  of  the  trajectory
  traveled  so far  is  highlighted and  the  other UAV  is no  longer
  shown.}
\label{fig:gameboard}
\end{figure}

In  order to  evaluate probability  judgments of  human  operators, we
formulated  a  strategic  game  involving  uncertainty.  
In  this  sequential  game,  participants  observe  a  UAV  (hereafter
referred to as \textit{participant's UAV}) moving through a 4 $\times$
4 theater of sectors. From an  initial sector, the UAV moves towards a
goal sector,  as we  show in Fig.~\ref{fig:gameboard}  using different
trajectories.   The environment  is  shared with  another hostile  UAV
(hereafter referred  to as the  {\em other UAV}).   While participants
are briefed  about the starting  sector of the  other UAV and  that it
moves  in a  loop,  no other  information  such as  the  speed or  the
specific trajectory is revealed to them.  Unknown to the participants,
the   other   UAV's   trajectory    is   fixed   and   is   controlled
programmatically.

A  trial   representing  the  completion   of  a  trajectory   by  the
participant's UAV is considered a win if the participant's UAV reaches
the  goal  sector, which  is  considered  safe, or  a  loss  if it  is
``spotted'' by the  other UAV. A spot occurs if both  UAVs move to the
same sector, after which the trial ends. Fig. \ref{fig:gameboard} represents the first two sectors visited (or \textit{decision points}) of a trial. The gameboard grants clairvoyance of the entire trajectory for the current trial, the initial location of the enemy, and the already traveled course.

The goal of the experiment was to gather the assessments of the overall likelihood of a trial's success from participants. Given the knowledge of the initial location of the enemy, as well as the growing knowledge of its movements based on losses, this game exemplifies a learning task.

Participants play a  total of 20 trials of the  game decomposed into a
training  phase of  10 trials  and  a test  phase of  10 trials.   The
training phase is  provided to acquaint the participant  with the game
and  the  experiment  protocol.   Within  each  trial,  a  participant
encounters  a  series of  decision  points.  At  each decision  point,
participants  are asked  to fill  a questionnaire  and  in particular,
asked to  assess the  probability of reaching  the goal  sector safely
without being  spotted.  In Fig.~\ref{fig:gameboard},  we illustrate a
trial and the first two decision  points in the trial. Notice that the
participant is shown the entire trajectory that her UAV will travel to
facilitate an informed judgment.

As cursory analysis to identify if participants were expressing learning, we  performed a trend analysis  utilizing a
generalized  linear  regression  model.   The model  was  designed  to
determine the mean slope of the assessments in a trial averaged across
all trials and participants, and determine the gradient of the changes
in  the  mean  slope  as  the  trials  progressed  averaged  over  all
participants.  The assessments are modeled as changing linearly across
the  assessment points  and  the  mean slope  is  modeled as  changing
linearly as the  trials progress, with a random  intercept for initial
probability assessments.  We model  the participant anticipating a win
or a loss as having an effect on their assessments, which is justified
by  the  significant  difference  in  values  of  the  variables  when
analyzing data on wins and losses separately.

\begin{table}[!ht]
\begin{center}
\begin{small}
		\begin{tabular}{|l|c|c|c|c|} \hline
                  \multirow{2}{*}{\textbf{Statistic}} &
                  \multicolumn{2}{|c|}{\textbf{Losses}} &
                  \multicolumn{2}{|c|}{\textbf{Wins}}\\
                  \cline{2-5}
                  & \textbf{Estimate} & \textbf{S.E.} & \textbf{Estimate} &
                  \textbf{S.E.} \\
                  \hline
		\textit{intercept} & 0.3315 & 0.032 & 0.5392 & 0.03\\

		\textit{slope within trial} & 0.02053 & 0.006 &
                0.05395 & 0.005\\
		\textit{slope grad. bet. trials} & -0.00486 & 0.001 &
                -0.00129 & 0.000 \\
                \hline
              \end{tabular}
\end{small}
\end{center}
\caption{\small Mixed effect linear regression on probability judgments
  of participants separated by wins and losses in the trials. All
  values are significant, $p\ll$ 0.01.}
\label{tbl:glimmix-separate}
\end{table}

In  Table~\ref{tbl:glimmix-separate},  we  show  the  results  of  the
statistical analysis.   Notice the  positive intercepts for  both wins
and  losses and the  very small  standard errors  (S.E.)  with  $p \ll
0.01$ indicating a significant  fit.  More importantly, the mean slope
within a trial is positive  and a significant $p$-value indicates that
participant probabilities increase as they approach the goal.
Furthermore, the  negative value for  the mean slope  gradient between
trials indicates that  the mean slope reduces as  the trials progress,
and that this reduction is significant.

The  positive   mean  slope  indicates   that  participants  generally
demonstrate  greater  certainty   as  reflected  in  their  increasing
probabilities of reaching  the goal without being spotted,  as a trial
progresses and  they get  closer to the  goal.  This remains  true for
losses  as  well although  the  mean  slope  is substantially  smaller
compared to that for wins.

Importantly,  the negative gradient in mean  slope between trials
indicates  that  participants   are  not  changing  their  probability
assessments in a  trial as much as they were  in previous trials.  For
the ideal case where participants  precisely know how the other UAV is
moving, they would be certain about the outcome given their trajectory
and their  assessments would  not vary within  a trial.   Therefore, a
reducing  change  in  the  judgments  is  indicative  of  participants
gradually demonstrating  greater confidence in  their assessments.  We
interpret  these  results  as  indicative of  learning  from  previous
experiences.

\section{Cognitive Biases in Human Decision Making}
\label{sec:biases}

As observed throughout computational psychology literature, humans suffer from a wide variety of cognitive biases~\cite{heuristics}. In this section, I discuss three significant cognitive biases that affect the capabilities of humans making decisions in sequential, partially observable tasks: \textit{forgetfulness}, or the phenomenon by which humans do not exhibit clear memories of past experiences~\cite{roth95}, \textit{spill over}, whereby humans attribute experience in one physical location to neighboring locations, and \textit{subproportional weighting}, in which probabilistic judgments of uncertainty by humans is often categorically over- or under-weighted at the extremes~\cite{prospecttheory}. Spill over is similarly observed when humans consider neighboring strategies as well~\cite{wagenaar84}.

\begin{figure*}[!t]
\begin{center}
 	\begin{minipage}{1.1in}
		\centerline{
		\includegraphics[width=1.1in]{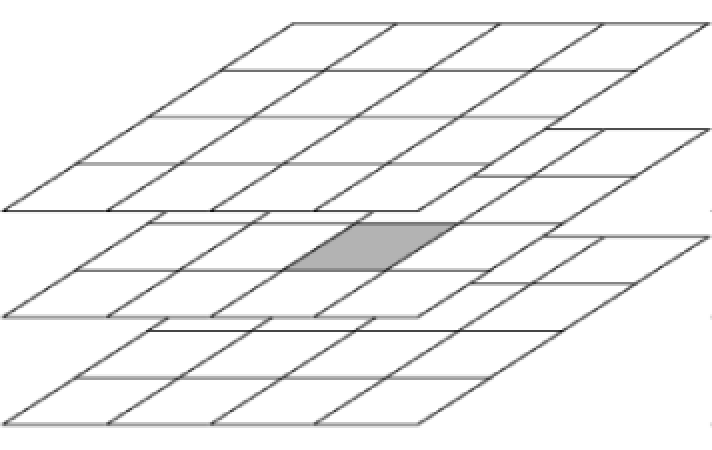}}
		{\centerline{{\small $(a)$}}}
	\end{minipage}
	\begin{minipage}{1.1in}
		\centerline{\includegraphics[width=1.1in]{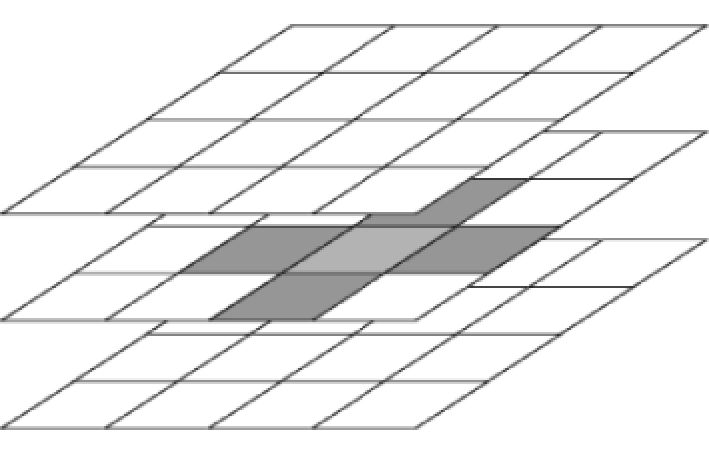}}
		{ \centerline{{\small $(b)$}}}		
	\end{minipage}
	\begin{minipage}{1.1in}
		\centerline{\includegraphics[width=1.1in]{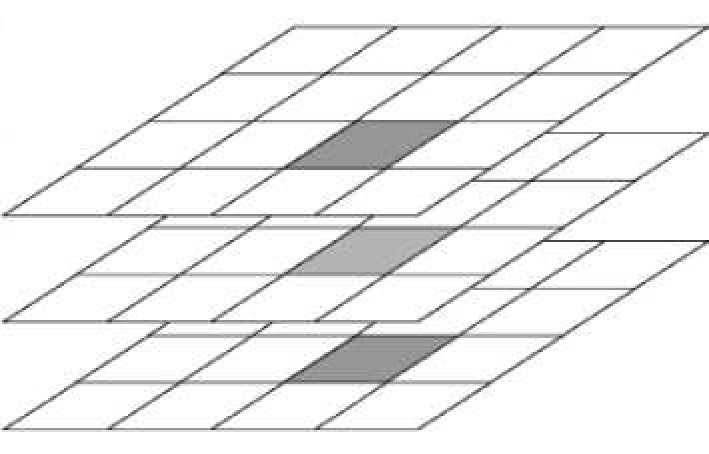}}
		{ \centerline{{\small $(c)$}}}		
	\end{minipage}
	\begin{minipage}{1.1in}
		\centerline{\includegraphics[width=1.1in]{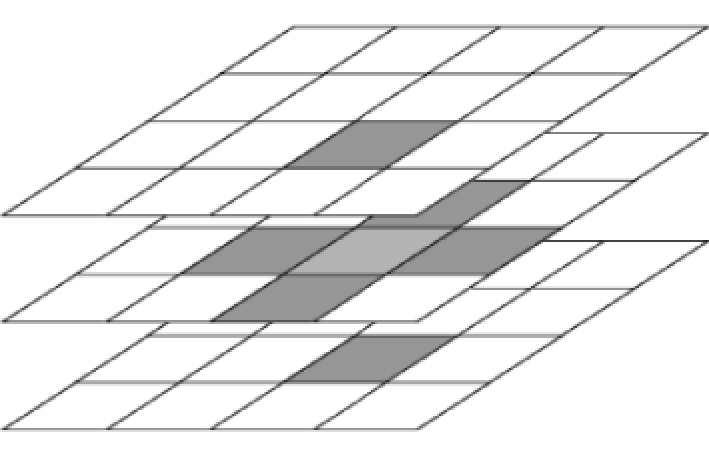}}
		{ \centerline{{\small $(d)$}}}		
	\end{minipage}
	\begin{minipage}{1.1in}
		\centerline{\includegraphics[width=1.1in]{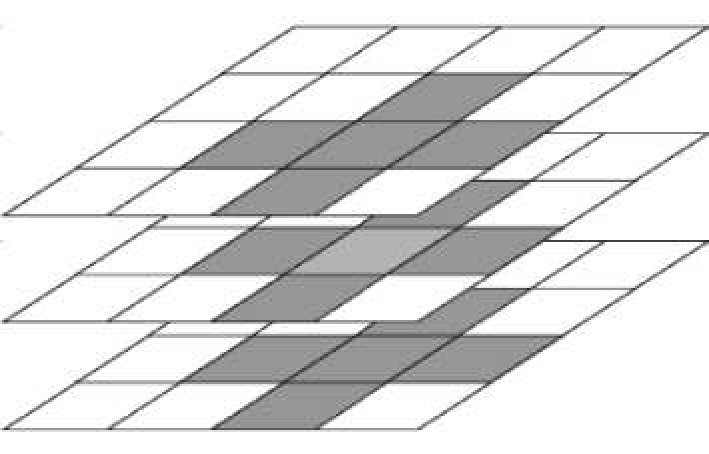}}
		{ \centerline{{\small $(e)$}}}		
	\end{minipage}
\end{center}
\caption{Differing ways  of implementing spill  over.  $(a)$ Normative
  case  where no  other  neighbors  are attributed  a  reward from  an
  experienced  sector.  $(b)$  Adjacent sectors  at current  time step
  receive some spill  over reward.  $(c)$ No neighbors  at the current
  time  step  receive a  spill  over but  the  visited  sector in  the
  preceding  and  following time  steps  are  attributed some  reward.
  $(d)$  Merges spill  overs of  $(b)$  and $(c)$  such that  adjacent
  sectors at  current time step,  and visited sector at  preceding and
  subsequent time  steps receive some  of the reward.   $(e)$ Adjacent
  sectors at current, preceding and following time steps receive spill
  over rewards.}
\label{fig:spillover}
\end{figure*}

While playing repeated games,  humans are observed as often forgetting
the        previous        history        of        actions        and
observations~\cite{roth95,camerer03}, thereby  violating the principle
of perfect  recall~\cite{kuhn}.  Such a phenomenon can be expected to arise when humans play strategic, sequential games, particularly those with high complexity, such as the partially observable UAV game present in this chapter. However, as the class of temporal difference RL algorithms maintain a learning rate parameter $\alpha$ (as appears in Sec.~\ref{sec:rl}), the computational effect of reward depreciation is already well-modeled. As such, I omit it from the descriptive RL model, though it is important to note that this phenomenon exists.

Spill over describes the generalizations of experiences to neighboring strategies, but surfaces as a subnormative misattribution of experience to locations outside of, but near, the location in which stimuli has occurred. Figure~\ref{fig:spillover} illustrates a variety of spill over models that potentially describe the behavior of human subjects in the UAV game. Not only do I consider the possibility that nearby sectors are affected by being caught by the enemy in a participant's mind, but also the possibility the state becomes taboo entirely; that is, participant's are concerned that, at any time, they may be caught in that sector. This latter case is well-established in eligibility traces~\cite{etracespeedup}, but the former remains relatively limited to the human computational modeling literature~\cite{wagenaar84}.

\begin{figure*}[!ht]
\begin{center}
 	\begin{minipage}{1.75in}
		\centerline{
		\includegraphics[width=1.75in]{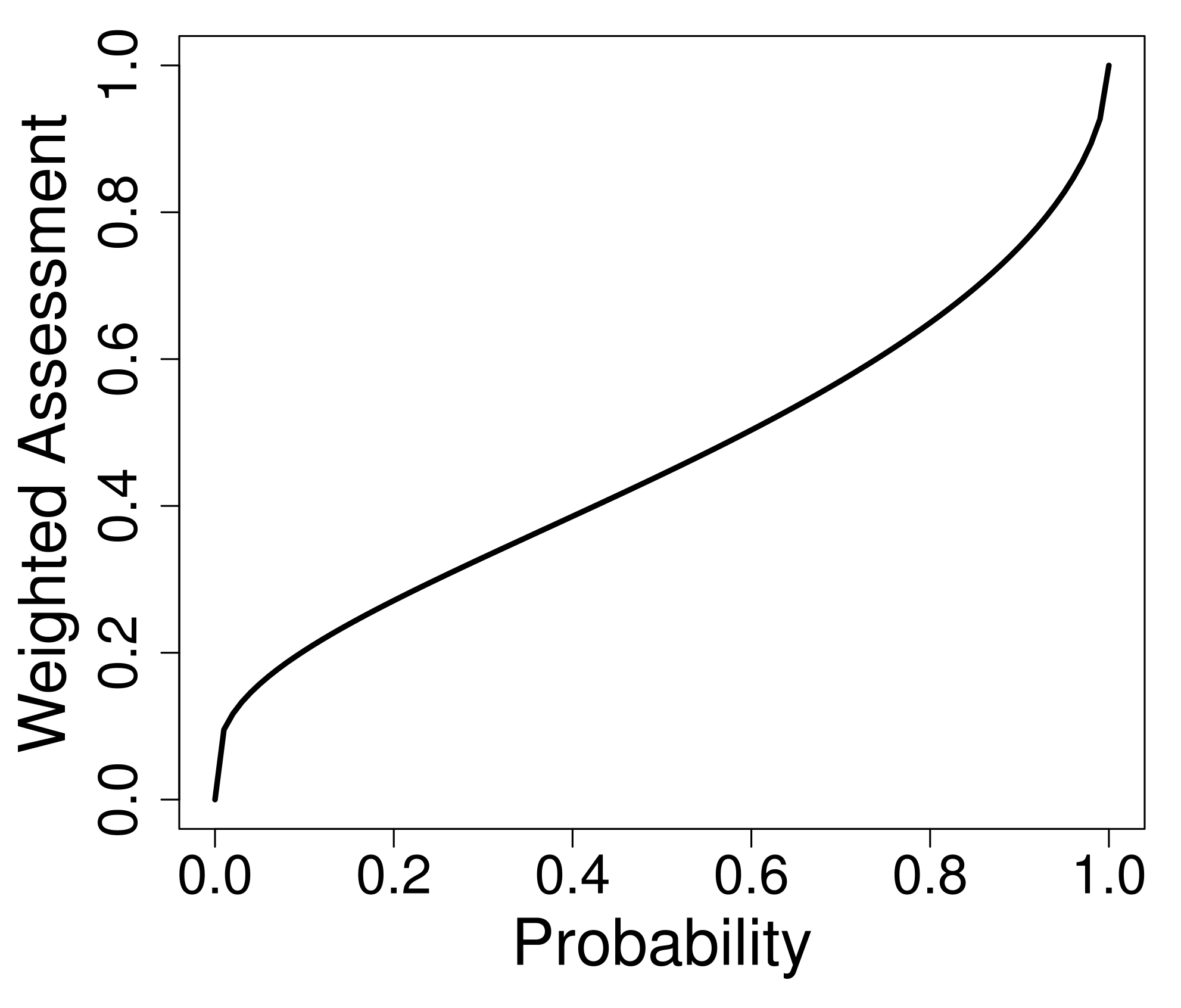}}
		{\centerline{{\small $(a)$}}}
	\end{minipage}
	\begin{minipage}{1.75in}
		 \centerline{\includegraphics[width=1.75in]{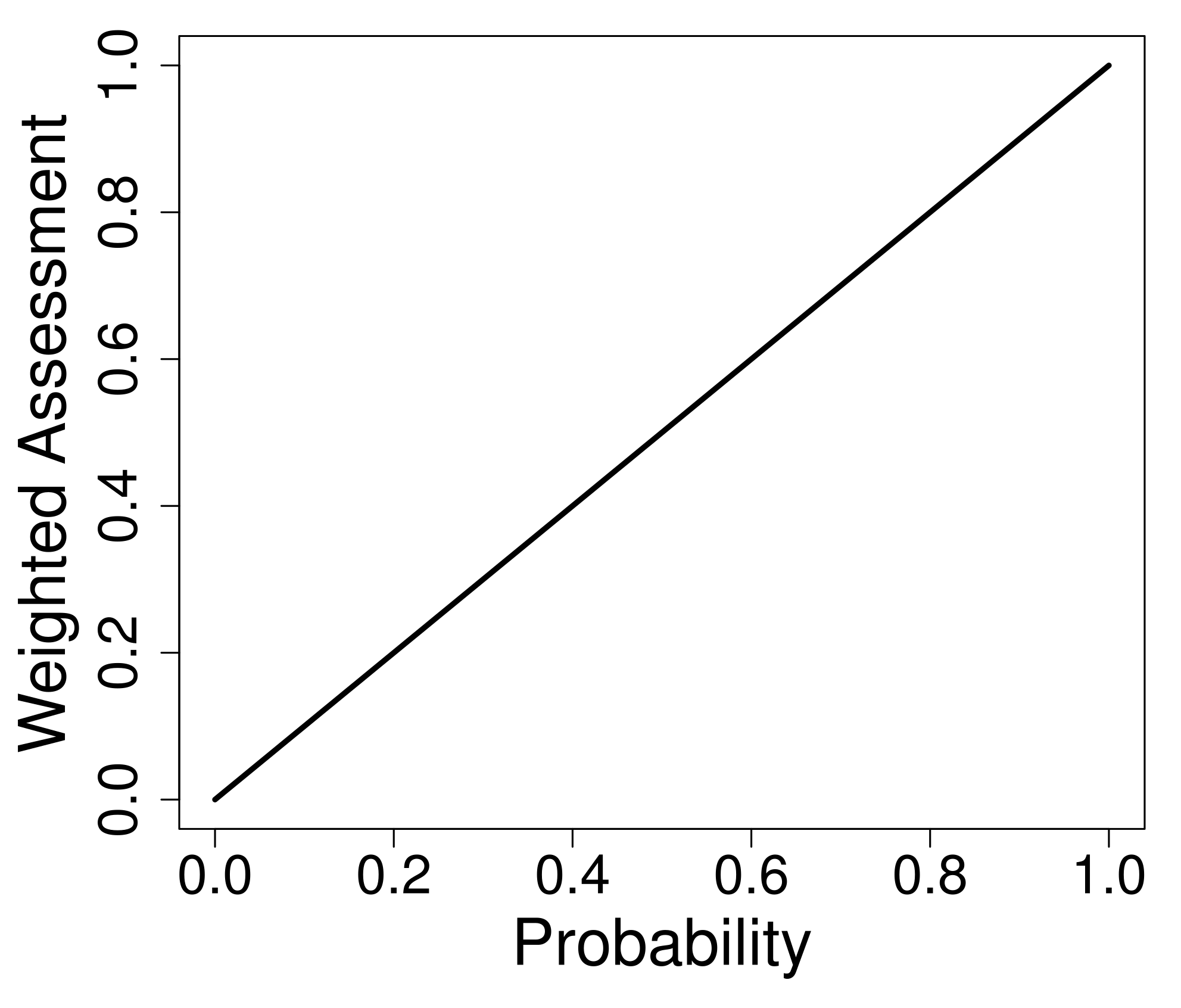}}
		{ \centerline{{\small $(b)$}}}		
	\end{minipage}
	\begin{minipage}{1.75in}
		 \centerline{\includegraphics[width=1.75in]{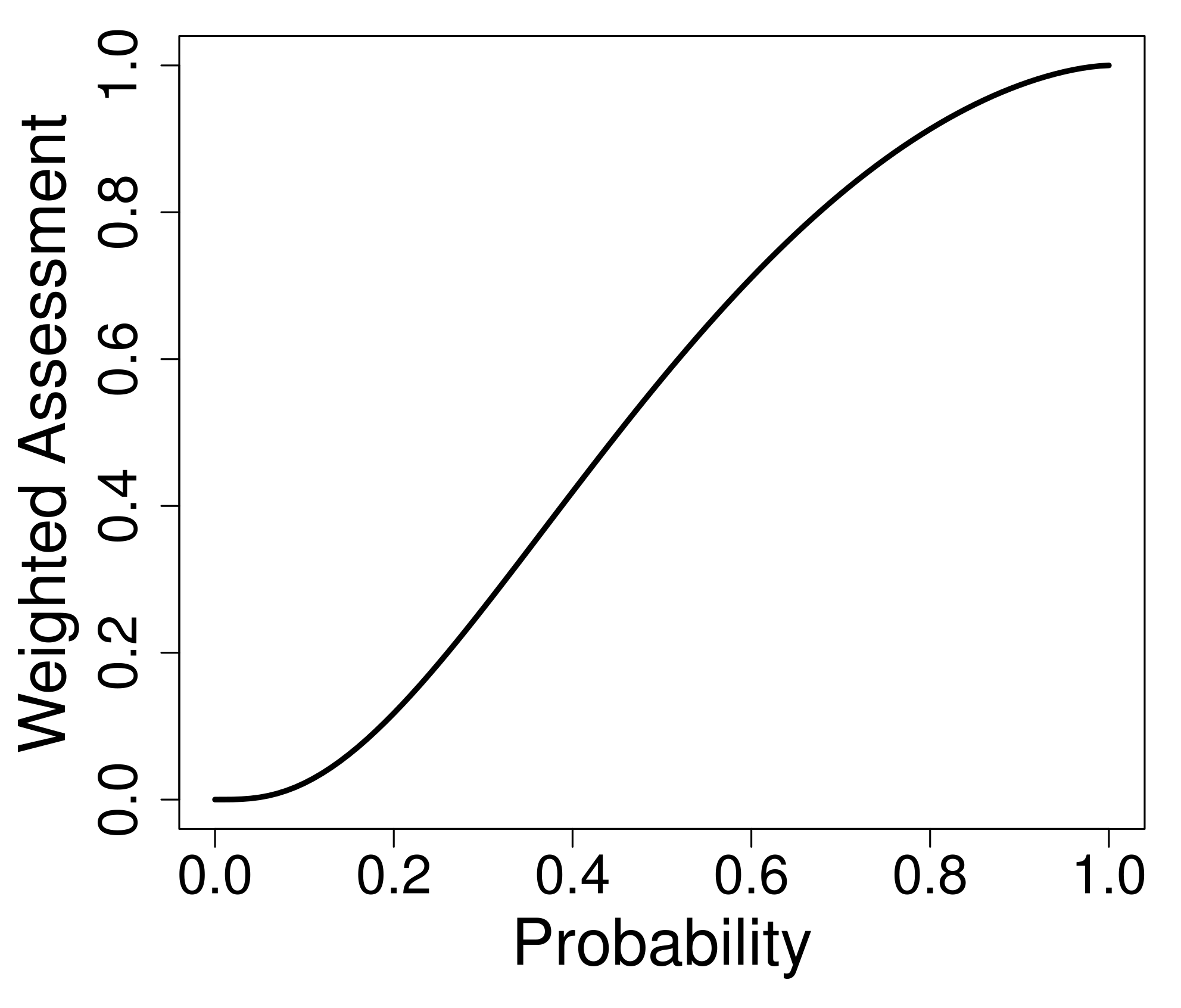}}
		{ \centerline{{\small $(c)$}}}		
	\end{minipage}
\end{center}
\caption{Prelec's  one  parameter  subproportional weighting  function
  with,  $(a)$  $\beta=0.56$,  $(b)$  $\beta=1$  (linear),  and  $(c)$
  $\beta=1.6$}
\label{fig:prob_weight}
\end{figure*}

\begin{align}
w(p;\beta)=exp\{-(-ln(p))^{\beta}\}
\label{eq:subweight}
\end{align}

It  is now  generally known  that humans  do not  weight probabilities
linearly in  their decisions. In response to Kahneman's original non-formulaic discussion of prospect theory~\cite{prospecttheory}, describing this phenomenon of \textit{subproportional weighting}, several parametric non-linear models arose to describe the behavior of humans under- or over-weighting probabilities when tasked with making judgments in games of chance. Initially proposed by Gonzales and Wu~\cite{gonzales99}, the two-parameter model allows the curvature of the probability weighting curve to vary in two ways: convexity as a sigmoid or, for negative values, an inverse sigmoid, and the center elevation. Prelec~\cite{prelec} simplified this model to just convexity. Figure~\ref{fig:prob_weight} illustrates the variety of probability weights that may be expressed, and Eq.~\ref{eq:subweight} describes the one-parameter model, where $p$ is the true believed probability, $\beta$ is the curvature exponent, and $w(\cdot)$ is the weighted, expressed probability.

\section{Process Model for Behavioral Reinforcement Learning}
\label{sec:behavioralqlearning}

Where learning may be characterized as the observed change in behavior due to experience~\cite{camerer03}, and where the change in behavior is due to maximizing the propagation of positive stimuli defines the concept of reinforcement~\cite{schedules_skinner}, the UAV game described in Sec.~\ref{sec:uavexp} well defines a domain in which humans are likely to exhibit reinforcement learning. Being caught by the enemy UAV provides negative reinforcement for being in that sector, while making it to the goal provides strong positive reinforcement for the sectors in the trajectory. Consequently, I hypothesize the data, which contains the probability assessments of the participants expressing their likelihood of victory, can best be explained with a model of reinforcement learning.

Using the class of temporal difference learners, TD$(\Lambda)$, Q-learning, and SARSA, as my departure point, I parameterize the effects of spill over and subproportional weighting as additions to the normative models. Recall Eq.~\ref{eq:td}, in which the algorithm for TD$(\Lambda)$ is described. I first parameterize the effect of spill over by attributing a fraction of the reward with a tuned parameter $\epsilon$. The augmented equation for the sector in which the stimuli occurred is as follows.

\begin{align*}
V(s;\alpha,\epsilon,\lambda)=V(s)+\alpha((1-\epsilon)r(s)+\gamma\cdot V(s')-V(s))e(s;\lambda)
\end{align*}

Dependent on the version of spill over that is used, as described in Fig.~\ref{fig:spillover}, the neighbors (defined as dark grey sectors) receive the following proportion of reward.

\begin{align}
V(s;\alpha,\epsilon,\lambda)=V(s)+\alpha(\epsilon\cdot r(s)+\gamma\cdot V(s')-V(s))e(s;\lambda)
\label{eq:nbr}
\end{align}

Q-learning, as an off-policy RL technique, learns the action-values for each state instead. However, the rewards can be simply propagated to neighboring states in a similar fashion.

\begin{align*}
Q(s,a;\alpha)\gets Q(s,a)+\alpha((1-\epsilon)r(s)+\gamma \max_{a'}Q(s',a')-Q(s,a))
\end{align*}

The on-policy Q-learning implementation SARSA is straightforward as well.

\begin{align*}
Q(s,a;\alpha)\gets Q(s,a)+\alpha((1-\epsilon)r(s)+\gamma Q(s',a')-Q(s,a))
\end{align*}

For both implementations, the neighbor equations are omitted, but follow analogously from Eq.~\ref{eq:nbr}.

In order to apply the effect of subproportional weighting, the empirical values calculated by these descriptive RL models must be mapped to probabilities.
Observe  that values
approaching -1  represent a path  likely to lead  to a loss  and those
approaching 1  indicate a  win from that  path.  Because the  value is
representative of the {\em desirability} of the state (and action), it
maps  to  the  likelihood  of   success  from  that  state  given  the
trajectory, naturally.  We may  then convert the values to assessments
by normalizing them  between 0 and 1. With this normative expression of the likelihood of success, the empirical value of the Q-function may be applied to Eq.~\ref{eq:subweight} as the parameter $p$.

\section{Experiments and Results}
\label{sec:biasresults}

With the descriptive RL model formalized in Sec.~\ref{sec:behavioralqlearning}, three parameters require tuning to predict the behavior of humans in the UAV game. $\alpha$ represents the learning rate, as in canonical Q-learning, but also the forgetfulness of the participant. $\epsilon$ captures the effect of spill over between sectors where stimuli occur (the goal sector and being caught). Lastly, $\beta$ represents the convexity of the subproportional weighting function, describing how participants are under- and over-weighting their assessments of success.

Data collected from the 43 participants were randomly partitioned into
5 folds, with data from  8-9 participants in each fold.  Utilizing the
Nelder-Mead method~\cite{neldermead} --  a downhill simplex method for
minimizing an objective function --  the model is trained over 4 folds
and then,  to test  the predictive capabilities  of the  model, tested
over the  remaining fold.  Additionally, I  consider simple baselines
for comparison.  These include the {\sf default} model, which does not
include any of  the behavioral factors, and the {\sf  random} model, which
estimates  the probabilities  at each  decision point  within  a trial
randomly.

Beginning at the first decision point of the first trial, the Q-values
are updated as the participant's  UAV follows its trajectory. On being
spotted, a reward of -1 is  obtained for that state. If it reaches the
goal sector,  a reward of 1  is obtained for the  state, otherwise the
reward  is  0.  I  update  the  function  and simultaneously  predict
probabilities for the 20  trials that each participant experiences and
for all the participants in the training folds. Parameters are learned
by minimizing the sum of  squared differences (SSD) between the stated
probabilities of participants, $n$, at  each decision point, $i$, in a
trial,  $t$,   $p_e(i,t,n)$,  and   those  predicted  by   our  model,
$p_m(i,t,n)$. We may interpret this difference as the {\em fit} of the
model with smaller differences signifying better fits. Formally,

\begin{equation}
	SSD = \sum \limits_{n=1}^N \sum \limits_{t=1}^{20} \sum
\limits_i (p_m(i,t,n) - p_e(i,t,n))^2
\label{eq:likelihood_func}
\end{equation}
where, $N$ is the number of participants in the training folds and $i$
is the number of UAV actions in a trial which vary.

I begin  by learning the  parameter values and establishing  the best
fitting spill over among  those shown in Fig.~\ref{fig:spillover}.  I
implement each spill over in each  of the default models and perform a
5-fold  cross  validation  summing  the  SSDs  over  the  test  folds.
Table~\ref{tbl:spillover_fits} lists the  SSDs for the different spill
over implementations in each model. Notice that each implementation in
SARSA provides the best fit  among the different learning models.  For
SARSA,  the implementation  which  spills the  reward across  adjacent
sectors ({\bf local}) results in the best fit, indicating that participants felt wary of the state, but considered how many steps had already occurred.

\begin{table}[ht]
\begin{center}
\begin{tabular}{|r|c|c|c|c|} \hline
 & \textbf{TD(0)} & \textbf{TD(1)} & \textbf{SARSA} & \textbf{Q-learning} \\ \hline
 \textbf{No Spill over} & 360.187 & 407.469 & {\bf 355.569} & 379.463 \\ \hline
 \textbf{Local} & 351.258 & 382.345 & {\bf \underline{341.923}} & 372.306 \\ \hline
 \textbf{Time Step} & 361.141 & 400.23 & {\bf 346.425} & 371.776 \\ \hline
 \textbf{Local \& Time Step} & 356.304 & 392.01 & {\bf 354.161} & 364.861 \\ \hline
 \textbf{All Neighbors} & 348.165 & 383.386 & {\bf 343.458} & 358.075 \\ \hline
\end{tabular}
\end{center}
\caption{{\small SSDs for the different spill
    over implementations shown in Fig.~\ref{fig:spillover} summed over
    all test folds. SARSA
    provides the lowest SSD for each spill over implementation and 
    {\bf Local} fits the best.}}
\label{tbl:spillover_fits}
\end{table}

Table~\ref{tbl:descriptive_output}  shows the  learned  values of  the
three behavioral  parameters in  each descriptive model  utilizing the
spill  over implementation  that results  in the  lowest SSD  for that
model.  For all models, $\gamma$ was fixed to 0.9.   A $1-\alpha$
value  of  0.421  for   SARSA  signifies  that  participants  place  a
moderately lower emphasis on their previous experiences as compared to
the  current   and  future   reward,  thereby  forgetting   them.   We
experimented with a  linearly varying $\alpha$ as well  resulting in a
worse fit. On  the other hand, the spill over  is negligible for SARSA
but   substantial   for   the   other  models.    Curvature   of   the
subproportional weighting as parameterized  by $\beta$ remains above 1
for all models indicating that the function is sigmoidal.

\begin{table}[!ht]
\begin{center}
\begin{tabular}{|r|c|c|c|c|} \hline
 & \textbf{TD(0)} & \textbf{TD(1)} & \textbf{SARSA} & \textbf{Q-learning} \\ \hline
 \textbf{$\alpha$} & 0.570 & 0.750 & 0.579 & 0.491 \\ \hline
 \textbf{$\epsilon$} & 0.215 & 0.463 & 0.0004 & 0.809 \\ \hline
 \textbf{$\beta$} & 1.905 & 1.785 & 2.045 & 1.420 \\ \hline
 \textbf{Total SSD} & 348.165 & 382.345 & 341.923 & 358.075 \\ \hline
\end{tabular}
\end{center}
\caption{\small Learned parameter values of the different models.}
\label{tbl:descriptive_output}
\end{table}

Table~\ref{tbl:fits}   shows  the   comparative  performance   of  the
different reinforcement  learning models including a  random model. Observe that the corresponding behavioral generalization improves each
default model with  {\sf Behavioral SARSA} showing the  lowest SSD and
therefore the  best fit.  It outperforms the  next best  model (TD(0))
significantly  (Student's  paired,  two-tailed t-test,  $p$-value  $<$
0.05), as well as its default model. 

\begin{table}[!ht]
\centering
\begin{small}
\begin{tabular}{|l|l|} \hline
{\bf Model} & {\bf total SSD}\\
\hline \hline
{\sf Behavioral SARSA} & \underline{\bf 341.923} \\ \hline
{\sf Behavioral TD(0)} & 348.165 \\ \hline
{\sf Default SARSA} & 355.569 \\ \hline
{\sf Behavioral Q-learning} & 358.075 \\ \hline
{\sf Default TD(0)} & 360.187 \\ \hline
{\sf Default Q-learning} & 379.463 \\ \hline
{\sf Behavioral TD(1)} & 382.345 \\ \hline
{\sf Default TD(1)} & 407.469 \\ \hline
{\sf Random} & 891.18 \\ \hline
\end{tabular}
\end{small}
\caption{\small Behavioral SARSA shows the  best fit and the
  differences with others are significant.}
\label{tbl:fits}
\end{table}

To illustrate the predictive performance of descriptive Q-learning, I select the best performing model
from  the previous subsection,  {\sf Behavioral  SARSA}, and  plot its
performance.    Because  the   experiment  utilized   trajectories  of
differing lengths for  the participant's UAV between the  win and loss
trials, and  the collected data  also exhibits difference  between the
two (for  e.g., see Table~\ref{tbl:glimmix-separate}),  we present the
results separately for the two types of trials for clarity. However, a
single model was trained over the participant data.

\begin{figure}[!ht]
\begin{center}
 	\begin{minipage}{1.63in}
		\centerline{
		\includegraphics[width=1.63in]{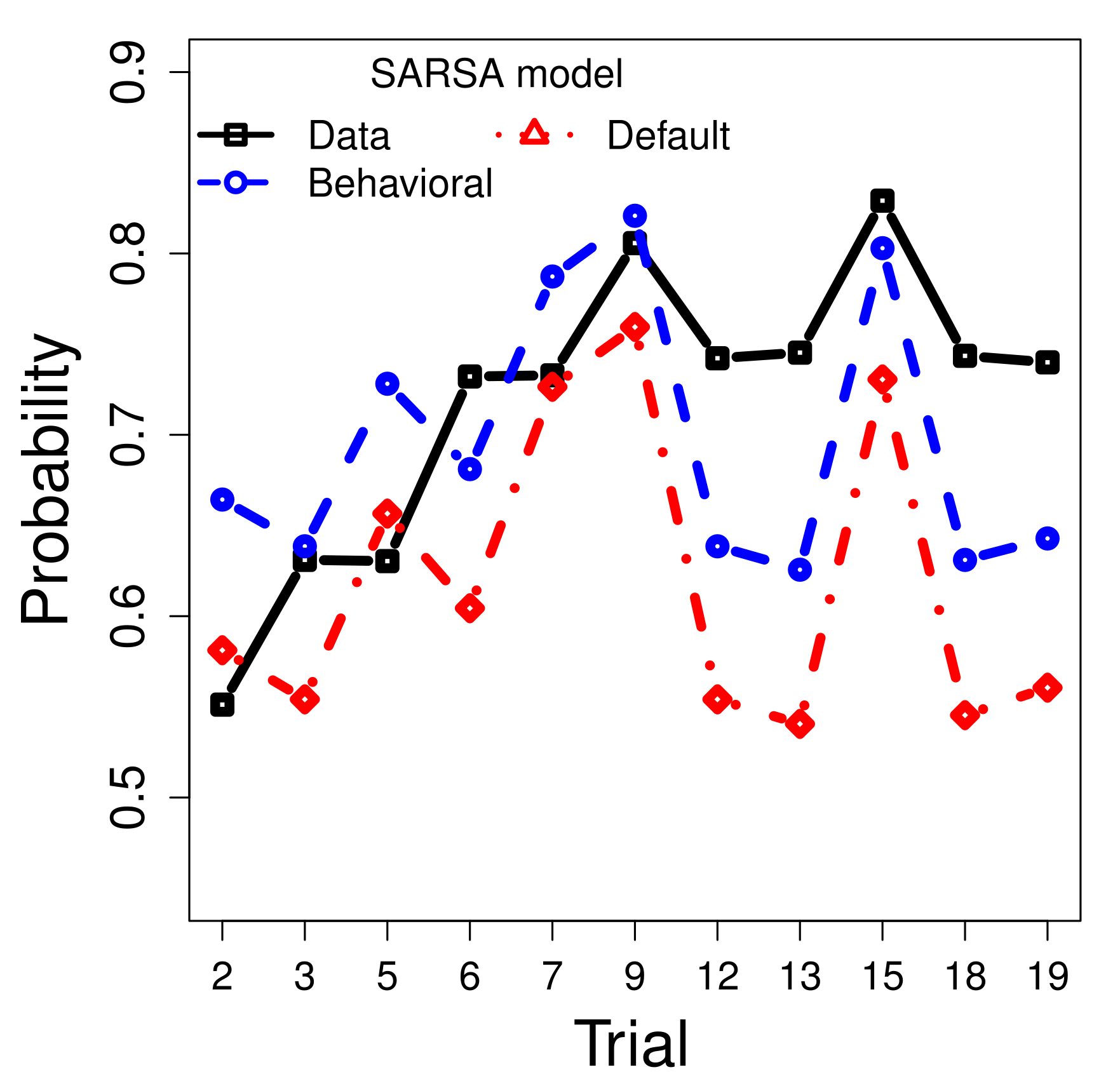}}
		\centerline{{\small $(a)$}}		
	\end{minipage}
 	\begin{minipage}{1.63in}
		\centerline{
		\includegraphics[width=1.63in]{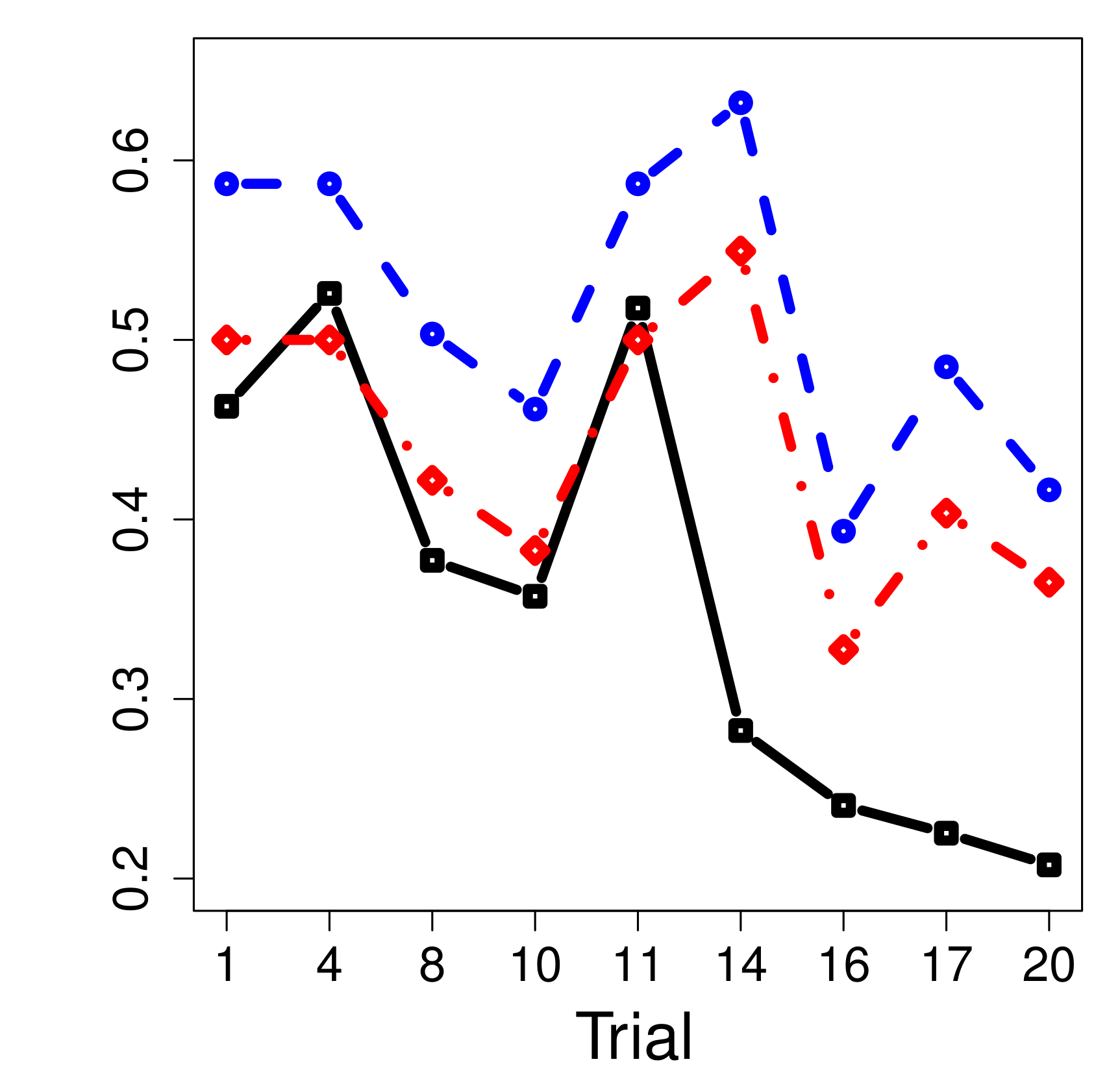}}
		\centerline{{\small $(b)$}}		
	\end{minipage}
\end{center}
\caption{\small  Average  probability  assessment  in each  trial  for
  trajectories  that  lead to  $(a)$  successfully  reaching the  goal
  sector, and $(b)$ being spotted by the other UAV.}
\label{fig:trials_across}
\end{figure}

\begin{figure*}[!t]
\begin{center}
 	\begin{minipage}{2.1in}
		\centerline{
		\includegraphics[width=2.1in]{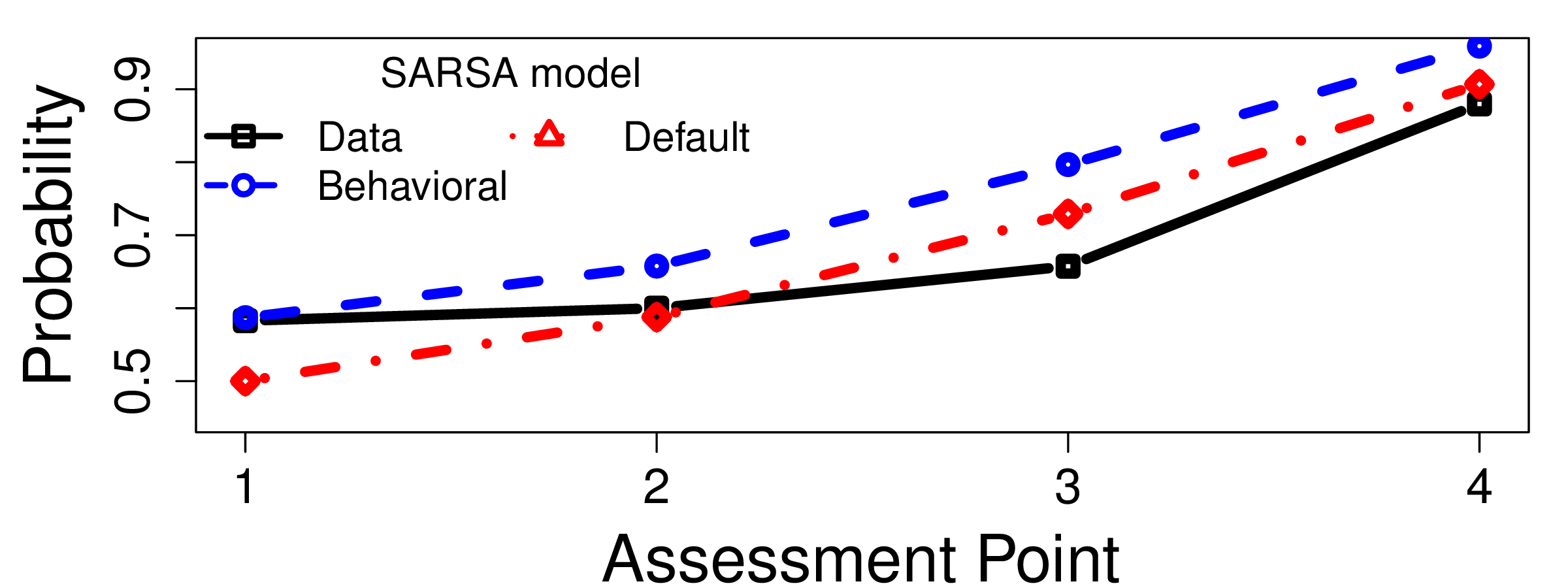}}
		\centerline{{\small $(a)$}}
	\end{minipage}
	\begin{minipage}{2.1in}
		 \centerline{\includegraphics[width=2.1in]{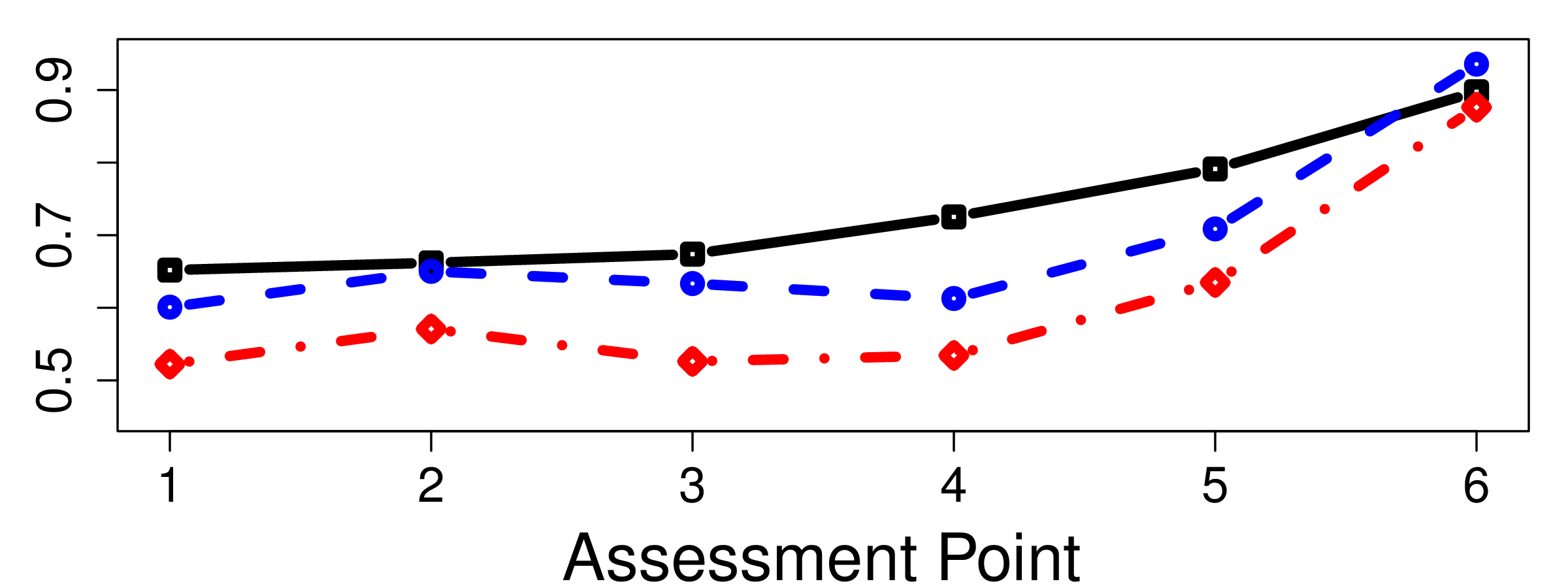}}
		\centerline{{\small $(b)$ }}
	\end{minipage}
	\begin{minipage}{2.1in}
		 \centerline{\includegraphics[width=2.1in]{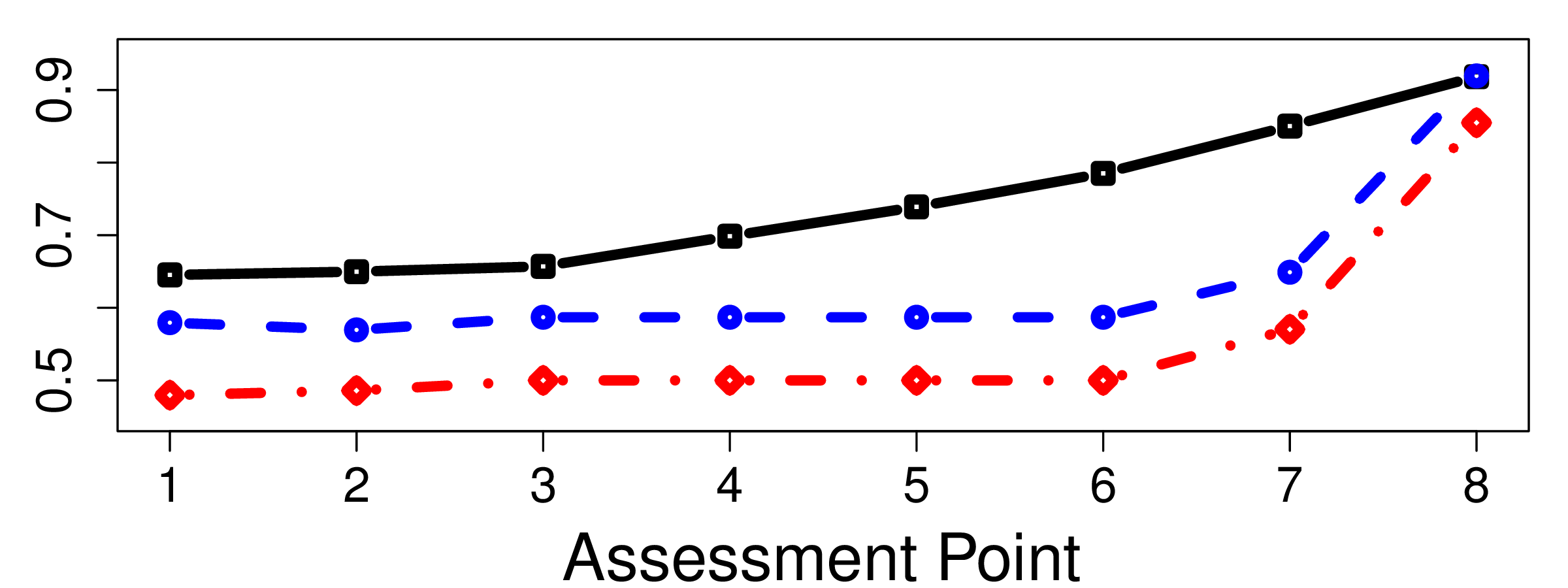}}
		\centerline{{\small $(c)$ }}
	\end{minipage}
\end{center}
\caption{\small  Comparison of  predicted judgments  by  the different
  models with the  experiment data for trajectory lengths  of, $(a)$ 4
  time steps, $(b)$ 6 time steps, and $(c)$ 8 time steps. Vertical
  bars are the standard errors.}
\label{fig:decpt_win}
\end{figure*}

In Fig.~\ref{fig:trials_across}, we  show the average probability over
all assessment  points and participants  for each trial,  separated by
winning and losing trials.  For wins and losses, model predictions fit
the  general shape  of the  probability changes  closely. For  the win
trials, {\sf  Behavioral SARSA}  effectively models the  changing mean
assessments per trial up to and  including the last trial. We show the
performance  of  the default  SARSA  across  the  trials as  well.  As
Table~\ref{tbl:fits}  suggests,  {\sf  Behavioral SARSA}  demonstrates
improved predictions across the trials compared to the default.

Trajectories  that result in  a win  are of  lengths 4,  6, or  8 time
steps.   Probability  assessments are  substantially  affected by  the
distance  to  the  goal sector,  so  we  analyze  the data  and  model
predictions    separately    for   each    of    these   lengths    in
Fig.~\ref{fig:decpt_win}   next.    While   {\sf   Behavioral   SARSA}
understates the probabilities in comparison to the data for the longer
trajectories,  it  exhibits  the   overall  trend  correctly  for  the
trajectories of different lengths.

\begin{figure}[!ht]
\begin{center}
 	\begin{minipage}{1.63in}
		\centerline{
		\includegraphics[width=1.5in]{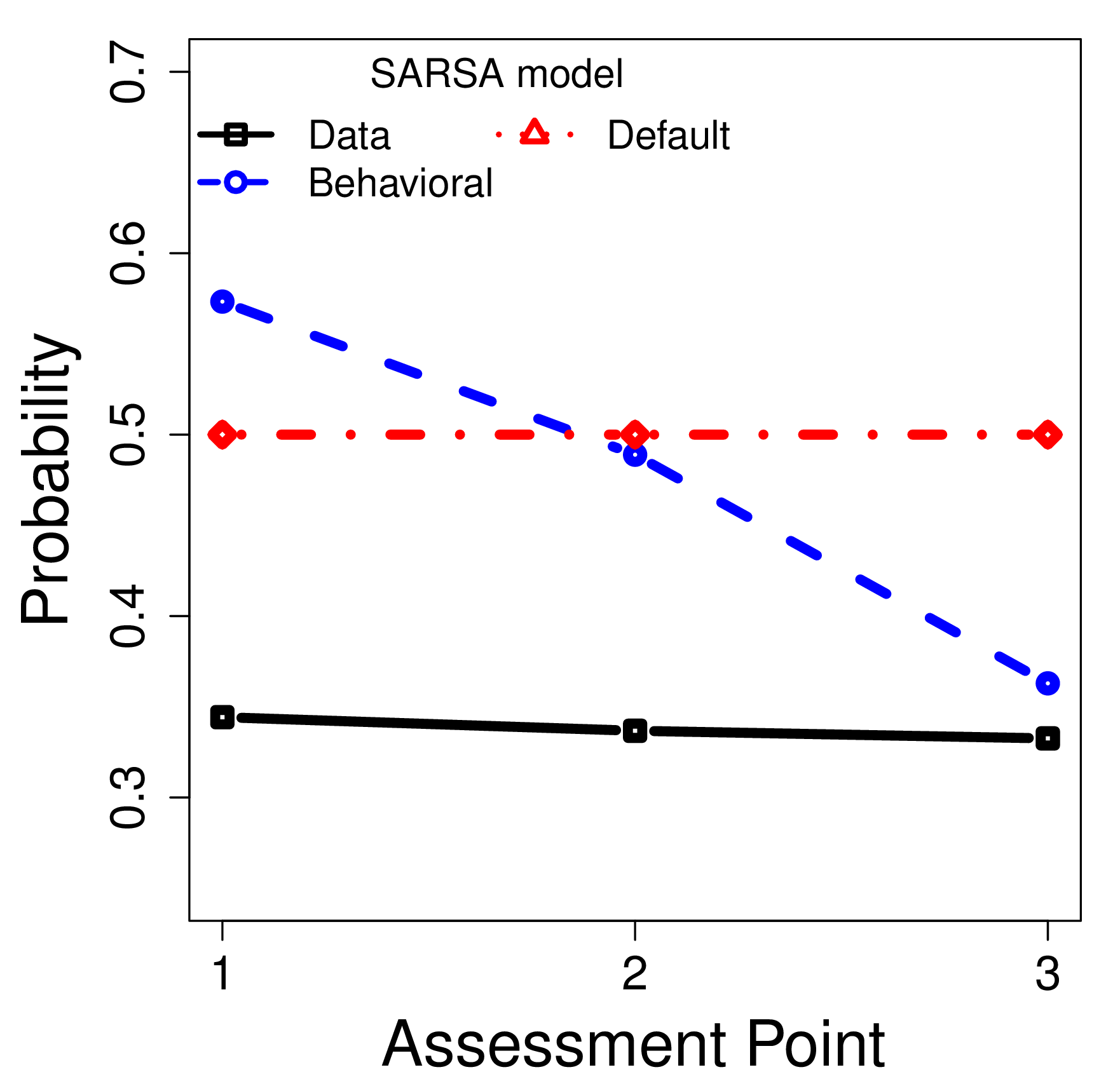}}
		\centerline{{\small $(a)$ }}
	\end{minipage}
	\begin{minipage}{1.63in}
		 \centerline{\includegraphics[width=1.5in]{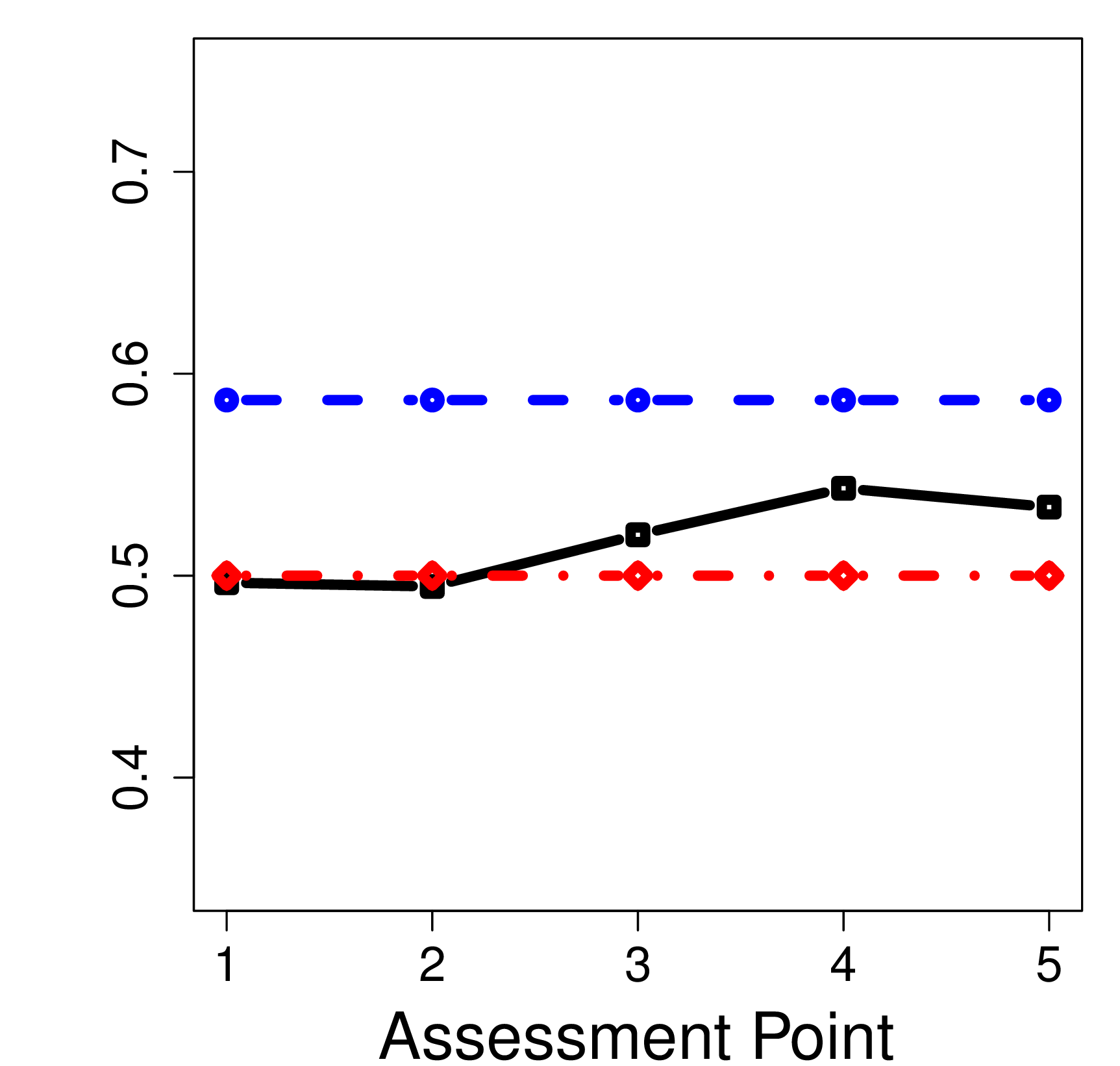}}
		\centerline{{\small $(b)$}}
	\end{minipage}
\end{center}
\caption{\small A  comparison of model predictions  with observed data
  for loss  trajectories. We show the comparisons  for trajectories of
  lengths $(a)$ 3 time steps, and $(b)$ 5 time steps.}
\label{fig:decpt_loss}
\end{figure}

In  Fig.~\ref{fig:decpt_loss}, we compare  the model  predictions with
the data averaged over all the trials that result in a loss. 
Participants,
on  average, start  with  lower assessments  compared  to trials  that
result in a win.  This  indicates that participants are generally good
at identifying  eventual losses and  retain their pessimism  as trials
progress.  The  models show  higher initial judgments  and exaggerated
decreases in  their probability predictions over time  compared to the
slight dip in  probabilities we observed for trajectories  of length 3
(Fig.~\ref{fig:decpt_loss}$(a)$).    For    the   longer   trajectory,
participants generally became more  optimistic until just before their
loss,  while the  models' predictions  averaged over  all  such trials
remain mostly  flat.  The  primary reason is  the lack  of substantive
data because each participant  experiences just one trial that results
in being spotted after 5 steps.

\section{Concluding Remarks}

Significant research has established that humans deviate from normative reasoning, largely due to the effects of cognitive biases that preclude rational decision making. Forgetfulness, which causes humans to depreciate the value of previous events, spill over, where humans generalize their experience to similar strategies or locations nearby, often erroneously, and subproportional weighting, when expressed probability deviates at high and low values, are some of these biases. In this chapter, I modified reinforcement learning techniques to capture the effects of these biases in actual human experiments.

By  utilizing  behavioral  parameters  in a  sequential  RL model (Q-learning) and testing its predictive capabilities on
data  collected  from  a   strategic  game,  I  established  improved
performance when compared to the default RL models
and other simple baseline models. In the experiment, participants were
required to estimate their likelihood of a UAV successfully navigating
to a goal sector without being spotted by a hostile UAV.
I further improved the performance  of the model by using a nonlinear
probability  weights  to  map   the  model's  probability
assessments accounting for under- or overstatements of probabilities.

The descriptive reinforcement learning model gets us close to modeling
the human  judgments in a strategic  task, but shows  room for further
improvement. Certain behaviors such  as that of participants assessing
increasing  likelihoods  of reaching  the  goal  sector without  being
spotted only to suddenly reduce it  in the later stages when they find
themselves about to  be spotted, as in Fig.~\ref{fig:decpt_loss}$(b)$,
are challenging to computationally  model. Indeed, here the model's average
predictions do not change much over time.

Despite this, descriptive Q-learning serves as a powerful method for computationally modeling the behavior of humans in sequential games under uncertainty, where participants refine their predictions as they experience more trials. I show that the parameterized models significantly outperform normative models of learning, and illustrate the predictive capabilities of the model in relation to real-world data.

\newpage
\chapter{Game-Delayed Reinforcement Learning in Robotic Team Precision Agriculture}
\label{chap:ag}

Like the previous chapter, this chapter examines the application of parameterized reinforcement learning in a real-world domain. In this chapter, I examine a novel formation of traditional reinforcement learning problems: game-delayed reinforcements. In most contemporary literature, RL is applied to either single-shot games or sequential domains wherein stimuli is achieved immediately following the last action. In some contexts, however, stimuli is achieved at some other stage the agent is not involved in, as is the case in teams of heterogeneous agents with specialized tasks.

In this setting, one agent generates the frame of a problem for another agent, and the subsequent agent resolves it, informing the first agent of the results. There are two unique challenges here: the first agent frequently makes decisions based on a prior that precedes recently executed actions and, additionally, noise in observations from subsequent agents are propagated to previous agents. I show, via empirical results, that agents are still able to reasonably learn under these constraints in a novel assembly of heterogeneous agents leveraging \mcesp{}.

The field of precision agriculture is rapidly growing due to availability of low-cost computation and robotic physical systems~\cite{precag}. I define a real-world RL implementation by leveraging popular image processing techniques~\cite{plant_detection} as discretized observations for a heterogeneous team of simulated agents and show \mcesp{} results in remarkable performance under game-delayed reinforcements. I introduce a heuristic that allows RL-like exploration of foregone strategies~\cite{whitehead} that dramatically improves both canonical Q-learning~\cite{watkins} and promotes significantly more positive identifications of crops under stress.

\section{Representing Heterogeneous Teams in Precision Agriculture}
\label{sec:domain}

\begin{figure} [htb]
\centerline{\includegraphics[width=8cm]{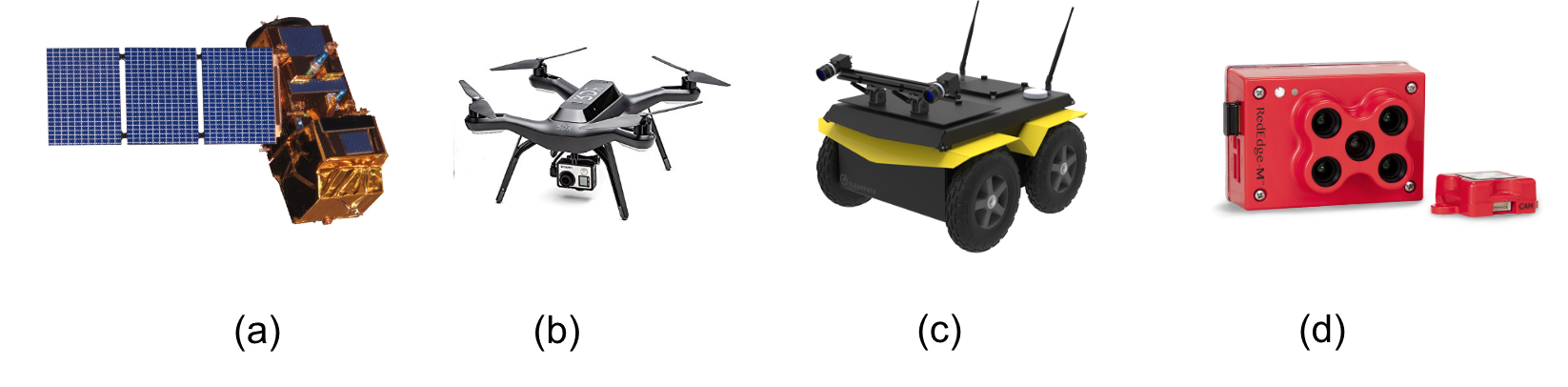}}
\caption{Available delivery mechanisms with sensor modalities: (a) Images from the Sentinal-II, (b) 3DR Solo, (c) ClearPath Jackal, and (d) Micasense Redenge (and others).}
\label{fig:modalities}
\end{figure}

In this work, I propose a novel aggregate framework for utilizing multispectral images and environmental metadata in heterogeneous teams of sensor modalities. In this precision agriculture application, I will introduce a framework that identifies the presence of phenotypically expressed crop stress. I label the framework the Agricultural Distributed Decision framework (ADDF). ADDF tackles individual and team learning and planning, as well as approaches for accomplishing individual tasks, representing the task of organizing satellites, autonomous uninhabited aerial vehicles (AUAVs), autonomous uninhabited ground vehicles (AUGVs), and ground-level environmental sensors, specified in Fig.~\ref{fig:modalities}.

Beyond sensor modalities, the environmental domain (used in Sec.~\ref{sec:results_image} and simulated in Sec.~\ref{sec:results_team}) is embodied by large-scale peanut crop fields in Tifton, GA, managed by the College of Agriculture and Environmental Sciences (CAES) at the University of Georgia, Tifton campus. Extant stresses include crop field erosion, damage done by local fauna, and several introduced stresses including fungus and pathogen introductions resulting in crop blight and lesions.

The functional capabilities of the sensor modalities are defined as follows:

\begin{itemize}
    \item \textbf{L3}: The highest level will be comprised of satellites which collect images of the crops on average every week. The Sentinel 2a and 2b satellites will be used for this layer of information. The data from these satellites are available for free and the ESA also provides a tool box for processing data collected. This level is passive, as it is not directly controlled, instead limited by a fixed temporal component of passing over the same location once every 5 days. Image data in this layer has a multi-spectral resolution of 10 meters per pixel  and a thermal resolution of 20 m per pixel.
    \item \textbf{L2}: Layer 2 represents the multispectral, high dimensional image data taken by AUAVs. While AUAVs have a very high degree of control and speed of execution with limited challenge in pathing, images are taken above their targets. Images generally have a resolution of 1-3 centimeters per pixel depending on the height of the AUAV and the spatial resolution of the camera.
    \item \textbf{L1}: Like L2, this layer represents a controllable physical subsystem, but is executed by AUGVs. Pathing is significantly more challenging and, even given high resolution geolocation sensors, may require exploration for the AUGV to reach its target. Though targeted exploration is more computationally costly and slower, image data can be collected from a variety of perspectives (individual leaves, fruit, flowers, whole plants) and have a similar resolution to L2, albeit from alternative angles.
    \item \textbf{L0}: Although often not directly measuring features due to the onset of stress, I include a layer representing an array of ground sensors which measure air and soil moisture, ground temperature, and relative humidity. This helps acquire information valuable for predictive learning approaches that enhance the data model beyond the other layers’ image data, enhancing model accuracy by capturing features that the resultant stresses are conditionally dependent on.
\end{itemize}

\begin{figure} [htb]
\centerline{\includegraphics[width=8cm]{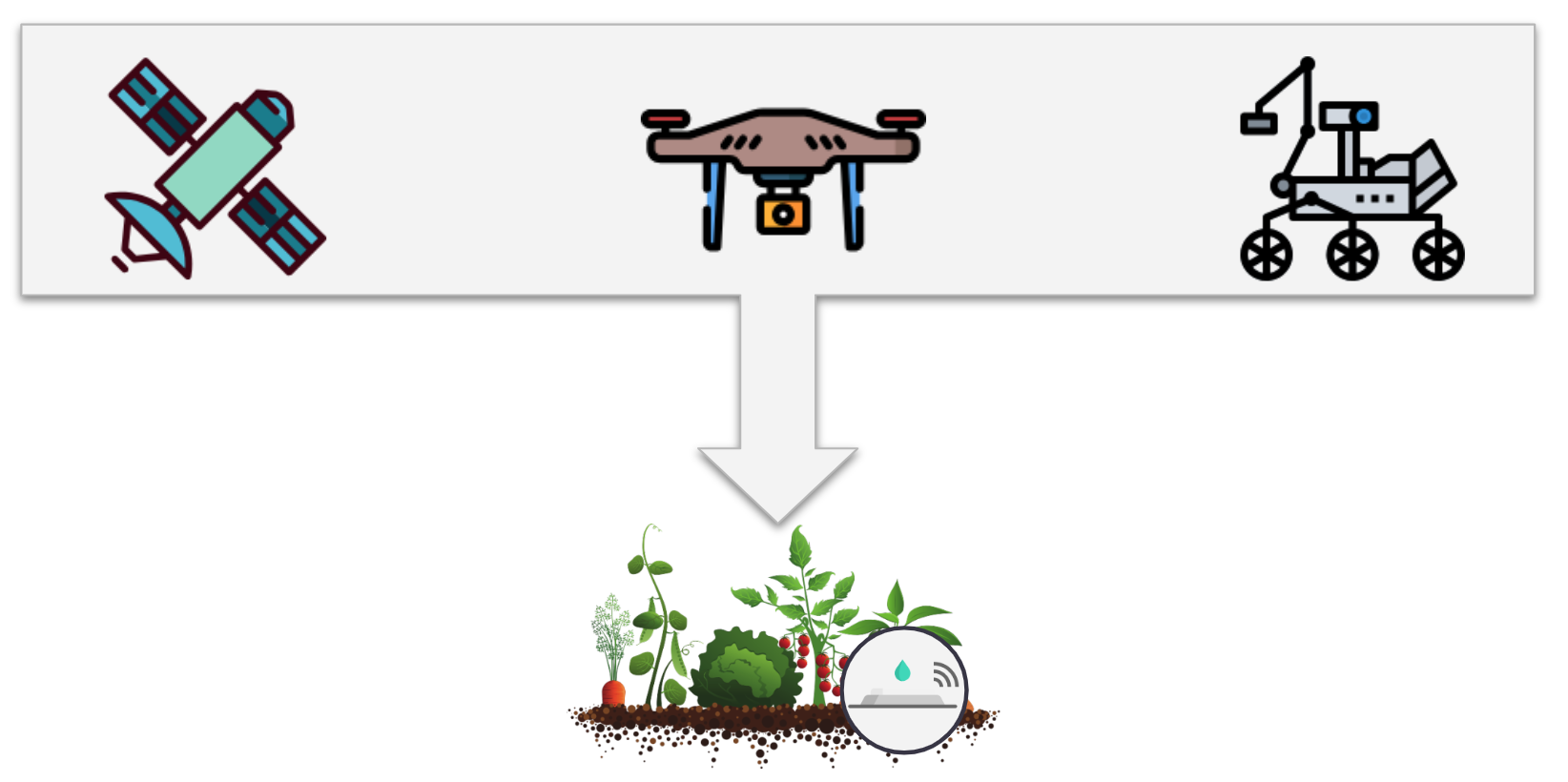}}
\caption{Arrangement of physical sensor modalities. Layers 3, 2, and 1 delegates to subsequent layers, and all layers take image data and collect environmental data from layer 0.}
\label{fig:domain_arrangement}
\end{figure}

As Fig.~\ref{fig:domain_arrangement} illustrates, layers 3, 2, and 1 collect image data at various resolutions and spectra of the crop field, while also collecting environmental data from the layer 0 sensors. Each layer is able to instantly communicate with a centralized server via a wireless network infrastructure. This centralized server then performs analysis and reasoning on the collected image and environmental data, described in the next section.

\section{Agricultural Distributed Decision Framework}
\label{sec:method}

The ADDF is comprised of two distinct functional capabilities: the processing of NDVI image data to compute the evolution of point-in-time crop growth metrics and the ability to learn when to raise a call-to-action, believing that processed image data indicates immediate or impending crop stress. The former capability requires collecting normalized difference vegetation index (NDVI) image data, computing a moving average pixel sufficient metric, and identifying deviating segments of crop fields. The latter capability introduces a reinforcement learning technique for exploring policies that, based on continuous-value observations, trigger a call-to-action. NDVI indices are computed by transforming ($nir$) and red wavelengths ($r$) as follows: $\frac{nir-r}{nir+r}\cdot 100$.

\subsection{Processing Image Data}
\label{sec:img}

Composing metrics of growth patterns via NDVI imaging is a popular methodology. While NDVI image data is effective at detecting stress in large and fully grown crops, it can struggle in smaller target crop sizes. Nonetheless, it is a highly effective mechanism. Unfortunately, much of the contemporary body of work leverages retrospective curve-fitting as a departure point. As the domain requires that immediate action be taken (i.e. at various points \textit{during} the growing season), I instead opt to develop a sufficient statistic computed from two or more NDVI images taken at potentially diverse intervals by proportionally merging error values.

Two distinct types of image processing using this methodology are required. The first task is simple: two aligned NDVI images, taken from highly similar perspectives (L3 and some L2 tasks), are compared via image difference with $n$-square-pixel approximation. Due to possible slight deviations of the images (due to erroneous alignment or evolution of crop size), the approximation, which takes the average index of an $p\times p$ square, can be used to alleviate information loss.

The second, more complex methodology required is analyzing images of crops from dramatically different angles, comprising some of L2's tasks and nearly all of L1's. In this case, I instead generate a distribution of the NDVI image coloration of a crop and compare the distribution to average coloration taken by the layer. In this work, I focus on the first task and the theory of composing a heterogeneous team, leaving this task and online experiments for future work.

\begin{figure} [htb]
\centerline{\includegraphics[width=8cm]{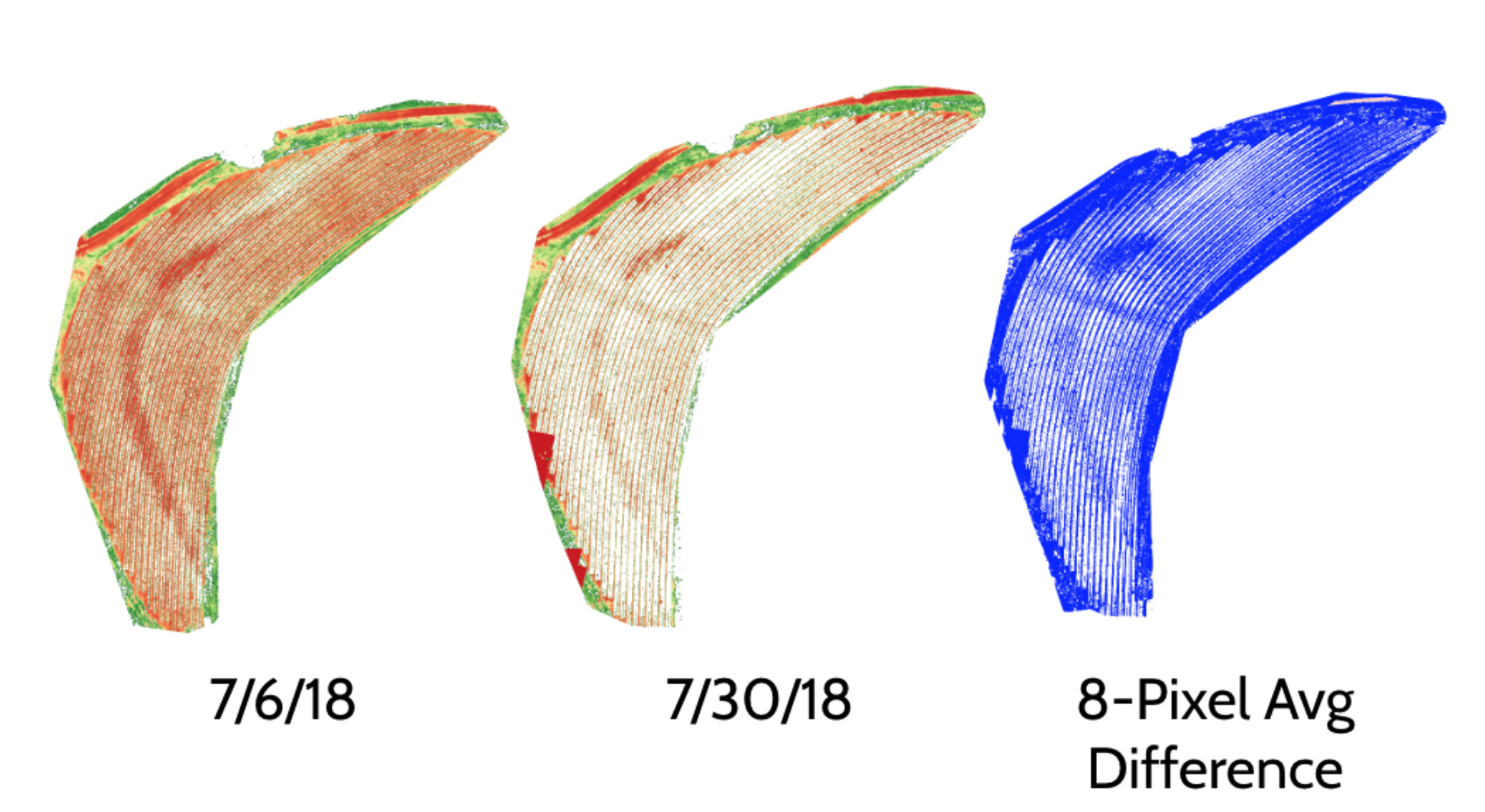}}
\caption{Comparison of two NDVI peanut field images taken over a 3 week period with $8\times 8$ pixel approximation.}
\label{fig:rmsd}
\end{figure}

Figure~\ref{fig:rmsd} shows a trivial implementation of a single pairwise comparison of two images taken 3 weeks apart of the domain's peanut field in the late 2017 growing season. Interestingly, the results are quite telling. Known to the field workers, three standout features are identifiable via this comparison: (1) the field is bisected by two lines as it was originally 4 fields, (2) near the bisection point significant crop erosion is occurring, and (3) the west, and particularly northwest, sector of the field is beset by damage from grazing deer.

\begin{algorithm}[!h]
  \caption{ADDF: NDVI image difference method}
  \label{alg:avg}
  \begin{algorithmic}[1]
    \REQUIRE Two NDVI images, $a, b$; image dimensions $x, y$; pixel approximation parameter $p$
    \STATE $i\gets 0$
    \STATE $j\gets 0$
    \STATE $diff\gets \emptyset$
    \WHILE{$i<x$}
    \WHILE{$j<y$}
    \STATE $a_{i,j}\gets$ average index of pixels $(i,j)$ to $(i+p,j+p)$ for image $a$
    \STATE $b_{i,j}\gets$ average index of pixels $(i,j)$ to $(i+p,j+p)$ for image $b$
    \STATE $diff_{i,j}\gets \min[(a_{i,j}-b_{i,j}),0]$
    \STATE $i\gets i+p$
    \STATE $j\gets j+p$
    \ENDWHILE
    \ENDWHILE
    \RETURN $diff$
  \end{algorithmic}
\end{algorithm}

Algorithm~\ref{alg:avg} annotates the relatively simplistic process of generating a difference sample between two NDVI images. The resultant matrix approximates the index across a $p$ square pixels, the set of which becomes a list of pairwise comparisons. I am interested in $p$-size sectors that, over the growing season, have very low variance, as it indicates that either (a) crops are not growing or are dying quickly, or (b), at the near infrared portion of the spectrum, low reflectance indicates low crop health from stress, such as from water deficiency.

\begin{algorithm}[!h]
  \caption{ADDF: Variance estimation}
  \label{alg:update}
  \begin{algorithmic}[1]
    \REQUIRE List of pairwise difference matrices $d$; individual diff matrix size $m,n$
    \STATE Variance matrix $v\gets m\times n$ matrix instantiated to $0$
    \IF{$|d|>1$}
    \FOR{$i$ in $[0,m]$; $j$ in $[0,n]$}
    \STATE $v_{i,j}=VAR\{d^k_{i,j}\}_{k=0}^{|d|}$, the variance of the approximate pixel at $i,j$ for all image differences
    \ENDFOR
    \STATE Normalize all values in $v$ between $0$ and $1$
    \ENDIF
    \RETURN $v$
  \end{algorithmic}
\end{algorithm}

As I am primarily interested in disproportionate changes in reflectance and growth, instead of examining pure average indexes across the pairwise differences of images, I focus on the variance of those images. Following the matrix generation in Alg.~\ref{alg:avg}, Alg.~\ref{alg:update} computes the variance across all pairwise differences and then normalizes relative to the highest variance. When analyzing the resultant variance matrix, lower values are more concerning than higher ones.

As a last step, since the resultant matrix serves as rudimentary image data where lower variance is represented as higher intensity pixels. In order to combat variance diffuseness, which is caused by crops separated by rows of soil, I crop the image to the field and apply a Gaussian blur where $\sigma = 2.5$. I then segment the image via straightforward applications of image segmentation algorithms, such as K-means~\cite{kmeans1,kmeans2}, to create afflicted sectors of the crop field. These sectors, based on their average variance, are then decision moments for the agents described in the following sections.

\subsection{Composing a Heterogeneous Team}
\label{sec:team}

I have 4 layers of sensor modalities, each with the capability to take either image or environmental data at varying resolutions. I am tasked with representing these layers, then, as a team, with the common goal of identifying a stress while balancing (a) speed of identification and (b) accuracy of the eventual categorization.

\begin{figure} [htb]
\centerline{\includegraphics[width=8cm]{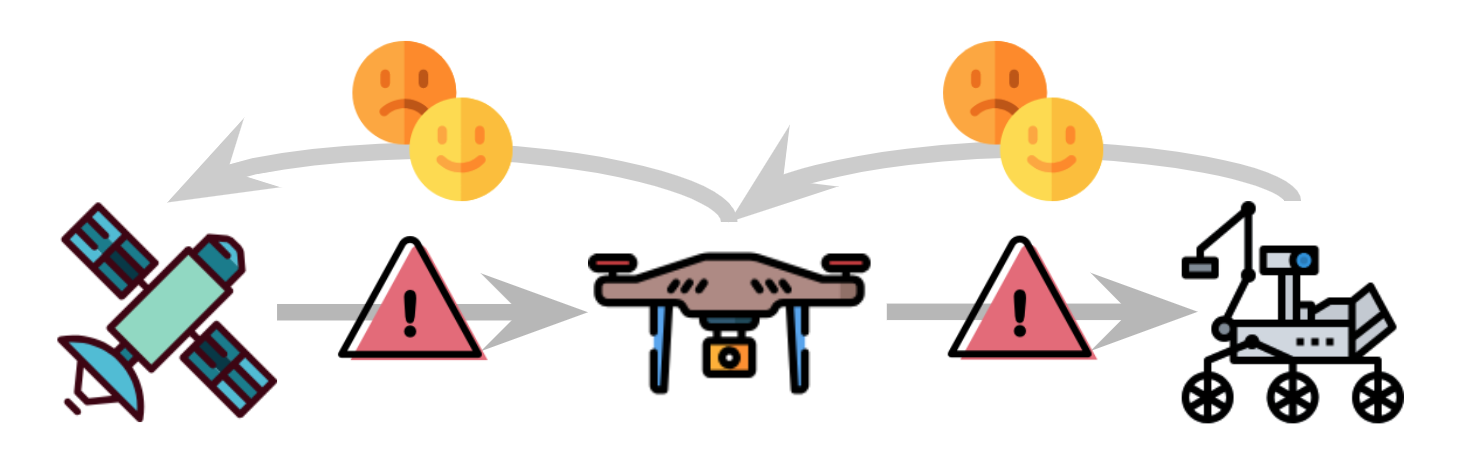}}
\caption{Image demonstrating the call-to-action and reinforcement process of layers 3, 2, and 1.}
\label{fig:rl}
\end{figure}

Casting this problem as decision-theoretic is rather straightforward. Each layer is solving an independent game formulation, though the state of the environment is (largely) identical for each of them. Since the eventual categorization is used to inform the decisions of higher level layers, I adopt the perspective of \textit{reinforcement learning}. Borrowing from Perkins' MCES-P, I explore a set of policies mapping call-to-actions to real-valued observations of the variance across images of a crop field. Figure~\ref{fig:rl} presents a visual representation of the actions of layers and the propagation of feedback.

I first present the general POMDP frame for layers 3, 2, and 1. The problem an individual agent faces is defined as a tuple $ADDF_i^L=\langle S,A,T,\Omega,O,R \rangle$, where $L$ and $i$ refers to the layer and agent respectively. $A^L$ refers to actions taken at this layer, and $A^{L-1}$ refers to the eventual action of the subsequent layer. Level 0 has no agency, represented in the framework as additional state observation information.

\begin{itemize}
\item $S$: the distribution of stress and agent location over a multi-row and column large-scale crop field
\item $A$: the set of actions, uniquely defined per layer
\begin{itemize}
    \item \textbf{L3}: take low-resolution images of the entire field on rare occasions
    \item \textbf{L2}: move; take image of a field or section
    \item \textbf{L1}: move; take image of an individual plants, leaves and fruit
\end{itemize}
\item $T=S\times A\times S$: state transition function dependent on agent movement
\item $\Omega$: the set of observations. Each agent receives information from level 0 and information as to how each sector/crop deviates from expected image data using Alg.~\ref{alg:update}. I discretize to levels of severity via clustering.
\item $O=S\times A$: the observation function, mapping observations to actions dependent on the state
\item $R=S\times A^{L}\times (A^{L-1})\rightarrow \mathcal{R}$: the reward function, dependent on the agents action and the decision made by the subsequent layer. The exception is at layer 1, which makes the final decision on the presence of a stress.
\end{itemize}

\begin{figure} [htb]
\centerline{\includegraphics[width=8cm]{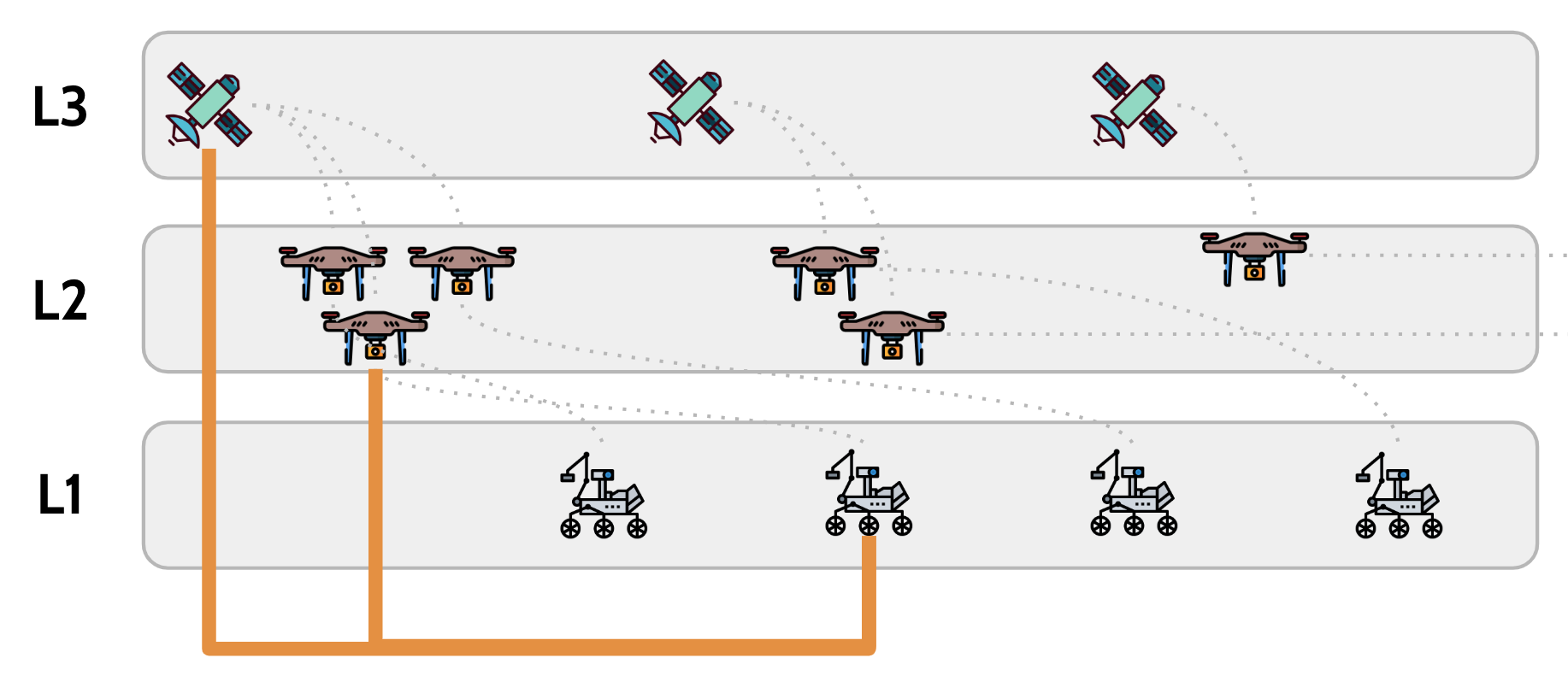}}
\caption{Layers 3 and 2 experience delayed reinforcements when a new game is started before subsequent layers make a decision.}
\label{fig:delayed}
\end{figure}

Examining the reward function $R$ raises an interesting caveat of the domain: reinforcements must be delayed due to relying on subsequent layer categorization, and games are played in parallel even within each layer. Since each game is a single horizon, and the policies that are learned are memory-less and reactive, this only means that games may be resolved outside the order they were played in. Figure~\ref{fig:delayed} demonstrates how a layer 3 agent may begin playing a game before a previous decision was reinforced.

Here I present the algorithms for ADDF. I generally require two flavors of ADDF: a high level agent that may tackle several sectors at once (such as for layer 3 and, sometimes, layer 2) and a lower level, sector- and crop-specific agent (layer 2 and 1). The high level agent creates multiple decision points for the lower agent, which often must tackle the sectors one-by-one.

\begin{algorithm}[!h]
  \caption{ADDF}
  \label{alg:addf}
  \begin{algorithmic}[1]
    \REQUIRE Maximum number of trajectories, $k$, to explore; sectors $S$; number of agents $Z$; observation and action spaces $O,A$ for each agent
    \FOR{$i\in (2,Z)$} \label{line:init}
    \STATE $Q^Li_{o,a}\gets 0$ for all $(O_i,A_i)$
    \STATE $c^Li_{o,a}\gets 0$ for all $(O_i,A_i)$
    \STATE $\pi^Li\gets $random initial policy
    \ENDFOR
    \REPEAT
    \STATE Generate call to actions $\vec{\tau}^{L3}=L3(\pi^{L3},k,S,c^{L3})$ \label{line:L3}
    \STATE Generate call to actions and collect rejections $\langle \vec{\tau}^{L2}, \vec{r}^{L2} \rangle =L2(\pi^{L2},k,S,c^{L2},\vec{\tau}^{L3})$ \label{line:L2}
    \STATE Collect classifications $\vec{r}^{L1} =L1(\vec{\tau}^{L2})$
    \STATE $\vec{r} = \vec{r}^{L1} + \vec{r}^{L2}$
    \STATE $Update(\vec{Q},\vec{c},\vec{\pi},\vec{r},\vec{\tau},Z)$ \label{line:learn}
    \UNTIL{End of growing season}
  \end{algorithmic}
\end{algorithm}

I begin by covering the high-level rotation of actions as a crop season progresses, defined in Alg.~\ref{alg:addf}. Line~\ref{line:init} initializes the Q-table, counts, and policies for each layer. Line~\ref{line:L3} runs the call-to-action generation for Layer 3, leveraging its current learned policy. Layer 2 is a bit different. While line~\ref{line:L2} also generates calls to action, any sector it considers without stress is served as stimuli for Layer 3. Layer 1 only generates stimuli. Line~\ref{line:learn} then updates all Q-values and transforms layer policies that have a new best action using reward stimuli.

As an important point, I omit the constraint of parallelism in the description of Alg.~\ref{alg:addf} for the purpose of brevity and clarity. Line~\ref{line:L2}, for example, doesn't occur every iteration. This is accomplished instead by creating a queue from line~\ref{line:L3} and iteratively executing line~\ref{line:L2} until the queue is empty, illustrated in Fig.~\ref{fig:delayed}. Line~\ref{line:learn} only executes when Layer 1 or 2 completes a task.

\begin{algorithm}[htbp]
  \caption{$L3(\pi,k,S,c)$}
  \label{alg:l3}
  \begin{algorithmic}[1]
    \REQUIRE Policy $\pi$; a maximum number of trajectories, $k$, to explore; vector of potentially stressed sectors $S$; counts $c$
    \STATE Choose random action $a$ and observation $o$
    \STATE Generate $\pi'$ as $\pi \gets (o, a)$
    \STATE Create variance samples $X$ from $S$ following Alg.~\ref{alg:update}
    \STATE Generate trajectories $\hat{\tau} = \{\tau_x\}_{x=0}^{|X|}$, according to $\pi'$
    \STATE Update counts $c_{o,a}\gets c_{o,a} + 1$
    \STATE Extract call-to-actions for subsequent layers, $\vec{\tau}=\{\tau \in \bar{\tau} : \tau_a > 0\}$ \label{line:extract}
    \RETURN $\vec{\tau}$
  \end{algorithmic}
\end{algorithm}

Algorithm~\ref{alg:l3} defines the L3 policy exploration and execution process. As in Alg.~\ref{alg:MCESP}, I select a random observation-action pair to explore, creating transformed policy $\pi'$. By taking variance samples of target sectors, L3 generates observations, which it then acts on using the transformed policies. For those actions that indicate a stress, L3 returns a call-to-action.

Since information must be passed between layers, I define $\tau$ differently than canonical \mcesp{}. Here, each element in $\tau^{Li}$ includes the observation, action, and sector index ($o,a,s$) of that layer, omitting rewards. When I return rewards back to preceding layers via $\vec{r}$, the preceding layer then knows which observation and action to update.

The algorithm for Layer 2 is omitted, as it differs from Layer 3 only in that it returns trajectories from $\vec{\tau}^{L3}$ as a negative reward if it fails to detect a stress in that sector, similar to Layer 1's line~\ref{line:reject} below, albeit only negative rewards.

\begin{algorithm}[!h]
  \caption{$L1(I)$}
  \label{alg:l1}
  \begin{algorithmic}[1]
    \REQUIRE Call-to-actions $I$
    \STATE $\vec{r}\gets\emptyset$
    \FOR{States from call-to-actions $s\in I$}
    \STATE Create variance samples $x$ from $i(s)$ following Alg.~\ref{alg:update}
    \STATE Classify non-uniformity $r=Classify(x)$ \label{line:classify}
    \STATE Dispense rewards $\vec{r} = APPEND(\vec{r},(s,r))$ \label{line:reject}
    \ENDFOR
    \RETURN $\vec{r}$
  \end{algorithmic}
\end{algorithm}

In this formulation, Layer 1 is considered objective in its classification, and generates stimuli for the other two (or more) layers. Like the previous layers, it generates an image of a particular crop, but if it detects significant non-uniformity, it immediately classifies the sector as stressed. 

\begin{algorithm}[!h]
  \caption{$Update(\vec{Q},\vec{c},\vec{\pi},\vec{r},\vec{\tau},Z,k)$}
  \label{alg:rlupdate}
  \begin{algorithmic}[1]
    \REQUIRE Q-tables $\vec{Q}$; count vectors $\vec{c}$; current policies $\vec{\pi}$; rewards $\vec{r}$; call-to-actions $\vec{\tau}$; number of agents $Z$; trajectory requirement count $k$
    \FOR{Agent $i\in (2,Z)$}
    \FOR{Agent's call-to-actions $t\in \vec{\tau}^{Li}$}
    \STATE Get reward for call-to-action $r = \vec{r}(t(s))$ \label{line:feedback}
    \STATE $Q^{Li}_{o, a} \gets (1-\alpha(c^{Li}_{o, a})) \cdot Q^{Li}_{o, a} +\alpha(c^{Li}_{o,a}) \cdot r$
    \IF {$\max_{a'} Q^{Li}_{o, a'} > Q^{Li}_{o, \pi^{Li}(o)} + \epsilon(k,c^{Li}_{o,a}, c^{Li}_{o,\pi^{Li}(o)})$}
    \STATE $\pi^{Li}(o) \gets a'$
    \STATE $c^{Li}_{o, a} \gets 0$ for all ${o,a}$
    \ENDIF
    \ENDFOR
    \ENDFOR
  \end{algorithmic}
\end{algorithm}

Each iteration is concluded with an update of the Q-table. As in MCESP, if the sample complexity $k$ is satisfied, then transformations are possible, if the agent has learned enough about each potential policy.

The domain differs from predominant exploration of POMDP domains in that the observation function is continuous. That is, I receive real-valued differentials between expected crop growth when processing image data. However, via utilizing lossless conversion to observation space partitions~\cite{hoey}, I can solve this continuous problem discretely, even considering layer 0 continuous-value environmental metadata. 

In layers that contain multiple agents (excluding layer 3 and 0), since the problem requires potentially reacting to multiple call to actions, I utilize a generalization of the POMDP that solves for instant-communication team play, the Multiagent POMDP~\cite{boutilier}. The MPOMDP is an interesting formulation, representing the dynamism of the decision problem by individually capturing the characteristics of each component of the system (in this case, the layers). However, it is well-known that an MPOMDP can be directly converted to an equally expressive POMDP and solved as such~\cite{amato}. The value is in its interpretability, which doesn’t affect the complexity to solve it.

I represent a layer’s homogeneous team as a tuple $ADDF^L=<Z,S,\vec{A},T,\Omega,O,R>$, with new or modified frame elements:
\begin{itemize}
\item $Z$: the set of agents
\item $\vec{A}={A^L_z}_{z=1}^Z$: the set of all agents’ actions
\item $O=S\times \vec{A}$: the team observation function, where the state and the team’s actions result in a single, shared observation
\end{itemize}

Utilizing a reinforcement learning perspective, I then learn the optimal policy for each layer as a Monte Carlo solution to the (M)POMDPs~\cite{thrun,perkins}.

\subsection{Tackling the Hyper-Conservative Local Optima}
\label{sec:heuristics}

One complication of the methodology is that layers are incentivized to be highly conservative in their call-to-actions. Since the presence of a stress is a rare event \textit{vis a vis} normal growth patterns, learning is inherently biased to rejecting the stress~\cite{rare_event}. Besides dramatically overweighting rewards that indicate stress, I propose two options: forced exploration even under rejection and random exploration in layer 1.

The first option considers reserving a few call-to-actions at each interval (demonstrated in Fig.~\ref{fig:delayed}) to create an objective for subsequent layers to explore near-accepted call-to-actions. For example, even if layer 3 rejects the notion of a stress in a particular sector, it will still inform layer 2 of a stress in that sector (along with other positively-identified stresses), a strategy similar to random policy exploration~\cite{wu13}. This may be added to Alg.~\ref{alg:l3} by adding rejected sectors to the call-to-actions $\tau$ with exponentially-decaying weight $w(\tau,m,i)=\frac{m}{m+|\tau|+i}$ for each $i$th rejected sector where $m$ controls the steepness of the exponential decay. This methodology is explored in Sec.~\ref{sec:results}.

The second option capitalizes on potential dead-time in the lower layers. When not exploring the crop field, I can opt to set layer 1 and 2 to an exploration mode, where they peruse the field randomly. When a stress is identified, a call-to-action may be simulated sequentially through the entire system and immediately rewarded appropriately, creating a positive sample to learn from. I reserve exploration of this methodology for real-world experiments in future work.

\section{Experiments and Results}
\label{sec:results}

I propose two toy implementations of the problem domains to test the approaches in Secs.~\ref{sec:img} and~\ref{sec:team}. I first examine the performance of the image segmentation technique for identifying stressed crop areas by utilizing actual AUAV images collected across several weeks of a growing season. I then introduce a two-layer POMDP implementation of the ADDF framework with a simulated toy environment, showing the effectiveness of learning a reward function \textit{vis a vis} a subsequent layer's decisions. In practice, Sec.~\ref{sec:results_image} produces inputs for Sec.~\ref{sec:results_team}, but I perform separate experiments in this work.

\subsection{Identifying Phenotypic Stress}
\label{sec:results_image}

Through technology supplied to the Department of Entomology at the University of Georgia, Tifton campus, I collected 5 images, each separated by 7 to 9 days, in July and August of 2017 of a peanut field. These images were collected by a 3DR Solo aerial drone indexed using NDVI at a resolution of 1-3 centimeters per pixel. The resultant files were encoded as Tagged Image File Format (TIFF), sized around 20 megabytes each.

Since these images took place in a previous growing season, the department is aware of several stresses that occurred during the growing season. First, the peanut field used to be four separate fields, bisected by two roads, and a sector just north of the east-west road suffered from stress due to soil erosion.

I experiment with several parameter settings for $p$, $\sigma$, and, in the case of K-means, $k$. Though I hand-selected parameters for the final result in this section, I hypothesize that performing gradient exploration methodologies, parameters could be fit by optimizing for fit in the final segmentation technique (such as the elbow method for K-means~\cite{elbow}).

\begin{figure}[!htb]
    \centering
    \begin{minipage}{.25\textwidth}
        \centering
        \includegraphics[width=0.9\textwidth]{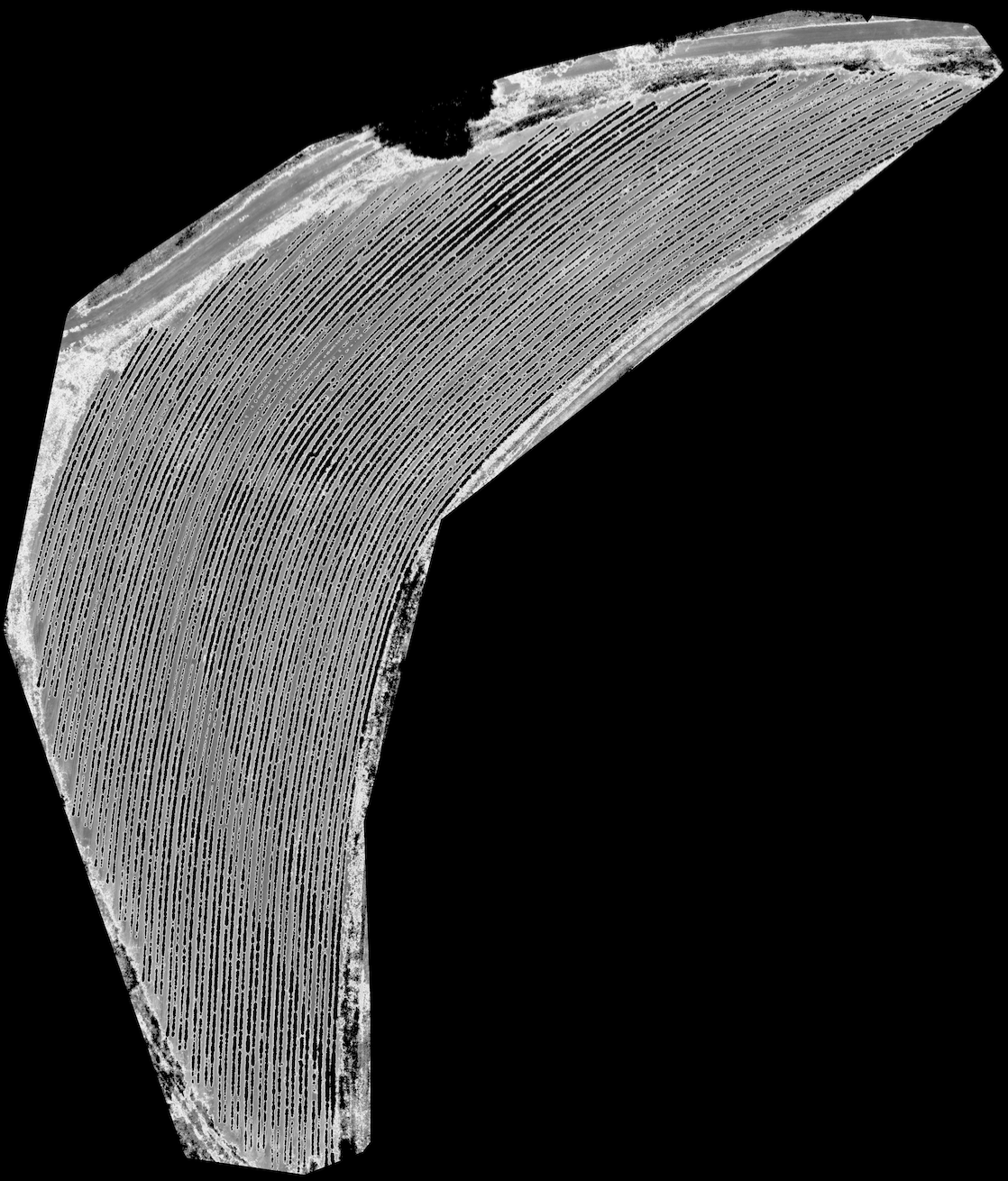}\\
        {$(a)$}
    \end{minipage}%
    \begin{minipage}{0.25\textwidth}
        \centering
        \includegraphics[width=0.9\textwidth]{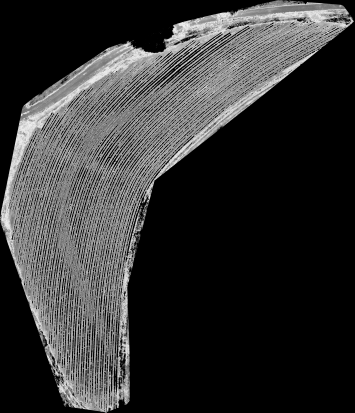}\\
        {$(b)$}
    \end{minipage}
    \caption{Illustration of applying Alg.~\ref{alg:avg} with $p>1$ to an original image $(a)$ resulting in an approximate image $(b)$.}
    \label{fig:approx}
\end{figure}

For Alg.~\ref{alg:avg}, I set $p=12$, converting original image dimensions of around $4900 \times 4200$ to $620 \times 530$. Figure~\ref{fig:approx} shows the effect of approximation, resulting in significantly less diffuseness between crop rows, though some is retained. Since the result of Gaussian blurring can lose too much information to compare between images, I wait until the final comparative image before applying it. Next, I compared the differences between two subsequent crop field images.

\begin{figure}[!htb]
    \centering
    \begin{minipage}{0.25\textwidth}
        \centering
        \includegraphics[width=0.7\textwidth]{figs/approx1}\\
        {$(a)$}
    \end{minipage}%
    \begin{minipage}{0.25\textwidth}
        \centering
        \includegraphics[width=0.7\textwidth]{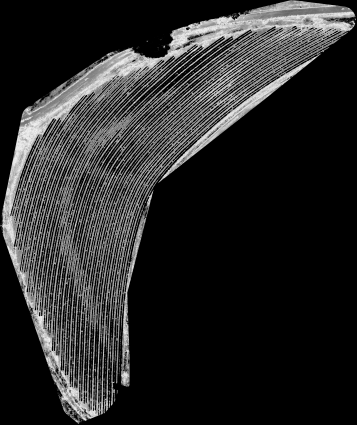}\\
        {$(b)$}
    \end{minipage}
    \begin{minipage}{0.01\textwidth}
        $\Large{\rightarrow}$
    \end{minipage}
    \begin{minipage}{0.25\textwidth}
        \centering
        \includegraphics[width=0.7\textwidth]{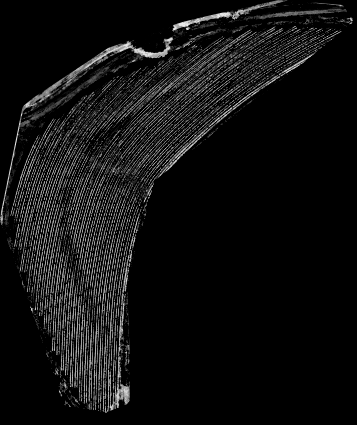}\\
        {$(c)$}
    \end{minipage}
    \caption{Result after full completion of Alg.~\ref{alg:avg}, where the difference of two subsequent images ($(a)$ July 6th and $(b)$ July 13th) is computed, resulting in $(c)$.}
    \label{fig:diff}
\end{figure}

In Fig.~\ref{fig:diff}, I illustrate the result of the first difference computation ($diff$) performed in the sample set for the first two dates. It confirms what I know about this data set: the north side of the field (which is darker, indicating less change) is performing significantly worse than the rest of the field, and roads bisecting the field are clearly visible.

\begin{figure}[!htb]
    \centering
    \begin{minipage}{0.25\textwidth}
        \centering
        \includegraphics[width=0.7\textwidth]{figs/diff1}\\
        {$(a)$}
    \end{minipage}%
    \begin{minipage}{0.25\textwidth}
        \centering
        \includegraphics[width=0.7\textwidth]{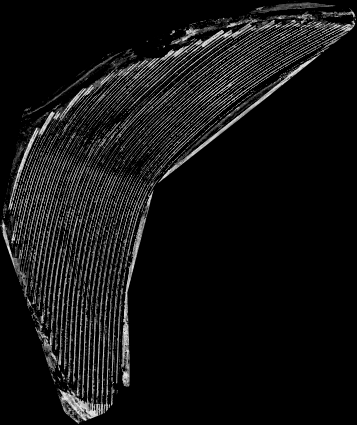}\\
        {$(b)$}
    \end{minipage}
    \begin{minipage}{0.01\textwidth}
        $\Large{\rightarrow}$
    \end{minipage}
    \begin{minipage}{0.25\textwidth}
        \centering
        \includegraphics[width=0.7\textwidth]{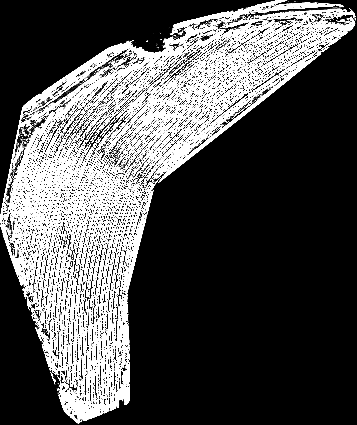}\\
        {$(c)$}
    \end{minipage}
    \caption{Calculating the variance between differences computed for $(a)$ July 6th and 13th and $(b)$ July 13th and 21st, resulting in variance map $(c)$.}
    \label{fig:var}
\end{figure}

\begin{figure}[!htb]
    \centering
    \begin{minipage}{0.25\textwidth}
        \centering
        \includegraphics[width=0.7\textwidth]{figs/var}\\
        {$(a)$}
    \end{minipage}%
    \begin{minipage}{0.01\textwidth}
        $\Large{\rightarrow}$
    \end{minipage}
    \begin{minipage}{0.25\textwidth}
        \centering
        \includegraphics[width=0.7\textwidth]{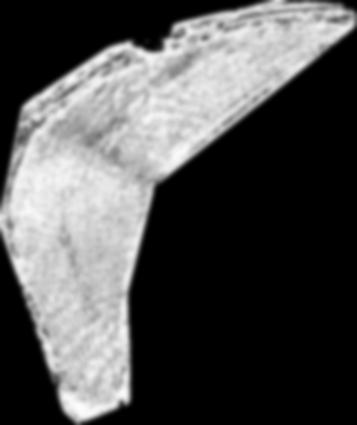}\\
        {$(b)$}
    \end{minipage}
    \begin{minipage}{0.01\textwidth}
        $\Large{\rightarrow}$
    \end{minipage}
    \begin{minipage}{0.25\textwidth}
        \centering
        \includegraphics[width=0.7\textwidth]{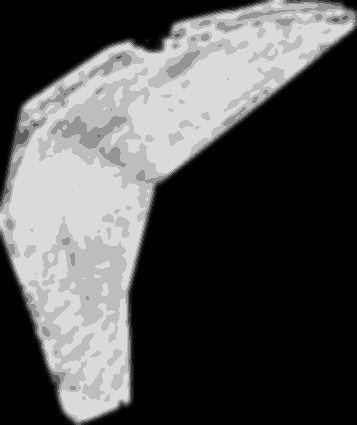}\\
        {$(c)$}
    \end{minipage}
    \caption{To identify affected sectors, I take an $(a)$ variance image, apply $(b)$ Gaussian blur, and then $(c)$ K-means.}
    \label{fig:kmeans}
\end{figure}

Examining the images in Fig.~\ref{fig:var} corroborates the growing concern from $diff$s computed on July 13th and 21st that the middle mass of the crop field is not producing. One drawback of this methodology is that the stress is certainly visible by the $diff$ on July 13th (recovering slightly by the 21st). After applying Alg.~\ref{alg:update}, since the variance image is still too diffuse for effective image segmentation, I apply a Gaussian blur and then segment, demonstrated in Fig.~\ref{fig:kmeans}.

Completing the approximation and segmenting technique then results in highly defined and contoured areas, show in Fig.~\ref{fig:kmeans}$.c$ for July 21st. 3 distinct layers are clearly visible (though the number of actual layers are determined by the input $k$ for the K-means method, $10$ in the experiments). The center road and previously mentioned erosion due north of the center of mass are both illustrated as very severe, with less severe sectors in the south and northeast.

\begin{figure}[!htb]
    \centering
    \begin{minipage}{0.25\textwidth}
        \centering
        \includegraphics[width=0.7\textwidth]{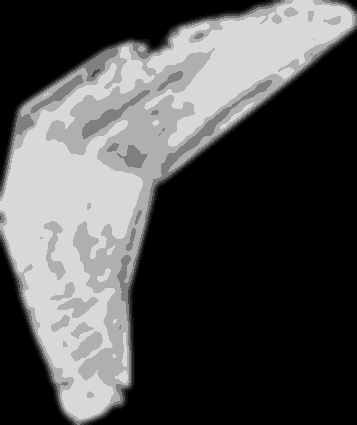}\\
        {$(a)$}
    \end{minipage}
    \begin{minipage}{0.25\textwidth}
        \centering
        \includegraphics[width=0.7\textwidth]{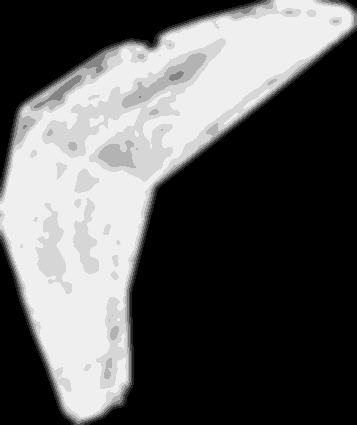}\\
        {$(b)$}
    \end{minipage}
    \caption{K-means processing for $(a)$ July 30th and $(b)$ August 11th}
    \label{fig:final_images}
\end{figure}

I complete this section by illustrating the K-means processing for the final two dates in the test set. Over the course of the first month and a half, I see that growth is stymied after the first completion and the problem worsens in the north side of the field, though it improves in the south. By the last date, August 11th, the field is significantly healthier, though the most severe problem areas from July 13th persist.

\subsection{Solving Inter-Team Objectives}
\label{sec:results_team}

\begin{figure} [htb]
\centerline{\includegraphics[width=8cm]{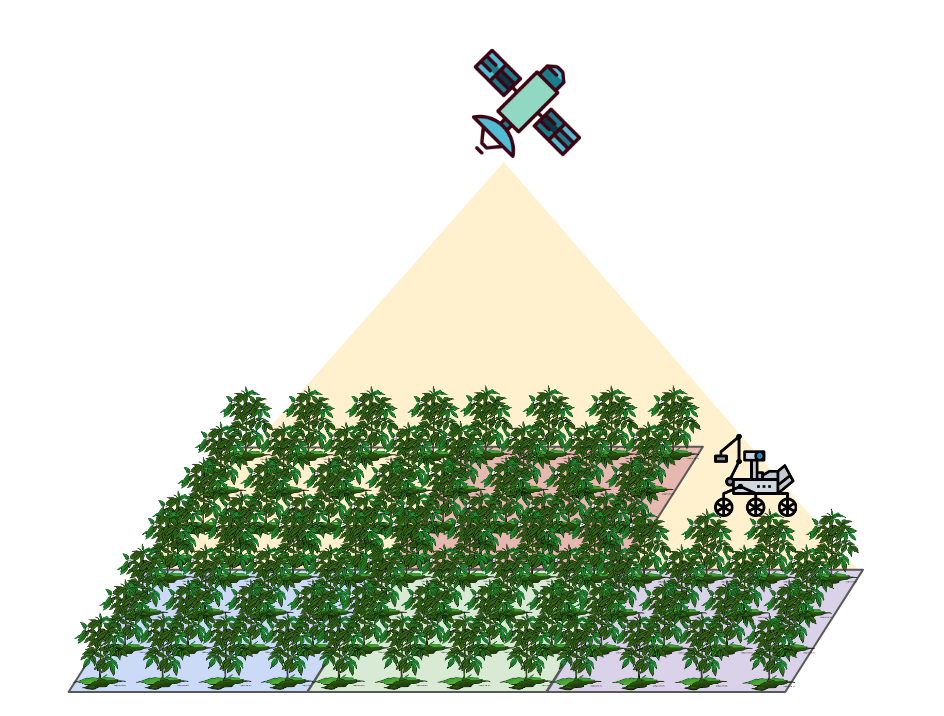}}
\caption{The toy crop field problem, where a fast and slow agent team up to learn crop stress probabilities.}
\label{fig:problem_domain}
\end{figure}

I test the effectiveness of the ADDF algorithm in a toy implementation of the crop field problem. In the experiment, two agents form a two-layer team, where the first agent (referred to as the "fast" agent) takes a low-precision image of an entire crop field, which contains 5 sectors, once every 3 days. The second, "slow" agent can collect image data only for a single sector, but can do it once a day. To simulate a L1 classifier, I use an oracle that returns the true state of the sector after each agent acts on a sector. Each agent receives one of $|O|$ observations, each correlating to a confidence of the stress, from low to high. I illustrate the domain in Fig.~\ref{fig:problem_domain}, where the fast and slow agents are represented by a satellite and AUGV, respectively.

The simulator generates a stress in each sector at the beginning of a growing season, which has $89$ days, with a $50\%$ probability. Early in the season, the stress of the state is unstable, having a $50\%$ chance to change status. This likelihood decreases exponentially each subsequent day. The fast agent gets the maximally correlated observation of the true stress of the environment (either $o=0$ or $o=|O|$ for the lack or presence of a stress, respectively) with $70\%$ probability, while the slow agent receives it with $85\%$ probability. Incorrect observations are received with the remaining probability with exponential weight towards the correct classification. For example, the slow agent's observations are received with probability $\{o0=0.8, o1=0.1, o2=0.05\}$ when no stress is present.

I test ADDF with configurations varying the number of observations and the inclusion of the heuristic in Sec.~\ref{sec:heuristics}, which adds rejected sectors to call-to-actions with exponentially decaying probabilities. Each configuration is tested with 500 seasons. I begin by setting the baseline with a canonical Q-learning technique, which does not perform policy exploration and always exploits the current highest action-value for an observation.

\begin{table}[h]
	\centering
	\scalebox{1}{%
		\begin{tabular}{|c|c|c|c|c|c|c|c|}
			\hline
			\multirow{3}{*}{\textbf{$|O|$}} & \multirow{3}{*}{\textbf{Agent}} & \multicolumn{5}{c|}{\textbf{Accuracy}} \\
			\cline{3-7}
			& & \multicolumn{2}{c|}{\textbf{True}} & \multicolumn{2}{c|}{\textbf{False}}  & \multirow{2}{*}{\textbf{Overall}}\\
			\cline{3-6}
			& & \textbf{Positive} & \textbf{Negative} &  \textbf{Positive} & \textbf{Negative} &\\
			\hline \hline
			\multirow{2}{*}{\textbf{3}} & \textbf{Fast} & 5 & 37,416 & 2 & 37,577 & $49.9\%$\\
			\cline{2-7}
			& \textbf{Slow} & 2 & 3 & 1 & 1 & \st{$71.4\%$} \\
			\hline
			\multirow{2}{*}{\textbf{5}} & \textbf{Fast} & 64 & 37,695 & 3 & 37,238 & $50.3\%$\\
			\cline{2-7}
			& \textbf{Slow} & 35 & 4 & 1 & 27 & \st{$56.3\%$} \\
			\hline
		\end{tabular}
	}
	\caption{Q-learning baseline, varying the observation and sector counts.}
	\label{tbl:qbaseline}
	\vspace{-0.9em}
\end{table}

As noted in Sec.~\ref{sec:heuristics}, the relative rarity of positive stimuli for negative events causes Q-learning to quickly converge to always rejecting the presence of a stress in a sector. Therefore, Q-learning performs essentially the same as the random baseline. With almost no samples to learn from, I strike the essentially random accuracy of the slow agent. I then test ADDF in this domain with trajectory limit $k=500$.

\begin{table}[h]
	\centering
	\scalebox{1}{%
		\begin{tabular}{|c|c|c|c|c|c|c|c|}
			\hline
			\multirow{3}{*}{\textbf{$|O|$}} & \multirow{3}{*}{\textbf{Agent}} & \multicolumn{5}{c|}{\textbf{Accuracy}} \\
			\cline{3-7}
			& & \multicolumn{2}{c|}{\textbf{True}} & \multicolumn{2}{c|}{\textbf{False}}  & \multirow{2}{*}{\textbf{Overall}}\\
			\cline{3-6}
			& & \textbf{Positive} & \textbf{Negative} &  \textbf{Positive} & \textbf{Negative} &\\
			\hline \hline
			\multirow{2}{*}{\textbf{3}} & \textbf{Fast} & 26,068 & 34,032 & 3,671 & 11,229 & $80.1\%$\\
			\cline{2-7}
			& \textbf{Slow} & 13,067 & 11,895 & 4,068 & 709 & $83.9\%$ \\
			\hline
			\multirow{2}{*}{\textbf{5}} & \textbf{Fast} & 20,756 & 36,737 & 3,732 & 14,775 & $75.6\%$\\
			\cline{2-7}
			& \textbf{Slow} & 12,052 & 7,612 & 701 & 4,123 & $80.3\%$ \\
			\hline
		\end{tabular}
	}
	\caption{ADDF with $k=500$.}
	\label{tbl:mcesp}
	\vspace{-0.9em}
\end{table}

ADDF performs dramatically better than the Q-learning baseline, achieving over $80\%$ accuracy for the slow agent and 3-obervation fast agent, performing relatively worse with more observations for the fast agent. The fast agent performs comparatively worse than the slow agent likely due to the increased noise in its observation function. These numbers are quite close to the theoretical maximum, considering noise for the environment.

Both the baseline and ADDF can potentially benefit from increased workload. The baseline rejects nearly every sector, and, referring to Tbl.~\ref{tbl:mcesp}, it is clear the slow agent is not given enough decision points. For example, when $|O|=3$, the fast agent generates just under 60 calls to the slow agent per season, when it can work 90. Without lowest-level agents working every day, many stressed crops will not be identified. I show the results for both the baseline and MCESP with the workload heuristic in Sec.~\ref{sec:heuristics} with $m=5$.

\begin{table}[h]
	\centering
	\scalebox{1}{%
		\begin{tabular}{|c|c|c|c|c|c|c|c|}
			\hline
			\multirow{3}{*}{\textbf{Method}} & \multirow{3}{*}{\textbf{$|O|$}} & \multirow{3}{*}{\textbf{Agent}} & \multicolumn{5}{c|}{\textbf{Accuracy}} \\
			\cline{4-8}
			& & & \multicolumn{2}{c|}{\textbf{True}} & \multicolumn{2}{c|}{\textbf{False}}  & \multirow{2}{*}{\textbf{Overall}}\\
			\cline{4-7}
			& & & \textbf{Positive} & \textbf{Negative} &  \textbf{Positive} & \textbf{Negative} &\\
			
			\hline \hline
			
			\multirow{4}{*}{\rotatebox{90}{\textbf{Baseline}}} & \multirow{2}{*}{\textbf{3}} & \textbf{Fast} & 30,552 & 26,445 & 14,202 & 3,801 & $76\%$ \\
			\cline{3-8}
			& & \textbf{Slow} & 22,238 & 13,629 & 5,110 & 5,023 & $78\%$ \\
			\cline{2-8}
			& \multirow{2}{*}{\textbf{5}} & \textbf{Fast} & 25,249 & 28,579 & 12,067 & 9,105 & $71.8\%$ \\
			\cline{3-8}
			& & \textbf{Slow} & 17,061 & 15,549 & 4,927 & 7,463 & $72.5\%$ \\
			\hline \hline
			\multirow{4}{*}{\rotatebox{90}{\textbf{ADDF}}} & \multirow{2}{*}{\textbf{3}} & \textbf{Fast} & 30,703 & 31,355 & 2,327 & 10,615 & $82.7\%$ \\
			\cline{3-8}
			& & \textbf{Slow} & 24,260 & 13,488 & 4,936 & 1,415 & $83.9\%$ \\
			\cline{2-8}
			& \multirow{2}{*}{\textbf{5}} & \textbf{Fast} & 28,025 & 31,995 & 6,239 & 8,741 & $80\%$ \\
			\cline{3-8}
			& & \textbf{Slow} & 17,865 & 18,242 & 3,991 & 4,898 & $80.2\%$ \\
			\hline
		\end{tabular}
	}
	\caption{The baseline and ADDF algorithms utilizing increased workload heuristics to employ the slow agent more frequently.}
	\label{tbl:heuristic}
	\vspace{-0.9em}
\end{table}

Table~\ref{tbl:heuristic} shows a dramatic performance boost for the baseline but, much more importantly, additionally demonstrates the slow agent working nearly every single available day. While ADDF still outperforms the baseline, it does not show remarkable improvement over the non-heuristic version. However, this isn't the major contribution of the heuristic. The impact is that, since the slow agent works more days, it identifies up to $83\%$ more stressed crops.

\section{Concluding Remarks}
\label{sec:conclusion}

In this chapter, I introduced an ensemble machine learning approach that learns to identify the onset of crop stress by employing a team of multiple sensor modalities to collect images, process them, and learn to identify crop stress via a decision theoretic approach optimized via reinforcement learning. The algorithm, dubbed the Agricultural Distributed Decision Framework (ADDF), takes multispectral, including near infrared, image data, computes the pairwise difference between two temporally successive images, uses the variance between differences to generate observations, and then learns whether the observation indicates a stress via a multi-layer POMDP approach. In this approach, high layers prevent low layers from intervening on sectors with low likelihood of stress, optimizing instead for sectors with high likelihood.

ADDF is derived from \mcesp{}, and each layer represents a sensor modality, including satellites, AUAVs, and AUGVs, with the observation space defined as the level of variance and ground sensor data. I show that \mcesp{} dramatically outperforms a canonical reinforcement learning technique, Q-learning, using over $75,000$ decision points to inform our results. Even when using a heuristic that allows subsequent layers to periodically act without previous layers' positive identification, ADDF is a better approach than the baseline.

\newpage
Several avenues exist for expanding the prototypical implementation we presented in this work. While I simulate layer 0 information in Sec.~\ref{sec:results_team}, we do not explore the ramifications it may have on image processing in Sec.~\ref{sec:results_image}. Additionally, layer 1 image processing is a completely different, and far more complex, task than the algorithms presented in Sec.~\ref{sec:img}. However, even in real-world experiments, the tasks accomplished by layer 1 may be readily replicated by a plant pathologist.

\newpage
\appendix
\chapter{Appendix}
\label{chap:appendix}

\section{Proof of Thm.~\ref{thm:mcesppacmas}}
\label{sec:mcesppacmas_proof}
\mcesppac{} in the multiagent setting allows for a PAC-style guarantee of $\epsilon$-local optimality. To show that the total error for \mcesppac{} is bounded by the user-defined $\epsilon$ with probability $1-\delta$, I first define the types of errors that can occur in selecting dominating neighboring policies and terminating when none is found after sampling. In this respect, our proof follows that of Greiner~\cite{PALO}.

We define $\mathcal{N}(\pi)$ as the set of neighboring policies of $\pi$. A policy is considered a neighbor if it differs from $\pi$ by only one action for all observation sequences.
\begin{enumerate}
\item After seeing $p$ samples (where $p<k$), \mcesppac{} selects some $\pi' \in \mathcal{N}(\pi)$, as $\pi'$ appears higher value than $\pi$, but it is not
\item After seeing $p$ samples (where $p<k$), \mcesppac{} cannot find a $\pi' \in \mathcal{N}(\pi)$ where $\pi'$ appears higher value than $\pi$, but there is one
\item After seeing all $k_m$ samples, \mcesppac{} selects some $\pi' \in \mathcal{N}(\pi)$, as $\pi'$ appears higher value than $\pi$, but it is not
\item After seeing all $k_m$ samples, \mcesppac{} cannot find a $\pi' \in \mathcal{N}(\pi)$ where $\pi'$ appears higher value than $\pi$, but there is one
\end{enumerate}

Recall $\epsilon$ and $k_m$ for \mcesppac{} are as follows.
\begin{align*}
	\epsilon(m,p,q,k_m) = \left\{
	\begin{array}{lr}
	\Lambda(\pi,\pi')\sqrt{\frac{1}{2p}{\ln}\frac{2(k_m-1)N}{\delta_m}} & \text{if } p=q < k_m\\
	\frac{\epsilon}{2} & \text{if } p=q=k_m \\
	+\infty & \text{otherwise}
	\end{array}
	\right.
\end{align*}
\begin{align*}
k_m = \left\lceil 2\frac{(\Lambda(\pi_i,\pi'_i))^2}{\epsilon^2} \ln\frac{2N}{\delta_m}\right\rceil
\end{align*}

Also, let $E[\pi]$ be the reward of the policy $\pi$, in the context of other agents and the environment, which the Q-value approximates.
\begin{align*}
a_{m}^n = &Pr\left[\exists \pi'_i \in \mathcal{N}(\pi^m_i) : (Q^{\pi'_i} - Q^{\pi^m_i}) \geq \epsilon(m,c^{\pi'_i}_i,c^{\pi^m_i}_i)  \text{ and }E[\pi'_i]<E[\pi^m_i]\right]\\
b_{m}^n = &Pr\left[\exists \pi'_i \in \mathcal{N}(\pi^m_i) : (Q^{\pi'_i} - Q^{\pi^m_i}) < \epsilon - \epsilon(m,c^{\pi'_i}_i,c^{\pi^m_i}_i) 
\text{ and }E[\pi'_i]>E[\pi^m_i]+\epsilon\right]\\
c_{m} = &Pr\left[\exists \pi'_i \in \mathcal{N}(\pi^m_i) : (Q^{\pi'_i} - Q^{\pi^m_i}) \geq \frac{\epsilon}{2} \text{ and }E[\pi'_i]<E[\pi^m_i]\right]\\
d_{m} = &Pr\left[\exists \pi'_i \in \mathcal{N}(\pi^m_i) : (Q^{\pi'_i} - Q^{\pi^m_i}) < \frac{\epsilon}{2} \text{ and }E[\pi'_i]>E[\pi^m_i]+\epsilon\right]
\end{align*}

We represent each of the probabilities above as disjoint sets over neighbors and the agents' policies. For example, $\mathcal{N}^<_i(\pi^m_i) = \{\pi_i\in \mathcal{N}(\pi^m)|E[\pi'_i]<E[\pi^m_i]\}$.

\begin{align*}
&a_{m}^n = Pr\Big[\bigvee_{\pi'_i\in \mathcal{N}^<_i(\pi^m_i)}(Q^{\pi'_i} - Q^{\pi^m_i}) \geq \epsilon(m,c^{\pi'_i}_i,c^{\pi^m_i}_i)\Big]\\
&b_{m}^n = Pr\Big[\bigvee_{\pi'_i\in \mathcal{N}^>_i(\pi^m_i)}(Q^{\pi'_i} - Q^{\pi^m_i}) < \epsilon - \epsilon(m,c^{\pi'_i}_i,c^{\pi^m_i}_i)\Big]\\
&c_{m} = Pr\Big[\bigvee_{\pi'_i\in \mathcal{N}^<_i(\pi^m_i)}(Q^{\pi'_i} - Q^{\pi^m_i}) \geq \frac{\epsilon}{2}\Big]\\
&d_{m} = Pr\Big[\bigvee_{\pi'_i\in \mathcal{N}^>_i(\pi^m_i)}(Q^{\pi'_i} - Q^{\pi^m_i}) < \frac{\epsilon}{2}\Big]
\end{align*}

Considering every possible $\pi'$ in $\pi^m$'s neighborhood leads to the following summation.
\begin{align*}
& a^n_{m} {\leq \sum\limits_{\pi'_i\in\mathcal{N}^<_i(\pi^m_i)}Pr[(Q^{\pi'_i}-Q^{\pi^m_i}) \geq \epsilon(m,c^{\pi'_i}_i,c^{\pi^m}_i,k_m)]}\\
&\leq \sum\limits_{\pi'_i\in\mathcal{N}^<_i(\pi^m_i)}Pr[(Q^{\pi'_i}-Q^{\pi^m_i}) \geq (E[\pi'_i]-E[\pi^m_i])+\epsilon(m,c^{\pi'_i}_i,c^{\pi^m}_i)]\\
&\leq {\sum\limits_{\pi'_i\in\mathcal{N}^<_i(\pi^m_i)}exp\left\{-2{p}\left(\frac{\epsilon(m,c^{\pi'_i}_i,c^{\pi^m_i}_i)}{\Lambda(\pi'_i,\pi^m_i,\pi_j)}\right)^2\right\} \numberthis \label{lab:eps1}}\\
&= \frac{|\mathcal{N}^<_i(\pi^m_i)|\delta_m}{2(k_m-1)|\mathcal{N}(\pi^m_i)|}
\end{align*}

Equation~\ref{lab:eps1} follows from Hoeffding's Inequality, where $(Q^{\pi'_i}-Q^{\pi^m_i})$ is the sample average of $E[\pi'_i]-E[\pi^m_i]$. I reduce \ref{lab:eps1} by utilizing $\epsilon$ when $p=q<k_m$. $b^n_m$ follows, 
\begin{align*}
\begin{split}
b^n_{m} &\leq \sum\limits_{\pi'_i\in\mathcal{N}^>_i(\pi^m_i)}Pr\left[(Q^{\pi'_i}-Q^{\pi^m_i}) < \epsilon - \right.\left.\epsilon(m,c^{\pi'_i}_i,c^{\pi^m}_i,k_m)\right] \leq \frac{|\mathcal{N}^>_i(\pi^m_i)|\delta_m}{2(k_m-1)|\mathcal{N}(\pi^m_i)|}
\end{split}
\end{align*}

The other two error types, where $p=k_m$, follow similarly but substitute $k_m$ for $n$.
\begin{align}
& c_{m} \leq \sum\limits_{\pi'_i\in\mathcal{N}^<_i(\pi^m_i)}Pr\left[(Q^{\pi'_i}-Q^{\pi^m_i}) \geq \frac{\epsilon}{2}\right]\\
& \leq \sum\limits_{\pi'_i\in\mathcal{N}^<_i(\pi^m_i)}exp\left\{-2k_m\left(\frac{\epsilon/2}{\Lambda(\pi'_i,\pi^m_i,\pi_j)}\right)^2\right\} 
 = \frac{|\mathcal{N}^<_i(\pi^m_i)|\delta_m}{2|\mathcal{N}(\pi^m_i)|} 
\label{eq:c2}
\end{align}
where Eq.~\ref{eq:c2} is obtained by substituting $k_m$ with its derived value in the previous expression and reducing.

\begin{align*}
d_{m,\pi_j} &\leq \sum\limits_{\pi'_i\in\mathcal{N}^>_i(\pi^m_i)}Pr\left[(Q^{\pi'_i}-Q^{\pi^m_i}) < \frac{\epsilon}{2}\right]
\leq \frac{|\mathcal{N}^>_i(\pi^m_i)|\delta_m}{2|\mathcal{N}(\pi^m_i)|}
\end{align*}

The sum total of the error is as follows.
\begin{align*}
&\sum\limits_{n=1}^{k_m-1}[a_m^n+b_m^n]+c_m+d_m\\
&=\sum\limits_{n=1}^{k_m-1}\Bigg[\frac{|\mathcal{N}^<_i(\pi^m_i)+|\mathcal{N}^>_i(\pi^m_i)|}{|\mathcal{N}(\pi_i^m)|}\frac{\delta_m}{2(k_m-1)}\Bigg]  +\frac{|\mathcal{N}^<_i(\pi^m_i)|+|\mathcal{N}^>_i(\pi^m_i)|}{\mathcal{N}(\pi_i^m)|}\frac{\delta_m}{2}=\delta_m
\end{align*}

Over all possible transformations, the total error is bounded by $\delta$.
\begin{align*}
\sum\limits_{m=1}^{\infty}\delta_m=\sum\limits_{m=1}^{\infty}\frac{6\delta}{m^2\pi^2}=\frac{6\delta}{\pi^2}\sum\limits_{m=1}^{\infty}\frac{1}{m^2}=\frac{6\delta}{\pi^2}\frac{\pi^2}{6}=\delta
\end{align*}

\section{Proof for Thm.~\ref{thm:mcesmppac}}
\label{sec:mcesmppac_proof}
The proof of $\epsilon$-local optimality for \mcesmppac{} strongly parallels that for \mcesppac{}. However, in the multiagent case, non-representative samples can cause multiplicatively more types of error to the order of agents in the arena. Since \mcesmppac{} is trying to solve for a joint policy comprised of $i$ and $j$'s respective policies, $\pi_i$ and $\pi_j$, there are the following types of error. In this proof, I establish bounds for only two agents for the sake of brevity, but it is straightforward to apply it to more than two agents.

We use $\mathcal{N}(\pi)$ as a function to denote the local neighborhood of policies that varies from $\pi$ by a single action-observation pair.

\begin{itemize}
\item[(A)] After seeing $n$ samples (where $n<k$), \mcesmppac{} could climb from some joint policy $\{\pi^m_i,\pi^m_j\}$ to a joint neighboring policy $\{\pi'_i,\pi'_j\}$ where $\pi'_i\in \mathcal{N}(\pi^m_i)$ and $\pi'_j\in\mathcal{N}(\pi^m_i)$, because $\{\pi'_i,\pi'_j\}$ appears to be better than $\{\pi^m_i,\pi^m_j\}$, but neither policy is an improvement
\item[(B)] After seeing $n$ samples (where $n<k$), \mcesmppac{} could climb from some joint policy $\{\pi^m_i,\pi^m_j\}$ to a joint neighboring policy $\{\pi'_i,\pi'_j\}$ where $\pi'_i\in \mathcal{N}(\pi^m_i)$ and $\pi'_j\in\mathcal{N}(\pi^m_i)$, because $\{\pi'_i,\pi'_j\}$ appears to be better than $\{\pi^m_i,\pi^m_j\}$, but $\pi'_i$ or $\pi'_j$ is not an improvement
\item[(C)] After seeing $n$ samples (where $n<k$), \mcesmppac{} cannot find a better joint policy $\{\pi'_i,\pi'_j\}$ where $\pi'_i\in \mathcal{N}(\pi^m_i)$ and $\pi'_j\in\mathcal{N}(\pi^m_i)$ that dominates joint policy $\{\pi^m_i,\pi^m_j\}$, as all $\pi'_i$ and $\pi'_j$ appear worse than $\{\pi^m_i,\pi^m_j\}$, but in fact there is a better joint policy
\item[(D)] After seeing $n$ samples (where $n<k$), \mcesmppac{} cannot find a better joint policy $\{\pi'_i,\pi'_j\}$ where $\pi'_i\in \mathcal{N}(\pi^m_i)$ and $\pi'_j\in\mathcal{N}(\pi^m_i)$ that dominates joint policy $\{\pi^m_i,\pi^m_j\}$, as either all $\pi'_i$ appear worse than $\pi^m_i$, or all $\pi'_j$ appear worse than $\pi^m_j$, but in fact there is a better joint policy
\item[(E)] After seeing all $k$ samples, \mcesmppac{} could climb from some joint policy $\{\pi^m_i,\pi^m_j\}$ to a joint neighboring policy $\{\pi'_i,\pi'_j\}$ where $\pi'_i\in \mathcal{N}(\pi^m_i)$ and $\pi'_j\in\mathcal{N}(\pi^m_i)$, because $\{\pi'_i,\pi'_j\}$ appears to be better than $\{\pi^m_i,\pi^m_j\}$, but neither policy is an improvement
\item[(F)] After seeing all $k$ samples, \mcesmppac{} could climb from some joint policy $\{\pi^m_i,\pi^m_j\}$ to a joint neighboring policy $\{\pi'_i,\pi'_j\}$ where $\pi'_i\in \mathcal{N}(\pi^m_i)$ and $\pi'_j\in\mathcal{N}(\pi^m_i)$, because $\{\pi'_i,\pi'_j\}$ appears to be better than $\{\pi^m_i,\pi^m_j\}$, but either $\pi'_i$ or $\pi'_j$ is not an improvement
\item[(G)] After seeing all $k$ samples, \mcesmppac{} cannot find a better joint policy $\{\pi'_i,\pi'_j\}$ where $\pi'_i\in \mathcal{N}(\pi^m_i)$ and $\pi'_j\in\mathcal{N}(\pi^m_i)$ that dominates joint policy $\{\pi^m_i,\pi^m_j\}$, as all $\pi'_i$ and $\pi'_j$ appear worse than $\{\pi^m_i,\pi^m_j\}$, but in fact there is a better joint policy
\item[(H)] After seeing all $k$ samples, \mcesmppac{} cannot find a better joint policy $\{\pi'_i,\pi'_j\}$ where $\pi'_i\in \mathcal{N}(\pi^m_i)$ and $\pi'_j\in\mathcal{N}(\pi^m_i)$ that dominates joint policy $\{\pi^m_i,\pi^m_j\}$, as either all $\pi'_i$ appear worse than $\pi^m_i$, or all $\pi'_j$ appear worse than $\pi^m_j$, but in fact there is a better joint policy
\end{itemize}

We apply the following to decide when to climb from one individual policy, $\pi$, to another, $\pi'$.

\begin{align*}
\epsilon(m,p,q,k_m) = \left\{
\begin{array}{lr}
\Lambda(\pi,\pi')\sqrt{\frac{1}{2p}ln\left(\frac{\sqrt[4]{6(k_m-1)}N}{\sqrt[4]{\delta_m}}\right)} & \text{if } p=q < k_m\\
\frac{\epsilon}{2} & \text{if } p = q = k_m \\
+\infty & \text{otherwise}
\end{array}
\right.
\end{align*}
where $\Lambda$ is defined in Sec.~{4} and $N=|\mathcal{N}(\pi^m)|$ is the size of the local neighborhood. Let $\vec{\pi}=\langle \pi_i,\pi_j\rangle$. Let $E^i[\vec{\pi}]$ represent the true value, as opposed to empirically sampled value $Q$, of following a policy $\pi_i$ when the other agent is following $\pi_j$, and $E^j[\vec{\pi}]$ analogously for the other agent. As it is clear from context and for the interest of brevity, I omit $k_m$ from $\epsilon(\cdot)$. I represent the previously mentioned types of error in \mcesmppac{} with the following expressions.

\small
\begin{align*}
a^n_m =& Pr[\exists\{\pi'_i,\pi'_j\}\in\mathcal{N}(\pi^m_i)\times\mathcal{N}(\pi^m_j):   (Q^{\pi'_i}-Q^{\pi^m_i})\geq\epsilon(m, c^{\pi'_i}_i, c^{\pi^m_i}_i), \\
& (Q^{\pi'_j}-Q^{\pi^m_j})\geq\epsilon(m, c^{\pi'_j}_j, c^{\pi^m_j}_j), \text{and }  E^i[\vec{\pi}']<E^i[\vec{\pi}^m], E^j[\vec{\pi}']<E^j[\vec{\pi}^m]]
\end{align*}
\begin{align*}
b^n_m =& Pr[\exists\{\pi'_i,\pi'_j\}\in\mathcal{N}(\pi^m_i)\times\mathcal{N}(\pi^m_j):  (Q^{\pi'_i}-Q^{\pi^m_i})\geq\epsilon(m, c^{\pi'_i}_i, c^{\pi^m_i}_i),   (Q^{\pi'_j}-Q^{\pi^m_j})\geq\epsilon(m, c^{\pi'_j}_j, c^{\pi^m_j}_j), \\
&\text{and }   [(E^i[\vec{\pi}']<E^i[\vec{\pi}^m], E^j[\vec{\pi}']>E^j[\vec{\pi}^m]+\epsilon)   \text{ or }
 (E^i[\vec{\pi}']>E^i[\vec{\pi}^m]+\epsilon, E^j[\vec{\pi}']<E^j[\vec{\pi}^m])]]
\end{align*}
\begin{align*}
c^n_m =& Pr[\exists\{\pi'_i,\pi'_j\}\in\mathcal{N}(\pi^m_i)\times\mathcal{N}(\pi^m_j): (Q^{\pi'_i}-Q^{\pi^m_i})<\epsilon-\epsilon(m, c^{\pi'_i}_i, c^{\pi^m_i}_i),  \\& (Q^{\pi'_j}-Q^{\pi^m_j})<\epsilon-\epsilon(m, c^{\pi'_j}_j, c^{\pi^m_j}_j),  \text{and }  E^i[\vec{\pi}']>E^i[\vec{\pi}^m]+\epsilon, E^j[\vec{\pi}']>E^j[\vec{\pi}^m]+\epsilon]
\end{align*}
\begin{align*}
d^n_m =& Pr[\exists\{\pi'_i,\pi'_j\}\in\mathcal{N}(\pi^m_i)\times\mathcal{N}(\pi^m_j):  (Q^{\pi'_i}-Q^{\pi^m_i})<\epsilon-\epsilon(m, c^{\pi'_i}_i, c^{\pi^m_i}_i),  (Q^{\pi'_j}-Q^{\pi^m_j})<\epsilon-\epsilon(m, c^{\pi'_j}_j, c^{\pi^m_j}_j), \\& \text{and }  [(E^i[\vec{\pi}']>E^i[\vec{\pi}^m]+\epsilon, E^j[\vec{\pi}']<C(\pi^m_j),\pi^m_i  \text{ or }
(E^i[\vec{\pi}']<E^i[\vec{\pi}^m], E^j[\vec{\pi}']>E^j[\vec{\pi}^m]+\epsilon)]]
\end{align*}
\begin{align*}
e_m =& Pr[\exists\{\pi'_i,\pi'_j\}\in\mathcal{N}(\pi^m_i)\times\mathcal{N}(\pi^m_j): (Q^{\pi'_i}-Q^{\pi^m_i})\geq\frac{\epsilon}{2}, (Q^{\pi'_j}-Q^{\pi^m_j})\geq\frac{\epsilon}{2}, \\& \text{and } E^i[\vec{\pi}']<E^i[\vec{\pi}^m], E^j[\vec{\pi}']<E^j[\vec{\pi}^m]]
\end{align*}
\begin{align*}
f_m =& Pr[\exists\{\pi'_i,\pi'_j\}\in\mathcal{N}(\pi^m_i)\times\mathcal{N}(\pi^m_j): (Q^{\pi'_i}-Q^{\pi^m_i})\geq\frac{\epsilon}{2}, (Q^{\pi'_j}-Q^{\pi^m_j})\geq\frac{\epsilon}{2}, \\& \text{and } [(E^i[\vec{\pi}']<E^i[\vec{\pi}^m], E^j[\vec{\pi}']>E^j[\vec{\pi}^m]+\epsilon)  \text{ or }
 (E^i[\vec{\pi}']>E^i[\vec{\pi}^m]+\epsilon, E^j[\vec{\pi}']<E^j[\vec{\pi}^m])]]
\end{align*}
\begin{align*}
g_m =& Pr[\exists\{\pi'_i,\pi'_j\}\in\mathcal{N}(\pi^m_i)\times\mathcal{N}(\pi^m_j): (Q^{\pi'_i}-Q^{\pi^m_i})<\frac{\epsilon}{2}, (Q^{\pi'_j}-Q^{\pi^m_j})<\frac{\epsilon}{2}, \\ &\text{and } E^i[\vec{\pi}']>E^i[\vec{\pi}^m]+\epsilon, E^j[\vec{\pi}']>E^j[\vec{\pi}^m]+\epsilon]
\end{align*}
\begin{align*}
h_m =& Pr[\exists\{\pi'_i,\pi'_j\}\in\mathcal{N}(\pi^m_i)\times\mathcal{N}(\pi^m_j): (Q^{\pi'_i}-Q^{\pi^m_i})<\frac{\epsilon}{2}, (Q^{\pi'_j}-Q^{\pi^m_j})<\frac{\epsilon}{2}, \\& \text{and } [(E^i[\vec{\pi}']>E^i[\vec{\pi}^m]+\epsilon, E^j[\vec{\pi}']<C(\pi^m_j),\pi^m_i) \text{ or }
(E^i[\vec{\pi}']<E^i[\vec{\pi}^m], E^j[\vec{\pi}']>E^j[\vec{\pi}^m]+\epsilon)]]
\end{align*}
\normalsize

We can eliminate the existential quantifiers in the previous expressions by representing them as finite disjunctions over the elements of the joint local neighborhood, $\mathcal{N}(\vec{\pi})$. For the individual agent, considering a set of neighboring policies that are truly worse than the current policy given the other agent is fixed can be represented as: $\mathcal{N}_i^<(\vec{\pi}^m) = \{\pi'_i\in\mathcal{N}(\pi^m_i)|E^i[\pi'_i,\pi^m_j]<E^i[\pi^m_i,\pi^m_j]\}$, and similarly for agent $j$. Subsequently, I represent the neighborhood of joint policies where both agents transformed policies are truly worse as $\mathcal{N}^{<<}(\vec{\pi}^m) = \{\vec{\pi}' \in\mathcal{N}(\vec{\pi}^m)|E^i[\vec{\pi}']<E^i[\vec{\pi}^m]\wedge E^j[\vec{\pi}']<E^j[\vec{\pi}^m]\}$. I omit the parameter as it is clear from context, resulting in $\mathcal{N}^{<<}$. Note that $|\mathcal{N}^{<<}|=|\mathcal{N}^{<}_i|\cdot|\mathcal{N}^{<}_j|$.

\begin{align*}
a^n_m = Pr[\bigvee\limits_{\vec{\pi}'\in\mathcal{N}^{<<}} &(Q^{\pi'_i}-Q^{\pi^m_i})\geq\epsilon(m, c^{\pi'_i}_i, c^{\pi^m_i}_i) \wedge (Q^{\pi'_j}-Q^{\pi^m_j})\geq\epsilon(m, c^{\pi'_j}_j, c^{\pi^m_j}_j)]
\end{align*}

We can also express probabilities for sets of policies that are truly greater as well, using  $\mathcal{N}_i^>(\vec{\pi}^m) = \{\pi'_i\in\mathcal{N}(\pi^m_i)|C(\pi'_i,\pi^m_j)>C(\pi^m_i,\pi^m_j)+\epsilon\}$. Again, this is written similarly for agent $j$. The joint notation follows identically as above.

\begin{align*}
c^n_m = Pr[\bigvee\limits_{\vec{\pi}'\in\mathcal{N}^{>>}} &(Q^{\pi'_i}-Q^{\pi^m_i})<\epsilon-\epsilon(m, c^{\pi'_i}_i, c^{\pi^m_i}_i) \wedge (Q^{\pi'_j}-Q^{\pi^m_j})<\epsilon-\epsilon(m, c^{\pi'_j}_j, c^{\pi^m_j}_j)]
\end{align*}

The other errors are expressed in a similar manner. I then evaluate the bounds as follows.

\small
\begin{align*}
a^n_m &\leq \sum\limits_{\vec{\pi}'\in\mathcal{N}^{<<}} Pr[(Q^{\pi'_i}-Q^{\pi^m_i})\geq\epsilon(m, c^{\pi'_i}_j, c^{\pi^m_i}_i) \wedge (Q^{\pi'_j}-Q^{\pi^m_j})\geq\epsilon(m, c^{\pi'_j}_j, c^{\pi^m_j}_j)]
\\
&\leq \sum\limits_{\vec{\pi}'\in\mathcal{N}^{<<}} Pr[(Q^{\pi'_i}-Q^{\pi^m_i})\geq (E^i[\vec{\pi}']-E^i[\vec{\pi}^m]) + \epsilon(m, c^{\pi'_i}_i, c^{\pi^m_i}_i)]\cdot \\
& \phantom{===} Pr[(Q^{\pi'_j}-Q^{\pi^m_j})\geq (E^j[\vec{\pi}']-E^j[\vec{\pi}^m]) + \epsilon(m, c^{\pi'_j}_j, c^{\pi^m_j}_j)] \\
&\leq \sum\limits_{\vec{\pi}'\in\mathcal{N}^{<<}} exp\left\{-2p\left[\left(\frac{\epsilon(m, c^{\pi'_i}_i, c^{\pi^m_i}_i)}{\Lambda(\pi_i,\pi'_i)}\right)^2+ \left(\frac{\epsilon(m, c^{\pi'_j}_j, c^{\pi^m_j}_j)}{\Lambda(\pi_j,\pi'_j)}\right)^2\right]\right\}\numberthis \label{eqn:hoeffding}
\\
&= |\mathcal{N}^{<<}| \frac{\sqrt{\delta_m}}{\sqrt{6}\sqrt{k_m-1}|\mathcal{N}(\pi^m_i)|}\cdot \frac{\sqrt{\delta_m}}{\sqrt{6}\sqrt{k_m-1}|\mathcal{N}(\pi^m_j)|}  = \frac{|\mathcal{N}^{<<}| \delta_m}{6(k_m-1)|\mathcal{N}(\vec{\pi}^m)|}
\end{align*}
\normalsize

Line~\ref{eqn:hoeffding} follows from Hoeffding's Inequality, where $(Q^{\pi'_i}-Q^{\pi^m_i})$ is the sample average approximating $E^i[\vec{\pi}']-E^i[\vec{\pi}^m]$. 5 is reduced when satisfying the condition $p=q<k_m$. $B$ is computed similarly, but the space of possible neighbors is over the union of $\mathcal{N}^{<>}$ and $\mathcal{N}^{><}$.

\small
\begin{align*}
b^n_m &\leq \sum\limits_{\vec{\pi}'\in\mathcal{N}^{<>}\cup\mathcal{N}^{><}} Pr[ (Q^{\pi'_i}-Q^{\pi^m_i})\geq\epsilon(m, c^{\pi'_i}_i, c^{\pi^m_i}_i)\wedge (Q^{\pi'_j}-Q^{\pi^m_j})\geq\epsilon(m, c^{\pi'_j}_j, c^{\pi^m_j}_j)]
\\
&\leq \frac{(|\mathcal{N}^{<>}|+|\mathcal{N}^{><}|) \delta_m}{6(k_m-1)|\mathcal{N}(\vec{\pi}^m)|}
\end{align*}
\normalsize

$C$ follows from $A$, but in the case where \mcesmppac{} terminates without transformation.

\small
\begin{align*}
c^n_m &\leq \sum\limits_{\vec{\pi}'\in\mathcal{N}^{>>}} Pr[ (Q^{\pi'_i}-Q^{\pi^m_i})\leq \epsilon - \epsilon(m, c^{\pi'_i}_i, c^{\pi^m_i}_i)\wedge (Q^{\pi'_j}-Q^{\pi^m_j})\leq\epsilon - \epsilon(m, c^{\pi'_j}_j, c^{\pi^m_j}_j)]
\\
&\leq \sum\limits_{\vec{\pi}'\in\mathcal{N}^{>>}} Pr[(Q^{\pi'_i}-Q^{\pi^m_i})<  (E^i[\vec{\pi}']-E^i[\vec{\pi}^m]) - \epsilon(m, c^{\pi'_i}_i, c^{\pi^m_i}_i)]\cdot \\
& \phantom{===} Pr[(Q^{\pi'_j}-Q^{\pi^m_j})<  (E^j[\vec{\pi}']-E^j[\vec{\pi}^m]) - \epsilon(m, c^{\pi'_j}_j, c^{\pi^m_j}_j)] \numberthis \label{eqn:termeps}\\
&= \frac{|\mathcal{N}^{>>}| \delta_m}{6(k_m-1)|\mathcal{N}(\vec{\pi}^m)|}
\end{align*}
\normalsize

Though the sign is reversed in $C$, $\epsilon$ falls under the square and, as such, the proof follows identically to $A$ from Line~\ref{eqn:termeps}. As with $B$, $D$ just expands the neighborhood.

\small
\begin{align*}
d^n_m &\leq \sum\limits_{\vec{\pi}'\in\mathcal{N}^{><}\cup\mathcal{N}^{<>}} Pr[ (Q^{\pi'_i}-Q^{\pi^m_i})<\epsilon-\epsilon(m, c^{\pi'_i}_i, c^{\pi^m_i}_i)\wedge Q^{\pi'_j}-Q^{\pi^m_j})<\epsilon-\epsilon(m, c^{\pi'_j}_j, c^{\pi^m_j}_j)]
\\
&\leq \frac{(|\mathcal{N}^{><}|+|\mathcal{N}^{<>}|) \delta_m}{6(k_m-1)|\mathcal{N}(\vec{\pi}^m)|}
\end{align*}
\normalsize

We can bound at $p=k_m$ samples in the same fashion.

\small
\begin{align*}
e_m &\leq \sum\limits_{\vec{\pi}'\in\mathcal{N}^{<<}} Pr[ (Q^{\pi'_i}-Q^{\pi^m_i})\geq\frac{\epsilon}{2}\wedge  (Q^{\pi'_j}-Q^{\pi^m_j})\geq\frac{\epsilon}{2}]
\\
&\leq \sum\limits_{\vec{\pi}'\in\mathcal{N}^{<<}} exp\left\{-2k_m\left[\left(\frac{\epsilon/2}{\Lambda(\pi_i,\pi'_i)}\right)^2+ \left(\frac{\epsilon/2}{\Lambda(\pi_j,\pi'_j)}\right)^2\right]\right\} = \frac{|\mathcal{N}^{<<}|\delta_m}{6|\mathcal{N}(\vec{\pi}^m)|} \numberthis \label{eqn:kmreplace}
\end{align*}
\normalsize

where Line~\ref{eqn:kmreplace} is obtained by replacing $k_m$ with its derived value. $F$ is obtained by following the process in $B$ and the reduction in Line~\ref{eqn:kmreplace}.

\small
\begin{align*}
f_m &\leq \sum\limits_{\vec{\pi}'\in\mathcal{N}^{<>}\cup\mathcal{N}^{><}} Pr[ (Q^{\pi'_i}-Q^{\pi^m_i})\geq\frac{\epsilon}{2}\wedge (Q^{\pi'_j}-Q^{\pi^m_j})\geq\frac{\epsilon}{2}]\leq\frac{(|\mathcal{N}^{<>}|+|\mathcal{N}^{><}|)\delta_m}{6|\mathcal{N}(\vec{\pi}^m)|}
\end{align*}
\normalsize

$G$ and $H$ similarly follow from $C$ and $D$, replacing $k_m$ with its derived value.

\small
\begin{align*}
g_m &\leq \sum\limits_{\vec{\pi}'\in\mathcal{N}^{>>}} Pr[ (Q^{\pi'_i}-Q^{\pi^m_i})<\frac{\epsilon}{2}\wedge (Q^{\pi'_j}-Q^{\pi^m_j})<\frac{\epsilon}{2}]\leq \frac{|\mathcal{N}^{>>}|\delta_m}{6|\mathcal{N}(\vec{\pi}^m)|}
\end{align*}

\begin{align*}
h_m &\leq \sum\limits_{\vec{\pi}'\in\mathcal{N}^{><}\cup\mathcal{N}^{<>}} Pr[ (Q^{\pi'_i}-Q^{\pi^m_i})<\frac{\epsilon}{2}\wedge  (Q^{\pi'_j}-Q^{\pi^m_j})<\frac{\epsilon}{2}]\leq\frac{(|\mathcal{N}^{><}|+|\mathcal{N}^{<>}|)\delta_m}{6|\mathcal{N}(\vec{\pi}^m)|}
\end{align*}
\normalsize

Then, the probability that \mcesmppac{} makes an error after $n$ transformations is

\small
\begin{align*}
\sum\limits_{n=1}^{k_m}&[a^n_m + b^n_m + c^n_m + d^n_m] + e_m + f_m + g_m + h_m \\
&\leq \left[\frac{(k_m - 1)\delta_m}{6(k_m-1)|\mathcal{N}(\vec{\pi}^m)|}(|\mathcal{N}^{<<}|+|\mathcal{N}^{<>}|+|\mathcal{N}^{><}|+ |\mathcal{N}^{>>}|+|\mathcal{N}^{><}|+|\mathcal{N}^{<>}|)\right]+\\
& \phantom{==}\frac{\delta_m}{6|\mathcal{N}(\vec{\pi}^m)|}(|\mathcal{N}^{<<}|+|\mathcal{N}^{<>}|+|\mathcal{N}^{><}|+|\mathcal{N}^{>>}|+ |\mathcal{N}^{><}|+|\mathcal{N}^{<>}|)\\
&\leq \left[\frac{\delta_m}{6|\mathcal{N}(\vec{\pi}^m)|}(|N(\vec{\pi}^m)|+|\mathcal{N}^{<>}|+|\mathcal{N}^{><}|)\right]+ \frac{\delta_m}{6|\mathcal{N}(\vec{\pi}^m)|}(|\mathcal{N}(\vec{\pi}^m)|+|\mathcal{N}^{<>}|+|\mathcal{N}^{><}|) \numberthis \label{eqn:nbrreduce}\\
&\leq \left[\frac{\delta_m}{6|\mathcal{N}(\vec{\pi}^m)|}(3|N(\vec{\pi}^m)|\right]+\frac{\delta_m}{6|\mathcal{N}(\vec{\pi}^m)|}(3|\mathcal{N}(\vec{\pi}^m)|) \numberthis \label{eqn:nbrcomb}\\
&= \frac{\delta_m}{2}+\frac{\delta_m}{2}= \delta_m
\end{align*}
\normalsize

Line~\ref{eqn:nbrreduce} follows from the observations that $\mathcal{N}^{**}=\mathcal{N}^*_i\times\mathcal{N}^*_j$ and $|\mathcal{N}^<_i| + |\mathcal{N}^>_i| \leq |\mathcal{N}_i|$. Line~\ref{eqn:nbrcomb} follows from the observation that any product of subsets of $\mathcal{N}$ is bounded by the entire set, such as $|\mathcal{N}_i^{<<}| \leq |\mathcal{N}|$. Now I show that, over all transformations, $\delta_m$ is bound by $\delta$.

\begin{align*}
&\sum\limits_{m=1}^{\infty}[\sum\limits_{n=1}^k[a^n_m + b^n_m + c^n_m + d^n_m] + e_m + f_m + g_m + h_m] \\
&\leq \sum\limits_{m=1}^{\infty}\delta_m=\sum\limits_{m=1}^{\infty}\frac{6\delta}{m^2\pi^2}=\frac{6\delta}{\pi^2}\sum\limits_{m=1}^{\infty}\frac{1}{m^2}=\frac{6\delta}{\pi^2}\frac{\pi^2}{6}=\delta
\end{align*}
\normalsize

\section{Proof of Prop.~\ref{prop:sample_complexity}}
\label{sec:samplecomplexity_proof}

Consider the following expanded definition of the above proposition.\\

\begin{align*}
&\sum\limits_{t=0}^T[\max\limits_{s}\{R(a_{\pi'_i}^t,a_j^t,s)-R(a_{\pi^k_i}^t,a_j^t,s)\} - \min\limits_s\{R(a_{\pi'_i}^t,a_j^t,s)-R(a_{\pi_i^k}^t,a_j^t,s)\}]\\
&\leq \sum\limits_{t=0}^T[\max\limits_{s,a_j}\{R(a_{\pi'_i}^t,a_j,s)-R(a_{\pi^k_i}^t,a_j,s)\} -  \min\limits_{s,a_j}\{R(a_{\pi'_i}^t,a_j,s)-R(a_{\pi_i^k}^t,a_j,s)\}]
\end{align*}

\begin{itemize}
\item Assume $\text{min}_s=\text{min}_{s,a_j}$ for all $t\in T$, resulting in the expression $\sum\limits_{t=0}^T\text{max}_{s}\{R(a_{\pi'_i}^t,a_j^t,s)-R(a_{\pi^k_i}^t,a_j^t,s)\}\leq\\\sum\limits_{t=0}^T\text{max}_{s,a_j}\{R(a_{\pi'_i}^t,a_j,s)-R(a_{\pi^k_i}^t,a_j,s)\}$. It must be the case that the LHS must be at most equivalent to the RHS, as, if the $a_j$ selected in the LHS is the maximal value, the $\text{max}_{s,a_j}$ will select it. If the $a_j$ selected on the LHS is not the maximal value for a given $a_i$ and $s$, the RHS must then be greater.
\item Assume $\text{max}_s=\text{max}_{s,a_j}$ for all $t\in T$, resulting in the expression $\sum\limits_{t=0}^T\text{min}_{s}\{R(a_{\pi'_i}^t,a_j^t,s)-R(a_{\pi^k_i}^t,a_j^t,s)\}\geq\\\sum\limits_{t=0}^T[\text{min}_{s,a_j}\{R(a_{\pi'_i}^t,a_j,s)-R(a_{\pi^k_i}^t,a_j,s)\}]$. Analogously to point 1, if the $a_j$ on the LHS is the minimal value, then the $\text{min}_{s,a_j}$ on the RHS must select it. If it is not, the RHS must be less.
\item Following point 1 and 2, since each component of the LHS must be bounded by the equivalent component on the RHS, the LHS must be a smaller range than the RHS. Therefore, $\Lambda^{\vec{a_j}}(\pi'_i, \pi^m_i)\leq\Lambda(\pi'_i,\pi^m_i,\pi_j)$
\end{itemize}

\cleardoublepage
\phantomsection
\addcontentsline{toc}{chapter}{References}
\bibliographystyle{plain}
\bibliography{dissertation}

\end{document}